\documentclass[10pt,twocolumn,letterpaper]{article}

\usepackage[pagenumbers]{cvpr} 

\usepackage{graphicx}
\usepackage{amsmath}
\usepackage{amssymb}
\usepackage{booktabs}

\usepackage{times}
\usepackage{epsfig}
\usepackage{graphicx}
\usepackage{amsmath}
\usepackage{amssymb}
\usepackage{color} 
\usepackage[toc,page]{appendix}

\usepackage{amsfonts}
\usepackage{array}
\usepackage{booktabs}
\usepackage{mathtools}
\usepackage{microtype}
\usepackage{multirow}
\usepackage{nicefrac}
\usepackage{url}
\usepackage{amsfonts}
\usepackage{xcolor}
\usepackage{pifont}
\usepackage{float}
\usepackage{enumitem}
\graphicspath{ {./figures/} }

\usepackage[pagebackref=true,breaklinks=true,letterpaper=true,colorlinks,bookmarks=false]{hyperref}

\hypersetup{
  colorlinks=true,
  linkcolor=red,
  urlcolor=magenta,
}

\usepackage{tabulary}

\usepackage{graphbox}

\usepackage{tabularx}
\newcolumntype{Y}{>{\centering\arraybackslash}X}

\usepackage{graphicx,overpic,xcolor}
\usepackage{longtable}
\usepackage{colortbl}
\usepackage{wrapfig}
\usepackage{bbding}

\newcolumntype{x}[1]{>{\centering\arraybackslash}p{#1pt}}
\newcolumntype{y}[1]{>{\raggedright\arraybackslash}p{#1pt}}
\newcolumntype{z}[1]{>{\raggedleft\arraybackslash}p{#1pt}}

\newlength\savewidth\newcommand\shline{\noalign{\global\savewidth\arrayrulewidth
  \global\arrayrulewidth 1pt}\hline\noalign{\global\arrayrulewidth\savewidth}}
  
\newcommand{\tablestyle}[2]{\setlength{\tabcolsep}{#1}\renewcommand{\arraystretch}{#2}\centering\footnotesize}

\def\eg{\textit{e.g.}}
\def\ie{\textit{i.e.}}
\def\Eg{\textit{E.g.}}

\newcommand{\model}{GSS}
\newcommand{\gssa}{FF-R}
\newcommand{\gssb}{FF}
\newcommand{\gssc}{FT}
\newcommand{\gssd}{TF}
\newcommand{\gsse}{TT}
\newcommand{\fx}{\mathcal{X}}
\newcommand{\fxpi}{\mathcal{X}^{-1}}

\makeatletter\renewcommand\paragraph{\@startsection{paragraph}{4}{\z@}
  {.5em \@plus1ex \@minus.2ex}{-.5em}{\normalfont\normalsize\bfseries}}\makeatother

\usepackage[capitalize]{cleveref}
\crefname{section}{Sec.}{Secs.}
\Crefname{section}{Section}{Sections}
\Crefname{table}{Table}{Tables}
\crefname{table}{Tab.}{Tabs.}

\def\cvprPaperID{1187} 
\def\confName{CVPR}
\def\confYear{2023}

\begin{document}

\title{Generative Semantic Segmentation}

\author{%
  Jiaqi Chen$^1$
  \quad 
  Jiachen Lu$^1$
  \quad 
  Xiatian Zhu$^2$ 
  \quad 
  Li Zhang$^1$\thanks{Li Zhang (lizhangfd@fudan.edu.cn) is the corresponding author with School of Data Science, Fudan University.}
  \\
  $^1$Fudan University 
  \quad
  $^2$University of Surrey 
  \vspace{.5em} 
  \\
  \textcolor{black}{\url{https://github.com/fudan-zvg/GSS}}
}

\maketitle

\begin{abstract}
We present {\bf\em Generative Semantic Segmentation} (GSS), a generative learning approach for semantic segmentation. Uniquely, we cast semantic segmentation as an {\bf\em image-conditioned mask generation problem}. This is achieved by replacing the conventional per-pixel discriminative learning with a latent prior learning process. Specifically, we model the variational posterior distribution of latent variables given the segmentation mask. To that end, the segmentation mask is expressed with a special type of image (dubbed as {\em maskige}). This posterior distribution allows to generate segmentation masks unconditionally. To achieve semantic segmentation on a given image, we further introduce a conditioning network. It is optimized by minimizing the divergence between the posterior distribution of maskige (\ie{} segmentation masks) and the latent prior distribution of input training images. Extensive experiments on standard benchmarks show that our GSS can perform competitively to prior art alternatives in the standard semantic segmentation setting, whilst achieving a new state of the art in the more challenging cross-domain setting.
\end{abstract}

\section{Introduction}
The objective of semantic segmentation is to predict a label for every single pixel of an input image \cite{long2015fully}.
Conditioning on each pixel's observation, existing segmentation methods
\cite{chen2017rethinking,zheng2021rethinking,xie2021segformer, cheng2021per} naturally adopt the {\em discriminative learning} paradigm, along with dedicated efforts on integrating task prior knowledge (\eg{}, spatial correlation)
\cite{zheng2021rethinking,cheng2021per, ji2021fsnet,wanseaformer}.
For example, existing methods
\cite{chen2017rethinking,zheng2021rethinking,xie2021segformer} typically use a linear projection to optimize the log-likelihood classification for each pixel.
Despite the claim of subverting per-pixel classification, the bipartite matching-based semantic segmentation~\cite{cheng2021per,cheng2022masked} still 
cannot avoid the per-pixel max log-likelihood.

\begin{figure}[t]
	\begin{center}
		\includegraphics[width=1.0\linewidth]{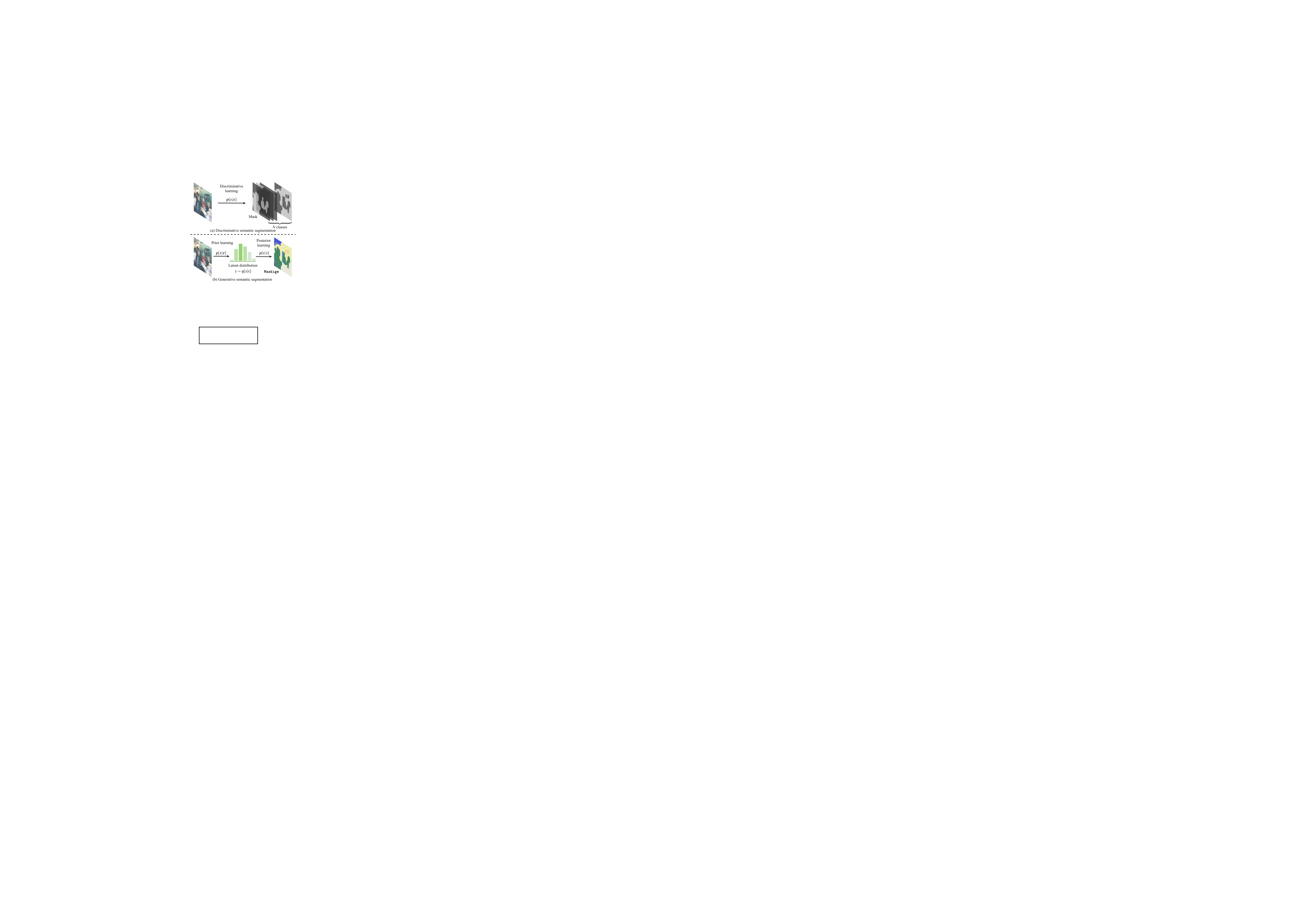}
	\end{center}
	\vspace{-0.5em}
	\caption{
 Schematic comparison between ({\bf a}) conventional discriminative learning and 
 ({\bf b}) our generative learning based model for semantic segmentation.
 Our \model{} introduces a latent variable $z$ and, given the segmentation mask $c$, it learns the posterior distribution of $z$ subject to the reconstruction constraint.
Then, we train a conditioning network to model the prior of $z$ by aligning with the corresponding posterior distribution.
This formulation can thus generate the segmentation mask for an input image.
	}
	\vspace{-1em}
	\label{fig:intro}
\end{figure}
In this paper, we introduce a new approach, {\bf\em Generative Semantic Segmentation} (GSS), that formulates semantic segmentation as {\em an image-conditioned mask generation problem}.
This conceptually differs from the conventional formulation of discriminative per-pixel classification learning, based on the log-likelihood of a conditional probability
(\ie\ the classification probability of image pixels).
Taking the manner of image generation instead~\cite{kingma2013auto, van2017neural}, 
we generate the {\em whole} segmentation masks
with {\em an auxiliary latent variable distribution} introduced.
This formulation is not only simple and more task-agnostic,
but also facilitates the exploitation of off-the-shelf big generative models 
(\eg\ DALL$\cdot$E~\cite{ramesh2021zero} trained by 3 billion iterations on a 300 million open-image dataset, far beyond both the data scale and training cost of semantic segmentation).

However, achieving segmentation segmentation in a generic generation framework (\eg{} the Transformer architecture \cite{esser2021taming}) is non-trivial due to drastically different data format.
To address this obstacle, 
we propose a notion of \texttt{maskige} that expresses 
the segmentation mask in the RGB image form.
This enables the use of a pretrained latent posterior distribution (\eg\ VQVAE~\cite{van2017neural}) of existing generative models.
Our model takes a two-stage optimization: 
{\bf(i)} Learning the posterior distribution of the latent variables conditioned on the semantic segmentation masks so that the latent variables can simulate the target segmentation masks;
To achieve this, we introduce an fixed pre-trained VQVAE from~\cite{ramesh2021zero} and a couple of lightweight transformation modules, which can be trained with minimal cost, or they can be manually set up without requiring any additional training. 
In either case, the process is efficient and does not add significant overhead to the overall optimization.
{\bf(ii)} Minimizing the distance between the posterior distribution and the prior distribution of the latent variables given input training images and their masks, enabling to condition the generation of semantic masks on the input images.
This can be realized by a generic encoder-decoder style architecture (\eg\ a Transformer).

We summarize the {\em contributions} as follows. 
\textbf{(i)} We propose a {\bf\em Generative Semantic Segmentation} approach that reformulates semantic segmentation as an image-conditioned mask generation problem.
This represents a {\em conceptual shift} from conventional discriminative learning based paradigm.
\textbf{(ii)}
We realize a \model{} model in an established conditional image generation framework,
with minimal need for task-specific architecture and loss function modifications 
while fully leveraging the knowledge of off-the-shelf generative models.
\textbf{(iii)}
Extensive experiments on several semantic segmentation benchmarks
show that our \model{} is competitive with prior art models in the standard setting, whilst achieving a new state of the art 
in the more challenging and practical cross-domain setting (\eg\ MSeg \cite{lambert2020mseg}).
\section{Related work}
{\noindent \bf Semantic segmentation}
Since the inception of FCN~\cite{long2015fully}, semantic segmentation have flourished by various deep neural networks with ability to classify each pixel. 
The follow-up efforts then shift to improve the limited receptive field of these models.
For example, PSPNet~\cite{zhao2017pyramid} and DeepLabV2~\cite{chen2014semantic} aggregate multi-scale context between convolution layers.
Sequentially, 
Nonlocal~\cite{wang2018nonlocal}, CCNet~\cite{huang2019ccnet}, and DGMN~\cite{zhang2020dynamic} integrate the attention mechanism in the convolution structure.
Later on, Transformer-based methods (\eg\ SETR~\cite{zheng2021rethinking} and Segformer~\cite{xie2021segformer}) are proposed following the introduction of Vision Transformers.
More recently, MaskFormer~\cite{cheng2021per} and Mask2Former~\cite{cheng2022masked} realize semantic segmentation with bipartite matching. 
Commonly, all the methods adopt the discriminative pixel-wise classification learning paradigm.
This is in contrast to our generative semantic segmentation.

{\noindent \bf Image generation}
In parallel, generative models \cite{esser2021taming, ramesh2021zero} also excel. They are often optimized in a two-stage training process: 
(1) Learning data representation in the first stage and 
(2) building a probabilistic model of the encoding in the second stage.
For learning data representation, VAE~\cite{kingma2013auto} reformulates the autoencoder by variational inference.
GAN~\cite{goodfellow2020generative} plays a zero-sum game.
VQVAE~\cite{van2017neural} extends the image representation learning to discrete spaces, making it possible for language-image cross-model generation.
\cite{larsen2016autoencoding} replaces element-wise errors of VAE with feature-wise errors to capture data distribution.
For probabilistic model learning, some works \cite{rombach2020making, esser2020disentangling, xiao2019generative} use flow for joint probability learning.
Leveraging the Transformers to model the composition between condition and images, Esser et al.~\cite{esser2021taming} demonstrate the significance of
data representation (\ie\ the first stage result) for the challenging high-resolution image synthesis, obtained at high computational cost.
This result is inspiring to this work in the sense that the diverse and rich knowledge about data representation achieved in the first stage could be transferable across more tasks such as semantic segmentation.

{\noindent \bf Generative models for visual perception}
Image-to-image translation made
one of the earliest attempts in generative segmentation, with far less success in performance~\cite{isola2017image}.
Some good results were achieved
in limited scenarios such as face parts segmentation and Chest X-ray segmentation~\cite{li2021semantic}.
Replacing the discriminative classifier with a generative Gaussian Mixture model, GMMSeg~\cite{liang2022gmmseg} is claimed as generative segmentation,
but the most is still of discriminative modeling.
The promising performance of Pix2Seq~\cite{chen2021pix2seq} on several vision tasks leads to the prevalence of sequence-to-sequence task-agnostic vision frameworks.
For example,
Unified-I/O~\cite{lu2022unified} supports a variety of vision tasks within a single model by seqentializing each task to sentences.
Pix2Seq-D~\cite{chen2022generalist} deploys a hierarchical VAE (\ie{} diffusion model) to generate panoptic segmentation masks.
This method is inefficient due to the need for iterative denoising.
UViM~\cite{kolesnikov2022uvim} realizes its generative panoptic segmentation by introducing latent variable conditioned on input images.
It is also computationally heavy due to the need for model training 
from scratch.
To address these issues,
we introduce a notion of {\tt maskige} for expressing segmentation masks
in the form of RGB images,
enabling the adopt of off-the-shelf data representation models (\eg\ VGVAE)
already pretrained on vast diverse imagery.
This finally allows for generative segmentation model training as efficiently as conventional discriminative counterparts.

\section{Methodology}
\subsection{GSS formulation}

\begin{figure*}[htb]
	\begin{center}
		\includegraphics[width=1.0\linewidth]{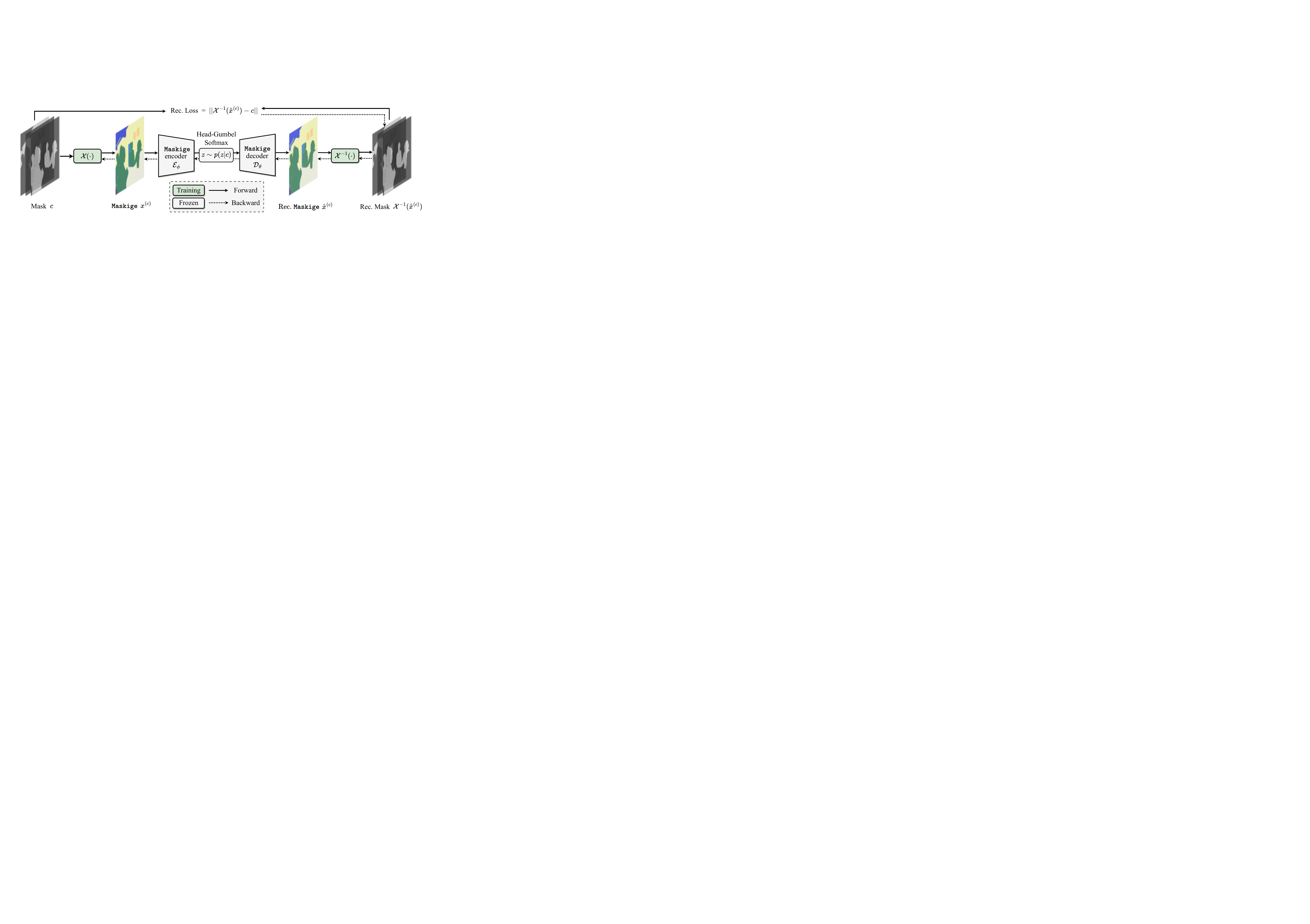}
	\end{center}
	\vspace{-1em}
	\caption{
\textbf{An illustration of our efficient latent posterior learning.}
Instead of training a \texttt{maskige} encoder-decoder, we utilize a well pretrained VQVAE~\cite{van2017neural} (gray blocks) and optimize the transformations $\mathcal{X}$ and $\mathcal{X}^{-1}$ (green blocks).
To optimize $\mathcal{X}$ and $\mathcal{X}^{-1}$ with gradient descent, we employ the Gumbel softmax relaxation technique~\cite{maddison2016concrete}. ``Rec.'': Reconstructed.
	}
	\vspace{-1em}
	\label{fig:stage1}
\end{figure*}
\begin{figure}[tb]
	\begin{center}
\includegraphics[width=1.0\linewidth]{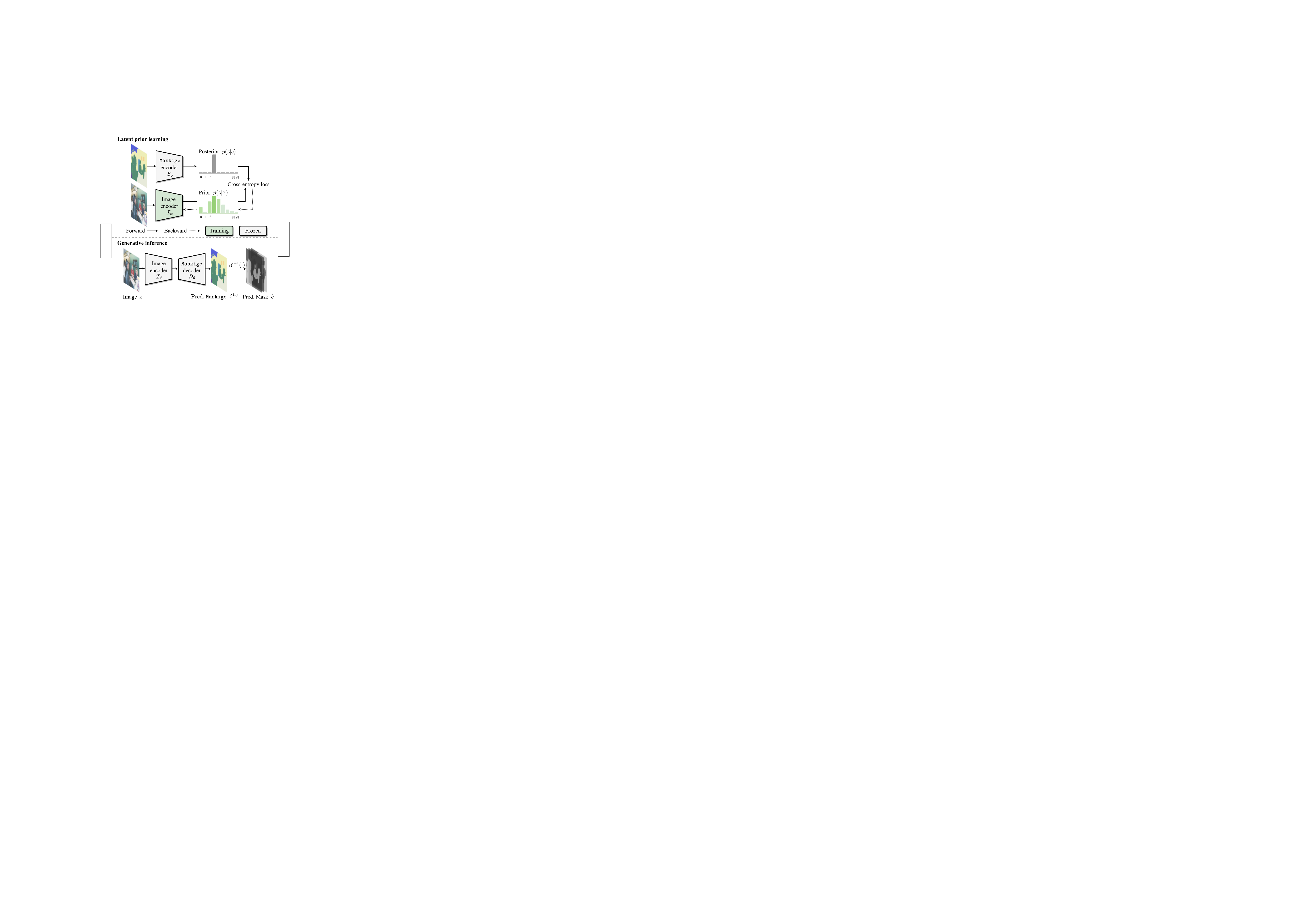}
	\end{center}
	\vspace{-1em}
	\caption{
\textbf{Illustration of latent prior learning} {(\bf\em top)} and \textbf{generative inference pipeline} {(\bf\em bottom)}.
During training, for latent prior learning, we optimize the image encoder $\mathcal{I}_\psi$ while freezing the {\tt maskige} encoder $\mathcal{E}_\phi$.
The objective is to minimize the divergence (\eg\ cross entropy loss) between the prior distribution and the posterior distribution of latent tokens.
During generative inference, we use the prior $z \sim p(z|x)$ inferred by $\mathcal{I}_\psi$ to generate the \texttt{maskige} with {\tt maskige} decoder $\mathcal{D}_\theta$.
``Pred.'': Predicted.
}
\vspace{-1em}
	\label{fig:stage2}
\end{figure}
Traditionally, semantic segmentation is formulated as a discriminative learning problem as
\begin{equation}
\label{equ:dis_ss}
    \max_\pi \log p_\pi(c|x)
\end{equation}
where $x\in \mathbb{R}^{H\times W\times3}$ is an input image, 
$c\in \{0,1\}^{ H\times W\times K}$ is a {\em segmentation mask} in $K$ semantic categories, and $p_\pi$ is a discriminative pixel classifier.
Focusing on learning the classification boundary of input pixels,
this approach enjoys high data and training efficiency~\cite{murphy2012machine}.

In this work, we introduce {\bf\em Generative Semantic Segmentation} (\model) by introducing a discrete $L$-dimension latent distribution $q_\phi(z|c)$ (with $z\in \mathbb{Z}^L$) to the above log-likelihood as:
\begin{align}
\label{eq:elbo2}
    \notag \log p(c|x) \geq \mathbb{E}_{q_\phi(z|c)}\left[
    \log \frac{p(z,c|x)}{q_{\phi}(z|c)}
    \right],
\end{align}
which is known as the Evidence Lower Bound (ELBO)~\cite{kingma2013auto} (details are given in the supplementary material).
Expanding the ELBO gives us
\begin{equation}
\label{equ:elbo_sem}
\mathbb{E}_{ q_{\phi}(z|c)} \left[
    \log p_\theta(c|z)\right] - D_{KL}\Big(q_{\phi}(z|c), p_{\psi}(z|x)\Big),
\end{equation}
where we have three components in our formulation:

\noindent \textbullet~$p_\psi$: An \textbf{image encoder} (denoted as $\mathcal{I}_\psi$) that models {\em the prior distribution} of latent tokens $z$ conditioned on the input image $x$.

\noindent \textbullet~$q_\phi$: A function that encodes the semantic segmentation mask $c$ into discrete latent tokens $z$, which includes a \textbf{\texttt{maskige} encoder} (denoted as $\mathcal{E}_\phi$, implemented by a VQVAE encoder~\cite{van2017neural}) and a linear projection (denoted as $\fx$, which will be detailed in Section~\ref{sec:maskige}).

\noindent \textbullet~$p_\theta$: A function that decodes the semantic segmentation mask $c$ from the discrete latent tokens $z$, which includes a 
\textbf{\texttt{maskige} decoder} (denoted $\mathcal{D}_\theta$, implemented by a VQVAE decoder~\cite{van2017neural}) and $\fxpi$ (the inverse process of $\fx$).

\paragraph{Architecture} 
The architecture of \model{} comprises three components: $\mathcal{I}_\psi$, $\mathcal{E}_\phi$ and $\mathcal{D}_\theta$. 
$\mathcal{E}\phi$ and $\mathcal{D}_\theta$ are implemented as VQVAE encoder and decoder~\cite{van2017neural}, respectively. 
Meanwhile, $\mathcal{I}_\psi$ is composed of an image backbone R(\eg\ esNet~\cite{he2016deep} or Swin Transformer~\cite{liu2021swin}) and a \emph{{M}ulti-{L}evel {A}ggregation} (MLA).
As an essential part, MLA is constructed using $D$ shifted window Transformer layers~\cite{liu2021swin} and a linear projection layer. The resulting output is a discrete code $z\in\mathbb{Z}^{H/d\times W/d}$, where $d$ denotes the downsample ratio. 

\paragraph{Optimization} 
Compared to the log-likelihood in discriminative models, optimizing the ELBO of a general model is more challenging \cite{murphy2012machine}.
End-to-end training cannot reach a global optimization.
For Eq. \eqref{equ:elbo_sem}, often we name the first term $\mathbb{E}_{ q_{\phi}(z|c)} \left[\log p_\theta(c|z)\right]$ as a reconstruction term and the second KL-divergence as the prior term.
In the next section we will introduce the optimization of this ELBO.

\subsection{ELBO optimization for semantic segmentation}
The ELBO optimization process for semantic segmentation involves two main steps, as described in \cite{van2017neural}.
The first step is {\bf \em latent posterior learning}, also known as reconstruction (see Figure~\ref{fig:stage1}). Here, the ELBO is optimized with respect to $\theta$ and $\phi$ through training a VQVAE~\cite{van2017neural} to reconstruct the desired segmentation masks.
The second step is {\bf \em latent prior learning} (see Figure~\ref{fig:stage2}). Once $\theta$ and $\phi$ are fixed, an image encoder $\psi$ is optimized to learn the prior distribution of latent tokens given an input image.

Typically, the first stage of ELBO optimization is
both most important and most expensive (much more than many discriminative learning counterparts) \cite{esser2021taming}.
To address this challenge, we propose an efficient latent posterior learning process.

\subsection{Stage I: Efficient latent posterior learning}
\label{sec:maskige}
During the first stage (shown in Figure~\ref{fig:stage1}), the initial prior $p_\psi(z|x)$ is set as the uniform distribution.
Conventionally, the first stage latent posterior training is conducted by
\begin{equation}
\label{equ:opt_stage1}
    \min_{\theta, \phi} \mathbb{E}_{q_\phi(z|c)}\|p_\theta(c|z)-c\|.
\end{equation}
To optimize Eq.~\eqref{equ:opt_stage1} more efficiently, we introduce a random variable transformation $\fx: \mathbb{R}^{K} \to \mathbb{R}^{3}$, its pseudo-inverse function $\fxpi : \mathbb{R}^{3} \to \mathbb{R}^{K}$, and the \texttt{maskige} decoder $\mathcal{D}_\theta$.
Applying expectation with transformed random variable, we have
\begin{align}
    \notag&\quad \min_{\hat{\phi},\hat{\theta}} \mathbb{E}_{q_{\hat{\phi}}(\hat{z}|\mathcal{X}(c))}\|\mathcal{D}_{\hat{\theta}}(\hat{z})-\mathcal{X}(c)\|\\
    &+\min_{\fxpi}\mathbb{E}_{q_{\hat{\phi}}(\hat{z}|\mathcal{X}(c))}\|\mathcal{X}^{-1}(\mathcal{D}_{\hat{\theta}}(\hat{z}))-c\|.
    \label{eq:posterior}
\end{align}
Please refer to supplementary material for more details.
Then we find that $\mathcal{X}(c) = x^{(c)}\in \mathbb{R}^{ H\times W\times 3}$ can be regarded as a kind of RGB image, where each category is represented by a specific color.
For convenience, we term it \texttt{maskige}. Therefore, the optimization can be rewritten as
\begin{align}
    \notag&\quad \min_{\hat{\phi},\hat{\theta}} \mathbb{E}_{q_{\hat{\phi}}(\hat{z}|x^{c})}\|\mathcal{D}_{\hat{\theta}}(\hat{z})-x^{(c)}\|\\
    \label{equ:opt_stage1_1}&+\min_{\fxpi}\mathbb{E}_{q_{\hat{\phi}}(\hat{z}|\mathcal{X}(c))}\|\mathcal{X}^{-1}(\hat{x}^{(c)})-c\|,
\end{align}
where $\hat{x}^{(c)}=\mathcal{D}_{\hat{\theta}}(\hat{z})$.
Now, the first term of Eq.~\eqref{equ:opt_stage1_1} can be regarded as an image reconstruction task (see Figure~\ref{fig:stage1}).
In practice, this has been already well optimized by \cite{esser2021taming,ramesh2021zero} using million-scale datasets.
This allows us to directly utilize 
the off-the-shelf pretrained models.
As such, our optimization problem can be simplified as:
\begin{equation}
\label{equ:opt_stage1_2}
    \min_{\fxpi}\mathbb{E}_{q_{\hat{\phi}}(\hat{z}|\mathcal{X}(c))}\|\mathcal{X}^{-1}(\hat{x}^{(c)})-c\|.
\end{equation}
Note that the parameters (0.9K$\sim$466.7K in our designs) of $\mathcal{X}$ and $\mathcal{X}^{-1}$ are far less than $\theta, \phi$ (totally 29.1M parameters with the VQVAE from~\cite{ramesh2021zero}), thus more efficient and cheaper to train.
Concretely, we only optimize the small $\mathcal{X}$ while
freezing $\theta,\phi$. 
Following \cite{kolesnikov2022uvim, chen2022generalist}, we use cross-entropy loss instead of MSE loss in Eq.~\eqref{equ:opt_stage1_2} for a better minimization between segmentation masks.

\paragraph{Linear maskige designs}
The optimization problem for $\fx$ and $\fxpi$ is non-convex, making their joint optimization challenging.
To overcome this issue, we optimize $\fx$ and $\fxpi$ separately.
For simplicity, we model both $\fx$ and $\fxpi$ as linear functions ({\em linear assumption}).
Specifically, we set $x^{(c)} = c\beta$ where $\beta\in \mathbb{R}^{K\times 3}$, and $\hat{c} = \hat{x}^{(c)}\beta^\dag$ where $\beta^\dag\in \mathbb{R}^{3\times K}$.
Under the {linear assumption}, we can transfer Eq.~\eqref{equ:opt_stage1_2} into a least squares problem with an explicit solution $\beta^\dag = \beta^\top(\beta\beta^\top)^{-1}$, so that $\mathcal{X}^{-1}$ is free of training.

To enable zero-cost training of $\fx$, we can also manually set the value of $\beta$ properly.
We suggest a {\bf \em maximal distance assumption} for selecting the value of $\beta$ to encourage the encoding of $K$ categories to be as widely dispersed as possible in the three-dimensional Euclidean space $\mathbb{R}^{3}$. More details are provided in the supplementary material.

\paragraph{Non-linear maskige designs}
For more generic design, 
non-linear models (\eg\ CNNs or Transformers) can be also used 
to express $\fxpi$. ({\em non-linear assumption})

\paragraph{Concrete maskige designs}
We implement four optimization settings for $\fx$ and/or $\fxpi$
with varying training budgets.
We define the naming convention of ``\model{}-[F/T]~[F/T]~(-$O$)'' in the following rules. 
{\em \textbf{(i)} Basic settings} ``-[F/T]~[F/T]'' on whether $\fx$ and/or $\fxpi$ require training: ``F" stands for {\em{}F}{ree of training}, and ``T'' for {\em{}T}{raining required}.
{\em \textbf{(ii)} Optional settings} ``-O'' (\eg\ ``R" or ``W") which will be explained later on. All \model{} variants are described below.

\noindent \textbullet~\textbf{\model{}-\gssb{}} (training free): Modeling both $\fx$ and $\fxpi$ using linear functions, with $\beta$ initialized under {\em maximal distance assumption}, and $\beta^\dag$ optimized using least squares. 
For comparison, we will experiment with \textbf{\model-\gssa{}}, where $\beta$ is \textbf{R}andomly initialized. 

\noindent \textbullet~\textbf{\model{}-\gssc{}} (training required): Modeling $\fx$ using a linear function but modeling $\fxpi$ using a non-linear function (\eg\ a three-layer convolutional neural network). We initialize $\beta$ with {\em maximal distance assumption} and optimize $\fxpi$ with gradient descent. 
A stronger design \textbf{\model{}-\gssc{}-W} utilizes a single-layer Shifted \textbf{W}indow Transformer block~\cite{liu2021swin} as the non-linear $\fxpi$. Notably, we train $\fx$ and $\fxpi$ separately, as described in Section~\ref{sec:set_up}. 

\noindent \textbullet~\textbf{\model{}-\gssd{}} (training required): Modeling both $\fx$ and $\fxpi$ using linear functions. We train $\beta$ with gradient descent and optimize $\beta^\dag$ with least squares according to $\beta$. 

\noindent \textbullet~\textbf{\model{}-\gsse{}} (training required): Modeling $\fx$ using a linear function but modeling $\fxpi$ using a non-linear function (\eg\ a three-layer CNN). We jointly train both functions using gradient descent. 

To perform end-to-end optimization of the $\fx$ function using gradient descent for both \model{}-\gssd{}\&\gsse{}, 
a hard Gumbel-softmax relaxation technique~\cite{maddison2016concrete} is used.
This involves computing the \texttt{argmax} operation during the forward step, while broadcasting the gradients during the backward step.
Our {\em linear} designs (\ie\ \model{}-\gssb{}\&\gssa{})
is training free with zero cost. 
Our {non-linear assumption} based designs (\eg\ \model{}-\gssc{}\&\gssc{}-W) has high performance potential at acceptable training cost (see Section~\ref{sec:ablation}).

\subsection{Stage II: Latent prior learning}
We show in Figure~\ref{fig:stage2} {(\bf \em Top)} the latent prior learning.
In this stage, we learn the prior joint distribution between mask latent representation $z$ and images $x$, with $\phi, \theta$ both fixed.

\paragraph{Objective} The optimization target of this stage is the second term of Eq.~\eqref{equ:elbo_sem}:
\begin{equation*}
    \min_\psi D_{KL}\Big(q_{\phi}(z|c), p_{\psi}(z|x)\Big),
\end{equation*}
where $z$ is in a discrete space of codebook-sized (\eg\ 8192 in \cite{ramesh2021zero}) integers.
The objective is to minimize the distance between the discrete distribution of $z$ predicted by latent prior encoder $p_\psi$ and the $z$ given by VQVAE.
Since the entropy of $q_\phi$ is fixed (\ie the ground truth), 
we can use the cross-entropy function to measure their alignment.

\paragraph{Unlabeled area auxiliary}
Due to high labeling cost and challenge, it is often the case that a fraction of areas per image are unlabeled (\ie\ unknown/missing labels).
Modeling per-pixel conditional probability $p(c|x)$ in existing discriminative models,
this issue can be simply tackled by ignoring all unlabeled pixels during training.

\begin{figure*}[tb]
	\def \imwidth {4.2cm}
	\def \adeheight {2.0cm}
	\def \cityheight {2.3cm}
	\centering
	\includegraphics[height=\cityheight, width=\imwidth, align=b]{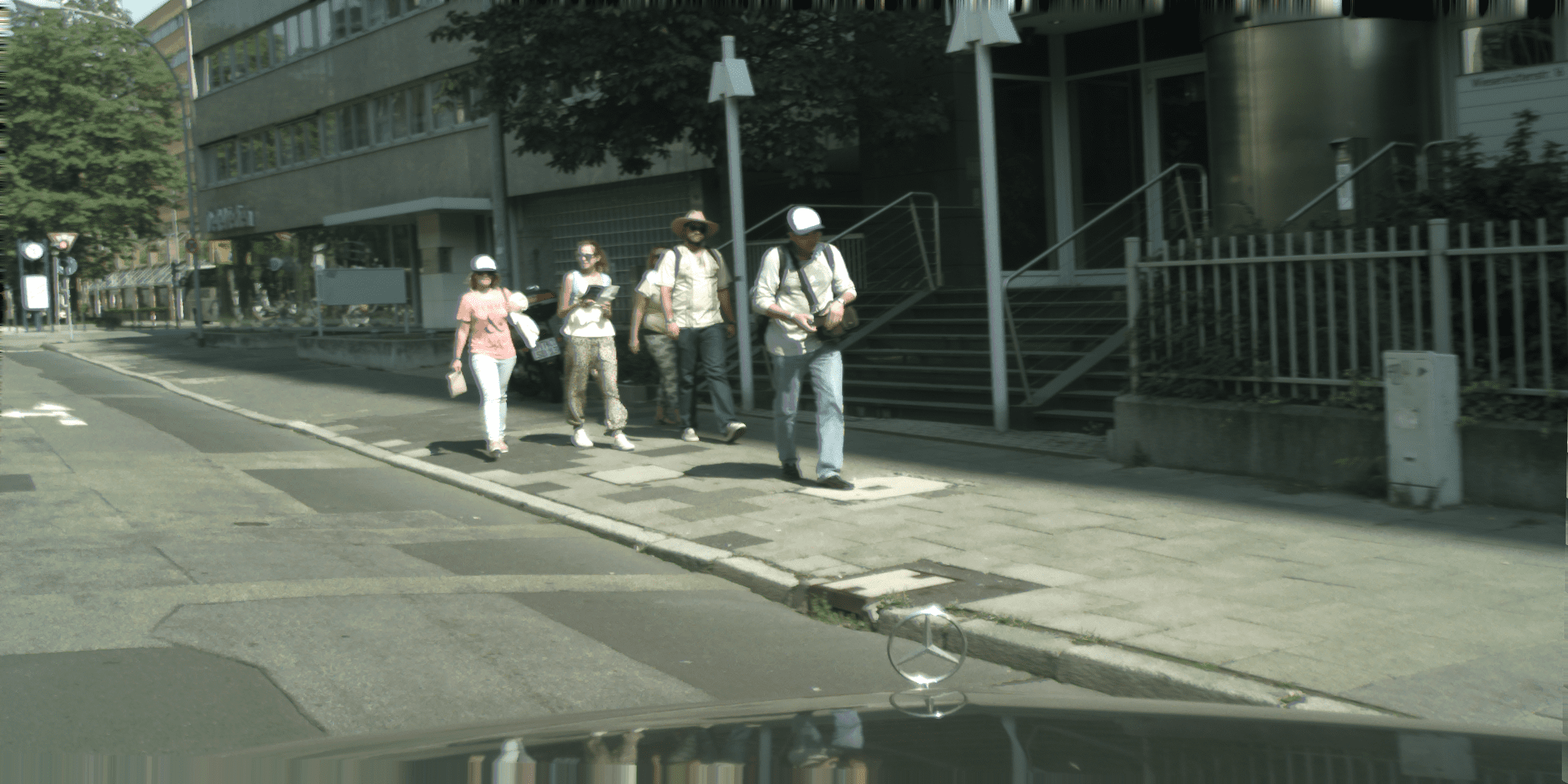}
	\includegraphics[height=\cityheight, width=\imwidth, align=b]{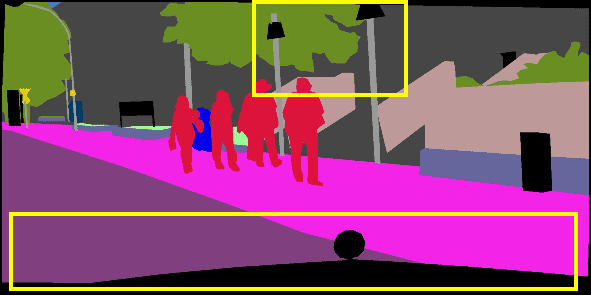}
	\includegraphics[height=\cityheight, width=\imwidth, align=b]{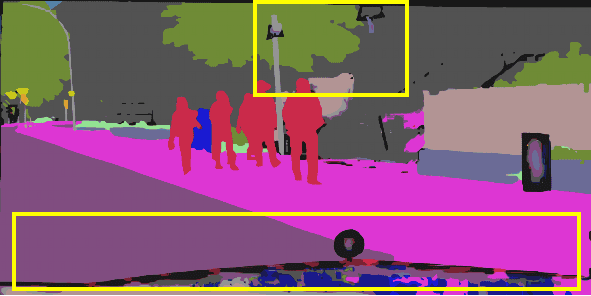}
	\includegraphics[height=\cityheight, width=\imwidth, align=b]{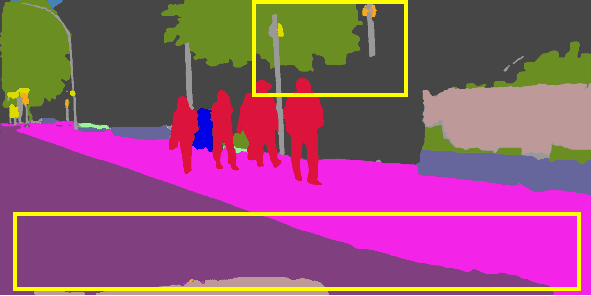}
	\\
    \rotatebox{0}{\textcolor{white}{--------------}Image}
	\rotatebox{0}{\textcolor{white}{------------------------}Ground Truth}
	\rotatebox{0}{\textcolor{white}{------------------}without auxiliary}
	\rotatebox{0}{\textcolor{white}{-----------------}with auxiliary\textcolor{white}{---------}}
\vspace{-1em}
	\caption{Qualitative results of {\em unlabeled area auxiliary} on Cityscapes~\cite{cordts2016cityscapes} dataset.
	}
	\vspace{-1em}
	\label{fig:unlabel_area_auxiliary}
\end{figure*}

In contrast, generative models (\eg\ UViM~\cite{kolesnikov2022uvim} and our \model{}) are trained at the latent token level, without flexible access to individual pixels. 
As a result, unlabeled pixels bring about extra challenges,
as they can be of objects/stuff of any categories heterogeneously.
Without proper handling, a generative model may learn to classify difficult pixels as the unlabelled and hurting the final performance (see Figure~\ref{fig:unlabel_area_auxiliary}).

To address this problem, we exploit a pseudo labeling strategy. 
The idea is to predict a label for each unlabeled pixel.
Specifically, we further introduce an auxiliary head $p_\xi(\bar{c}|z)$ during latent prior learning (\ie\ state II)
to label all unlabeled areas.
Formally, we form an enhanced ground-truth mask
by $\tilde{c}=M_u\cdot \bar{c} + (1-M_u)\cdot c$ where $M_u$ masks out labeled pixels, $\bar{c}$ denotes the pseudo labels, and $\tilde{c}$ denotes the labels after composition.
The final training objective of this stage cane be then revised as:
\begin{equation}
\label{equ:opt_unlabel}
    \min_\psi D_{KL}\left(q_{\phi}(z|\tilde{c}), p_{\psi}(z|x)\right) + p_\xi(\bar{c}|z).
\end{equation}

\subsection{Generative inference}
As illustrated in Figure~\ref{fig:stage2} {\bf\em(bottom)}, we first take the latent tokens $z$ that are predicted by the image encoder $\mathcal{I}_\psi$, and feed them into the {\tt maskige} decoder $\mathcal{D}_\theta$ to generate the predicted \texttt{maskige} $\hat{x}^{(c)}$. Next, we apply the inverse transformation $\mathcal{X}^{-1}$ (Section~\ref{sec:maskige}) to the predicted \texttt{maskige} to obtain the final segmentation mask $\hat{c}$.

\section{Experiment}
\subsection{Experimental setup}
\label{sec:set_up}
\paragraph{Cityscapes~\cite{cordts2016cityscapes}}
provides pixel-level annotations for 19 object categories in urban scene images at a high resolution of $2048 \times 1024$.
It contains 5000 finely annotated images, split into 2975, 500 and 1525 images for training, validation and testing respectively.
\paragraph{ADE20K~\cite{zhou2019semantic}}
is a challenging benchmark for scene parsing with 150 fine-grained semantic categories.
It has 20210, 2000 and 3352 images for training, validation and testing.

\paragraph{MSeg~\cite{lambert2020mseg}} is a composite dataset that unifies multiple semantic segmentation datasets from different domains. In particular, the taxonomy and pixel-level annotations are aligned by relabeling more than 220,000 object masks in over 80,000 images. We follow the standard setting: the train split~\cite{lin2014microsoft, caesar2018coco, chen2014semantic, cordts2016cityscapes, neuhold2017mapillary, varma2019idd, yu2020bdd100k, song2015sun} for training a unified semantic segmentation model,
the test split (unseen to model training)~\cite{everingham2010pascal, mottaghi2014role,brostow2009semantic, geiger2013vision, dai2017scannet, zendel2018wilddash} for cross-domain validation.

\paragraph*{Evaluation metrics}
The mean Intersection over Union (mIoU) and pixel-level accuracy (mAcc) are reported for all categories, following the standard evaluation protocol~\cite{cordts2016cityscapes}.

\paragraph{Implementation details}
We operate all experiments on {\em mmsegmentation} \cite{contributors2020openmmlab} with 8 NVIDIA A6000 cores. \emph{(i)~Data augmentation}: 
Images are resized to $1024 \times 2048$ on Cityscapes, $512 \times 2048$ on ADE20K and MSeg, and random cropped ($768 \times 768$ on cityscapes and $512\times512$ on ADE20K and MSeg) and random horizontal flipped during training.
\textbf{No test time augmentation} is applied. \emph{(ii)~Training schedule for latent prior learning}: 
The batch size is 16 on Cityscapes and MSeg and 32 on ADE20K. The total number of iterations is 80,000 on Cityscapes and 160,000 on ADE20K and MSeg.

\begin{table}[tb]
\tablestyle{1.8pt}{1.05}
\renewcommand\tabcolsep{20pt}
\renewcommand\arraystretch{1.2}
\small
\begin{center}
 
\begin{tabular}{l|cc}
\hline

\hline

\hline

\hline
GSS variants & {mIoU} & Training time \\
\hline

\hline

GSS-\gssa{} & 62.83 & ~~~~~~\textbf{0}\\
\textbf{GSS-\gssb{}} & 84.31 & ~~~~~~\textbf{0}\\
GSS-\gssc{} & 86.10 & ~~~$\leq$20\\
GSS-\gssd{} & 84.37 & ~~~$\leq$5\\
GSS-\gsse{} & 36.11 & ~~~$\leq$5\\
\rowcolor[gray]{.9} 
\textbf{GSS-\gssc{}-W} & \textbf{87.73} & ~~~~~~~~$\leq$350\\
\hline

\hline

\hline

\hline
\end{tabular}
\end{center}
\vspace{-1.5em}
\caption{
\textbf{Ablation on the variants of latent posterior learning} on the {\tt val} set of ADE20K.
Metrics: The \texttt{maskige} reconstruction performance in mIoU,
as well as the training time in \texttt{GPU hours} (\ie, effective single-core hours).
}
\vspace{-0.5em}
\label{tab:vqvae}
\end{table}
\begin{table}[tb]
\tablestyle{1.8pt}{1.05}
\renewcommand\tabcolsep{3.2pt}
\renewcommand\arraystretch{1.2}
\small
\begin{center}
 
\begin{tabular}{lc|ccl}
\hline

\hline

\hline

\hline
{Design} & {\texttt{Maskige}?} & {Cityscapes} & {ADE20K} & Train time \\
\hline

\hline
VQGAN~\cite{esser2021taming}& \XSolidBrush & 82.16 & 81.89 & $\leq$500\\
VQGAN~\cite{esser2021taming} &\checkmark & 75.09 & 42.70 & $\leq$100\\
UViM~\cite{kolesnikov2022uvim}\ &  \XSolidBrush & 89.14 & 78.98 & $\leq$2,000 \\
\rowcolor[gray]{.9} 
DALL$\cdot$E~\cite{ramesh2021zero} &\checkmark & \textbf{95.17} & \textbf{87.73} & $\leq$350\\

\hline

\hline

\hline

\hline
\end{tabular}
\end{center}
\vspace{-1.5em}
\caption{
\textbf{Ablation on \texttt{maskige} reconstruction by different VQVAE designs} 
on Cityscapes semantic {\tt val} split and ADE20k {\tt val} split.
In  case of \textit{no \texttt{maskige}}, we directly reconstruct the segmentation mask with $K$ (the number of classes) channels.
Unit for \textit{training time} is \texttt{GPU hour} (\ie\ the effective single-core hour).
}
\vspace{-0.5em}
\label{tab:vqvae_struct}
\end{table}
\begin{table}[tb]
\tablestyle{1.8pt}{1.05}

\renewcommand\tabcolsep{12.3pt}
\renewcommand\arraystretch{1.2}
\small
\begin{center}
 
\begin{tabular}{lcc|cc}
\hline

\hline

\hline

\hline
$d$&Unlabel& MLA &  ~{mIoU}~ & ~{mAcc}~ \\
\hline

\hline

1/8 &&&  {40.64} & {52.55} \\ 
1/8 &\checkmark&&  {43.72} & {56.08}\\ 
1/4&\checkmark& &   {43.98} & {56.11} \\ 
\rowcolor[gray]{.9}
1/4&\checkmark& \checkmark &  \textbf{46.29} & \textbf{57.84} \\ 

\hline

\hline

\hline

\hline
\end{tabular}
\end{center}
\vspace{-1.5em}
\caption{
\textbf{Ablation on latent prior learning} 
on the {\tt val} split of ADE20K. ``Unlabel'' denotes unlabeled area auxiliary, and ``MLA" denotes Multi-Level Aggregation. ``$d$'' is the downsample ratio of discrete mask representation size between input image size.
}
\vspace{-1.5em}
\label{tab:transformer}
\end{table}
\begin{table}[tb]
\tablestyle{1.8pt}{1.1}
\renewcommand\tabcolsep{7.1pt}
\renewcommand\arraystretch{1.2}
\begin{tabular}{x{44}x{14}x{44}x{30}|c}
\hline

\hline

\hline

\hline
Method & Pretrain & Backbone & Iteration&mIoU \\
\shline
\multicolumn{4}{l}{\emph{- Discriminative modeling:}} \\
\shline
\multicolumn{1}{l}{FCN~\cite{long2015fully}} & 1K & ResNet-101 & {80k} &{77.02}\\ 
\multicolumn{1}{l}{PSPNet~\cite{zhao2017pyramid}} & 1K& ResNet-101 &{80k}  & {79.77}\\
\multicolumn{1}{l}{DeepLab-v3+~\cite{chen2018encoder}}& 1K & ResNet-101 &{80k} & {80.65}\\

\multicolumn{1}{l}{NonLocal~\cite{wang2018nonlocal}} & 1K& ResNet-101 &{80k} & {79.40}\\
\multicolumn{1}{l}{CCNet~\cite{huang2019ccnet}}& 1K & ResNet-101 &{80k} &{79.45} \\
\multicolumn{1}{l}{Maskformer~\cite{cheng2021per}} & 1K& ResNet-101 & {90k} & {78.50} \\ 
\multicolumn{1}{l}{Mask2former~\cite{cheng2022masked}} & 1K& ResNet-101 & {90k} & {80.10} \\ 
\multicolumn{1}{l}{SETR~\cite{zheng2021rethinking}} & 22K& ViT-Large & {80k} & {78.10} \\
\multicolumn{1}{l}{UperNet~\cite{xiao2018unified}}& 22K & Swin-Large & {80k}  & {82.89}\\ 
\multicolumn{1}{l}{Mask2former~\cite{cheng2022masked}}& 22K & Swin-Large & {90k}  & \textbf{83.30}\\
\multicolumn{1}{l}{SegFormer~\cite{xie2021segformer}} & 1K& MiT-B5 & {160k} & {82.25} \\
\shline
\multicolumn{3}{l}{\emph{- Generative modeling:}} \\
\shline
\multicolumn{1}{l}{UViM$^\dag$~\cite{kolesnikov2022uvim}}& 22K & Swin-Large & 160k &70.77 \\
\rowcolor[gray]{.9}
\multicolumn{1}{l}{\model-\gssb{}~(Ours)} & 1K & ResNet-101 & 80k & {77.76} \\
\rowcolor[gray]{.9}
\multicolumn{1}{l}{\model-\gssc{}-W~(Ours)}& 1K & ResNet-101 & 80k & {78.46} \\
\rowcolor[gray]{.9}
\multicolumn{1}{l}{\model-\gssb{}~(Ours)} & 22K& Swin-Large & 80k & {78.90} \\
\rowcolor[gray]{.9}
\multicolumn{1}{l}{\model-\gssc{}-W~(Ours)}& 22K & Swin-Large & 80k & \textbf{80.05}\\

\hline

\hline

\hline

\hline
\end{tabular}
\vspace{-0.5em}
\caption{\textbf{Performance comparison on the Cityscapes {\tt val} split:} UViM$^\dag$~\cite{kolesnikov2022uvim} is reproduced by us on PyTorch. ``1K" means pretrained on ImageNet 1K~\cite{deng2009imagenet} while ``22K" means pretrained on ImageNet 22K~\cite{deng2009imagenet}.}
\label{tab:cityscapes_val}
\vspace{-0.5em}
\end{table}
\begin{table}[tb]
\tablestyle{1.8pt}{1.1}
\renewcommand\tabcolsep{7.0pt}
\renewcommand\arraystretch{1.2}
\begin{tabular}{x{44}x{14}x{44}x{30}|c}
\hline

\hline

\hline

\hline
Method & Pretrain & Backbone & Iteration &mIoU \\
\shline
\multicolumn{3}{l}{\emph{- Discriminative modeling:}}\\
\shline
\multicolumn{1}{l}{FCN~\cite{long2015fully}}&1K & ResNet-101 & {160k} &{41.40} \\
\multicolumn{1}{l}{CCNet~\cite{huang2019ccnet}}&1K & ResNet-101 & {160k}&{43.71} \\
\multicolumn{1}{l}{DANet~\cite{fu2019dual}}&1K & ResNet-101 &{160k} &  {44.17}\\
\multicolumn{1}{l}{UperNet~\cite{xiao2018unified}}&1K & ResNet-101 & {160k}&  {43.82}\\
\multicolumn{1}{l}{Deeplab-v3+~\cite{chen2018encoder}}&1K & ResNet-101 &{160k} & {45.47}\\
\multicolumn{1}{l}{Maskformer~\cite{cheng2021per}} & 1K& ResNet-101 & {160k}& {45.50}\\
\multicolumn{1}{l}{Mask2former~\cite{cheng2022masked}} & 1K& ResNet-101 & {160k}& {47.80}\\
\multicolumn{1}{l}{OCRNet~\cite{yuan2019object}}&1K & HRNet-W48 &{160k} &  {43.25}\\
\multicolumn{1}{l}{SegFormer~\cite{xie2021segformer}}&1K & MiT-B5 & {160k}& \textbf{50.08}\\
\multicolumn{1}{l}{SETR~\cite{zheng2021rethinking}}&22K & ViT-Large& {160k}& {48.28}\\
\shline
\multicolumn{3}{l}{\emph{- Generative modeling:}} \\
\shline
\multicolumn{1}{l}{UViM$^\dag$~\cite{kolesnikov2022uvim}} & 22k & Swin-Large & 160k &43.71 \\
\rowcolor[gray]{.9}
\multicolumn{1}{l}{\model-\gssb{}~(Ours)}&22K & Swin-Large & 160k &46.29 \\
\rowcolor[gray]{.9}
\multicolumn{1}{l}{\model-\gssc{}-W~(Ours)}&22K & Swin-Large & 160k &\textbf{48.54} \\

\hline

\hline

\hline

\hline
\end{tabular}
\vspace{-0.5em}
\caption{\textbf{Performance comparison with previous art methods on the ADE20K {\tt val} split}. UViM$^\dag$~\cite{kolesnikov2022uvim} is reproduced by ourselves. 
}
\vspace{-2em}
\label{tab:ade20k_val}
\end{table}
\begin{table*}[htb]
\tablestyle{3.8pt}{1.05}
\vspace{-2mm}
\centering
\renewcommand\tabcolsep{2.7pt}
\renewcommand\arraystretch{1.2}
\vspace{-2mm}
\small
\begin{tabular}{lllcccccc|c}
\hline

\hline

\hline

\hline
Method 
& Backbone
& Iteration 
&VOC~\cite{everingham2010pascal} & Context~\cite{mottaghi2014role} & CamVid~\cite{brostow2009semantic} & WildDash~\cite{zendel2018wilddash}   & KITTI~\cite{geiger2013vision}  & ScanNet~\cite{dai2017scannet} & \textit{h.\ mean} \\
\hline

\hline
\multicolumn{3}{l}{\emph{- Discriminative modeling:}}\\
\shline
CCSA~\cite{motiian2017unified} & HRNet-W48 &500k &48.9 & - & 52.4 & 36.0 & - & 27.0 & 39.7 \\
MGDA~\cite{sener2018multi} & HRNet-W48 & 500k &69.4 & - & 57.5 & 39.9 & - & 33.5 & 46.1 \\
MSeg~\cite{lambert2020mseg}  & HRNet-W48 &  500k & 70.7 & 42.7  & \textbf{83.3} & 62.0 & 67.0 & 48.2 & 59.2\\
MSeg$^\dag$~\cite{lambert2020mseg} & HRNet-W48 & 160k & 63.8& 39.6 & 73.9 & 60.9 & 65.1 & 43.5 & 54.9\\
 MSeg$^\dag$~\cite{lambert2020mseg}& Swin-Large &160k & 78.7 & 47.5 & 75.1 & \textbf{66.1} & \textbf{68.1} & 49.0 & 61.7 \\
\hline

\hline
 \multicolumn{3}{l}{\emph{- Generative modeling:}}\\
 \shline
  \rowcolor[gray]{.9}
 \model-\gssb{}~(Ours) & HRNet-W48 & 160k &64.1&37.1&72.3&59.3&62.0&40.6& 52.6\\
 \rowcolor[gray]{.9}
 \model-\gssc{}-W~(Ours) & HRNet-W48 & 160k & 65.2&  38.8& 75.2 & 62.5 & 66.2 &43.1& 55.2 \\
\rowcolor[gray]{.9}
\model-\gssb{}~(Ours) & Swin-Large & 160k & 78.7 & 45.8 & 74.2 & 61.8 & 65.4 & 46.9 & 59.5\\
\rowcolor[gray]{.9}
\model-\gssc{}-W~(Ours) & Swin-Large & 160k & \textbf{79.5} & \textbf{47.7} & 75.9 & 65.3 & 68.0 &  \textbf{49.7} & \textbf{61.9} \\
\hline

\hline

\hline

\hline
\end{tabular}
\vspace{-0.5em}
\caption{\textbf{Cross-domain semantic segmentation performance on MSeg dataset {\tt test} split.} ``\textit{h. mean}" is the harmonic mean~\cite{lambert2020mseg}.
MSeg$^\dag$~\cite{lambert2020mseg} is reproduced by us on MMSegmentation~\cite{contributors2020openmmlab}}
\label{tab:mseg}
\vspace{-1em}
\end{table*}
\begin{table}[tb]
\tablestyle{1.8pt}{1.05}
\renewcommand\tabcolsep{6.6pt}
\renewcommand\arraystretch{1.2}
\begin{tabular}{x{44}x{74}|cc}
\hline

\hline

\hline

\hline
Sharing $\mathcal{I}_\psi$ & Sharing {\tt maskige} & \model{}-\gssb & \model{}-\gssc{}-W \\
\shline
 &  &\textbf{78.9} &\textbf{80.5}\\
 & \checkmark & 78.0& 79.5\\
 \rowcolor[gray]{.9}
\checkmark & \checkmark & 76.6& 78.4\\

\hline

\hline

\hline

\hline
\end{tabular}
\vspace{-1em}
\caption{\textbf{Transferring the {\tt maskige} and image encoder $\mathcal{I}_\psi$ from MSeg to Cityscapes ({\tt val} split).} 
Metric: mIoU.
}
\vspace{-2em}
\label{tab:share_maskiage_between_cityscapes_and_mseg}
\end{table}
\subsection{Ablation studies}
\label{sec:ablation}
{\noindent\bf Latent posterior learning}
We evaluate the variants of latent posterior learning as described in Section \ref{sec:maskige}.
We observe from Table~\ref{tab:vqvae} that: 
\textbf{(i)} \model{}-\gssb{} comes with no extra training cost. 
Whilst UViM consumes nearly 2K GPU hours for training a VQVAE with similar performance achieved~\cite{kolesnikov2022uvim}.
\textbf{(ii)} The randomly initialized $\beta$ (\ie\ \model{}-\gssa{}) leads to considerable degradation on the least square optimization.
\textbf{(iii)} With our {\em maximal distance assumption}, the regularized $\beta$ of \model{}-\gssb{} brings clear improvement,
suggesting the significance of initialization and its efficacy of our strategy.
\textbf{(iv)} With three-layer conv network with activation function for $\fxpi$, \model{}-\gssc{} achieves good reconstruction.
Further equipping with a two-layer Shifted Window Transformer block~\cite{liu2021swin} for $\mathcal{X}^{-1}$ (\ie\ {\model{}-\gssc{}-W}) leads to the best result at a cost of extra 329.5 GPU hours.
This is due to more accurate translation from predicted {\tt maskige} to segmentation mask.
\textbf{(v)} Interestingly, with automatic $\beta$ optimization,
\model{}-\gssd{} brings no benefit over \model{}-\gssb{}.
\textbf{(vi)} 
Further, joint optimization of both $\fx$ and $\fxpi$ (\ie\ \model{}-\gsse{}) fails to achieve the best performance.
\textbf{(vii)} In conclusion, \model{}-\gssb{} is most efficient with reasonable accuracy,
whilst {\model{}-\gssc{}-W} is strongest with good efficiency.

{\noindent\bf VQVAE design}
We examine the effect of VQVAE in the context of \texttt{maskige}.
We compare three designs:
{\bf(1)} {\em UViM-style} \cite{kolesnikov2022uvim}: Using images as auxiliary input to reconstruct a segmentation mask in form of $K$-channels ($K$ is the class number) in a ViT architecture.
In this no {\tt maskige} case, the size of segmentation mask 
may vary across different datasets, leading to a need for dataset-specific training.
This scheme is thus more expensive in compute.
{\bf (2)} {\em VQGAN-style} \cite{esser2021taming}: 
Using a CNN model for reconstructing natural images ({\tt maskige} needed for segmentation mask reconstruction) or $K$-channel segmentation masks (no {\tt maskige} case) separately,
both optimized in 
generative adversarial training manner with a smaller codebook.
{\bf (3)} {\em DALL$\cdot$E-style} \cite{ramesh2021zero}:
The one we adopt, as discussed earlier.
We observe from Table~\ref{tab:vqvae_struct} that:
\textbf{(i)} 
Due to the need for dataset specific training,
UViM-style is indeed more costly than the others.
This issue can be well mitigated by our \texttt{maskige} with 
the first stage training cost compressed dramatically,
as evidenced by DALL$\cdot$E-style and VQGAN-style.
Further, the inferiority of UViM over DALL$\cdot$E
suggests that our \texttt{maskige} is a favored strategy 
than feeding image as auxiliary input.
\textbf{(ii)} 
In conclusion, using our \texttt{maskige} and DALL$\cdot$E pretrained VQVAE yields the best performance in terms of both accuracy and efficiency.

{\noindent\bf Latent prior learning}
We ablate the second training stage for learning latent joint prior.
The baseline is 
{\model{}-\gssb{}} without the unlabeled area auxiliary and Multi-Level Aggregation (MLA, including a 2-layer Swin block~\cite{liu2021swin}), under 1/8 downsample ratio.
We observe from Table~\ref{tab:transformer} that: 
\textbf{(i)} 
Our unlabeled area auxiliary boosts the accuracy by $3.1\%$,
 suggesting the importance of complete labeling which however is extremely costly
 in semantic segmentation.
\textbf{(ii)} 
Increasing the discrete mask representation resolution is slightly useful. 
\textbf{(iii)} 
The MLA plays another important role,
\eg\ giving a gain of $2.3\%$.

\begin{figure*}[htb]
	\def \imwidth {3.3cm}
	\def \imheight {2.3cm}
	\centering
	\includegraphics[height=\imheight, width=\imwidth, align=b]{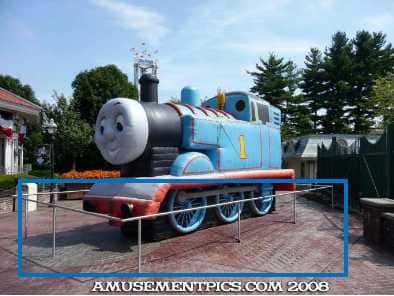}
	\includegraphics[height=\imheight, width=\imwidth, align=b]{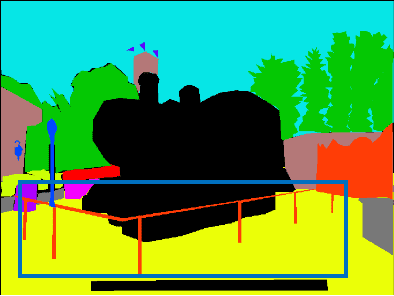}
	\includegraphics[height=\imheight, width=\imwidth, align=b]{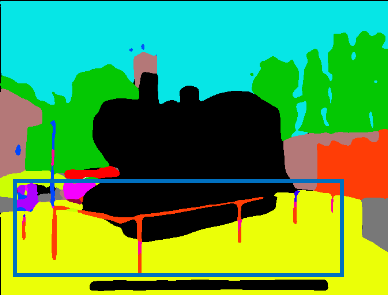}
	\includegraphics[height=\imheight, width=\imwidth, align=b]{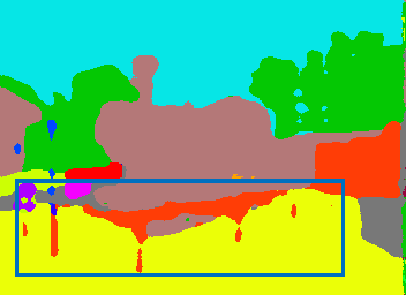}
	\includegraphics[height=\imheight, width=\imwidth, align=b]{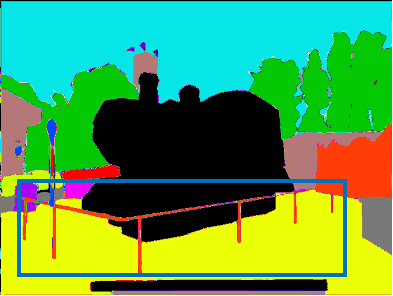}\\
 	\includegraphics[height=\imheight, width=\imwidth, align=b]{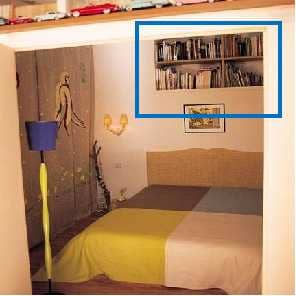}
	\includegraphics[height=\imheight, width=\imwidth, align=b]{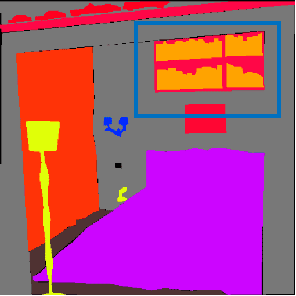}
	\includegraphics[height=\imheight, width=\imwidth, align=b]{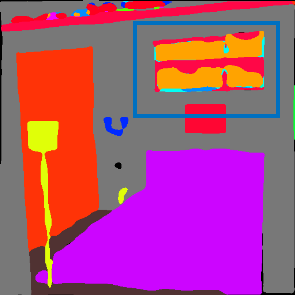}
	\includegraphics[height=\imheight, width=\imwidth, align=b]{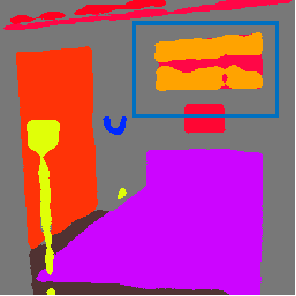}
	\includegraphics[height=\imheight, width=\imwidth, align=b]{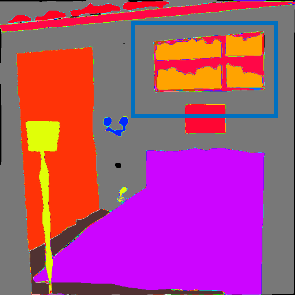} \\

	\rotatebox{0}{\textcolor{white}{---}Image}
	\rotatebox{0}{\textcolor{white}{----------------}Ground Truth}
	\rotatebox{0}{\textcolor{white}{--------------}VQGAN~\cite{esser2021taming}}
	\rotatebox{0}{\textcolor{white}{--------------}UViM$^\dag$~\cite{kolesnikov2022uvim}}
	\rotatebox{0}{\textcolor{white}{-------------}\model{} (ours)}
	\vspace{-0.5em}
	\caption{Qualitative results of {\texttt maskige} reconstruction on ADE20K~\cite{zhou2019semantic} dataset. Note that the black areas in the Ground Truth correspond to unlabeled regions, and thus {\em no impact} on mIoU measurement.
	}
	\vspace{-0.3em}
	\label{fig:visualization_stage_1}
\end{figure*}
\begin{figure*}[bth]
	\def \imwidth {2.8cm}
	\def \adeheight {2.0cm}
	\def \cityheight {1.5cm}
	\centering
	\includegraphics[height=\cityheight, width=\imwidth, align=b]{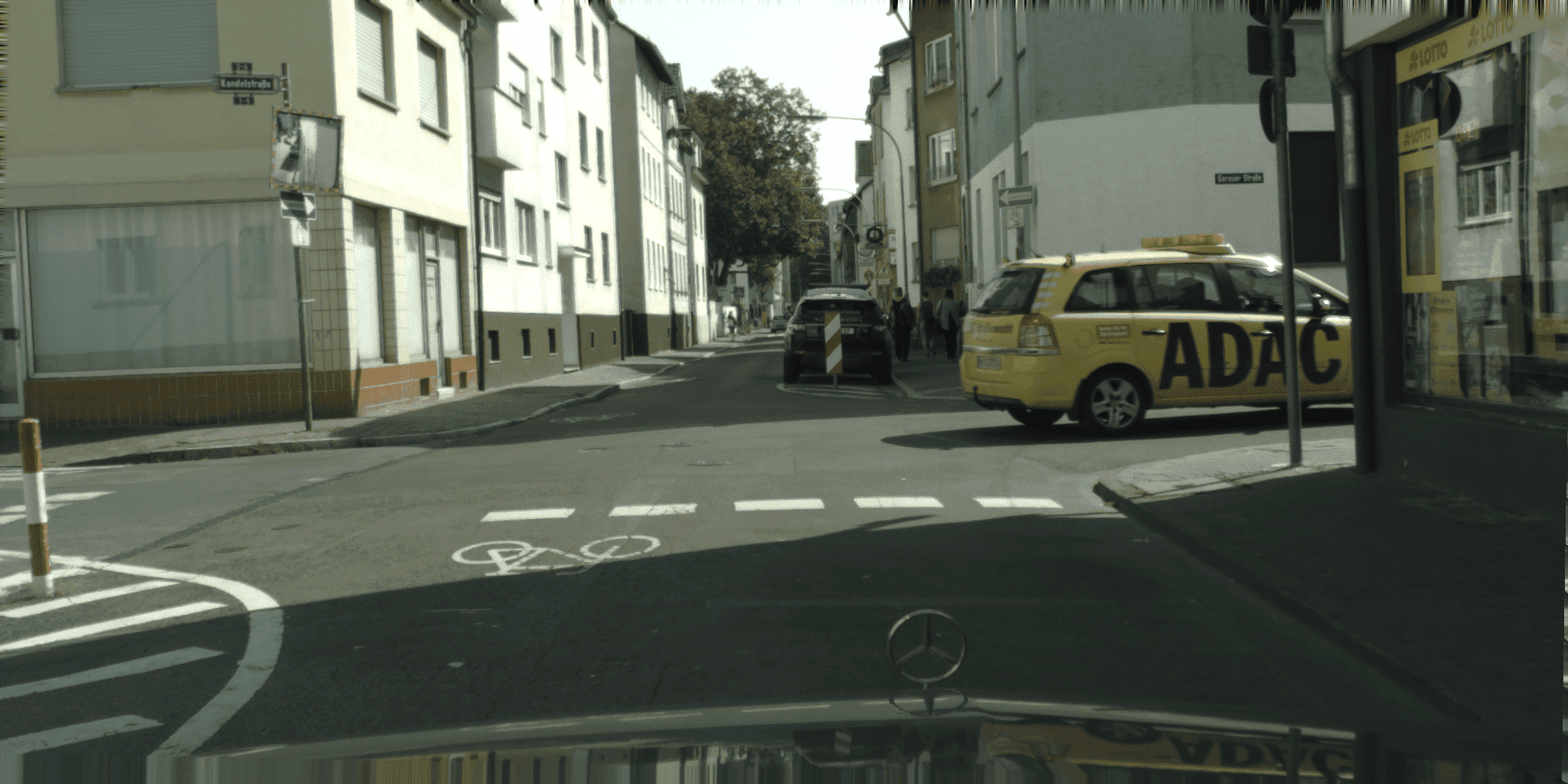}
	\includegraphics[height=\cityheight, width=\imwidth, align=b]{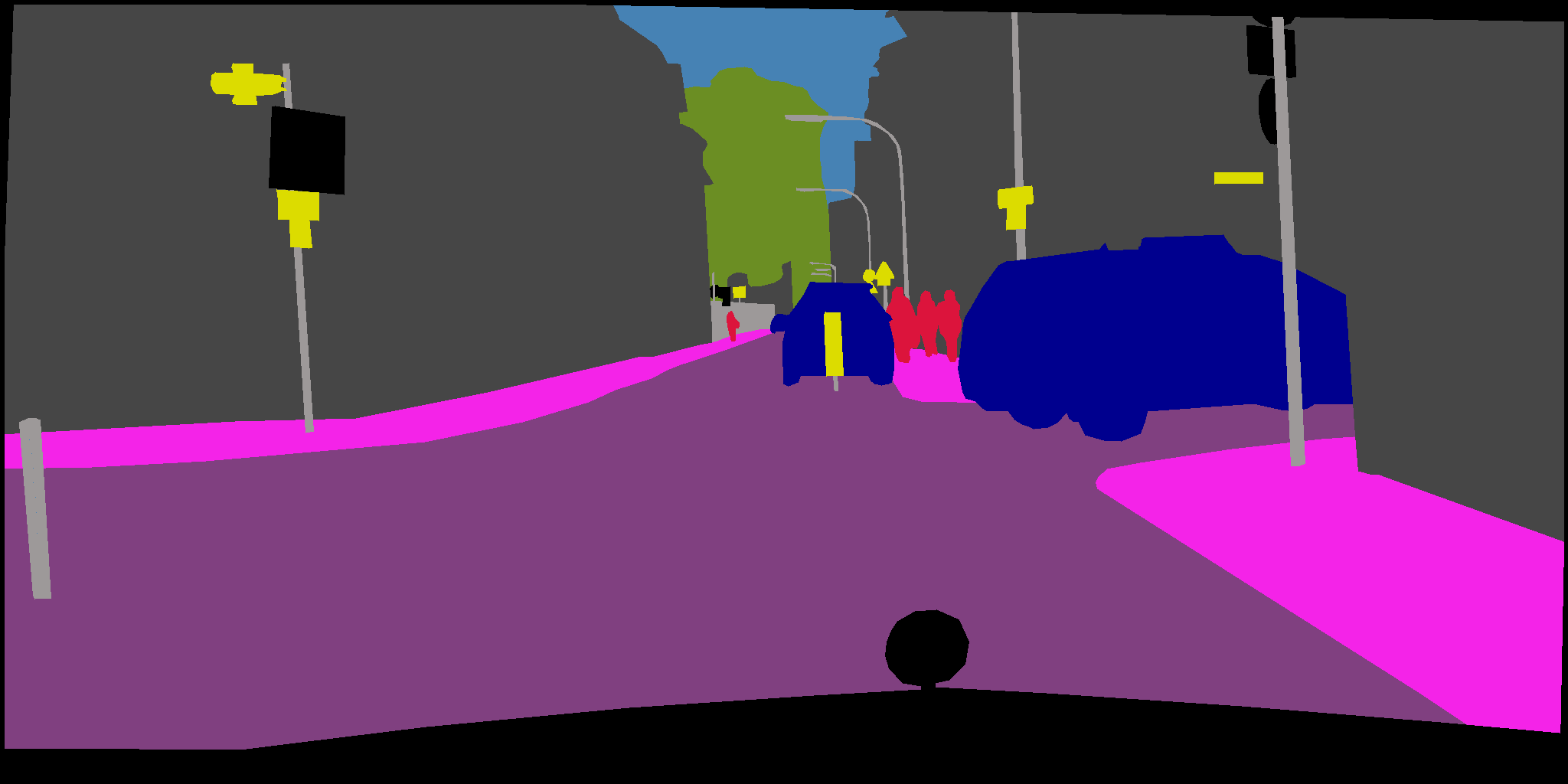}
	\includegraphics[height=\cityheight, width=\imwidth, align=b]{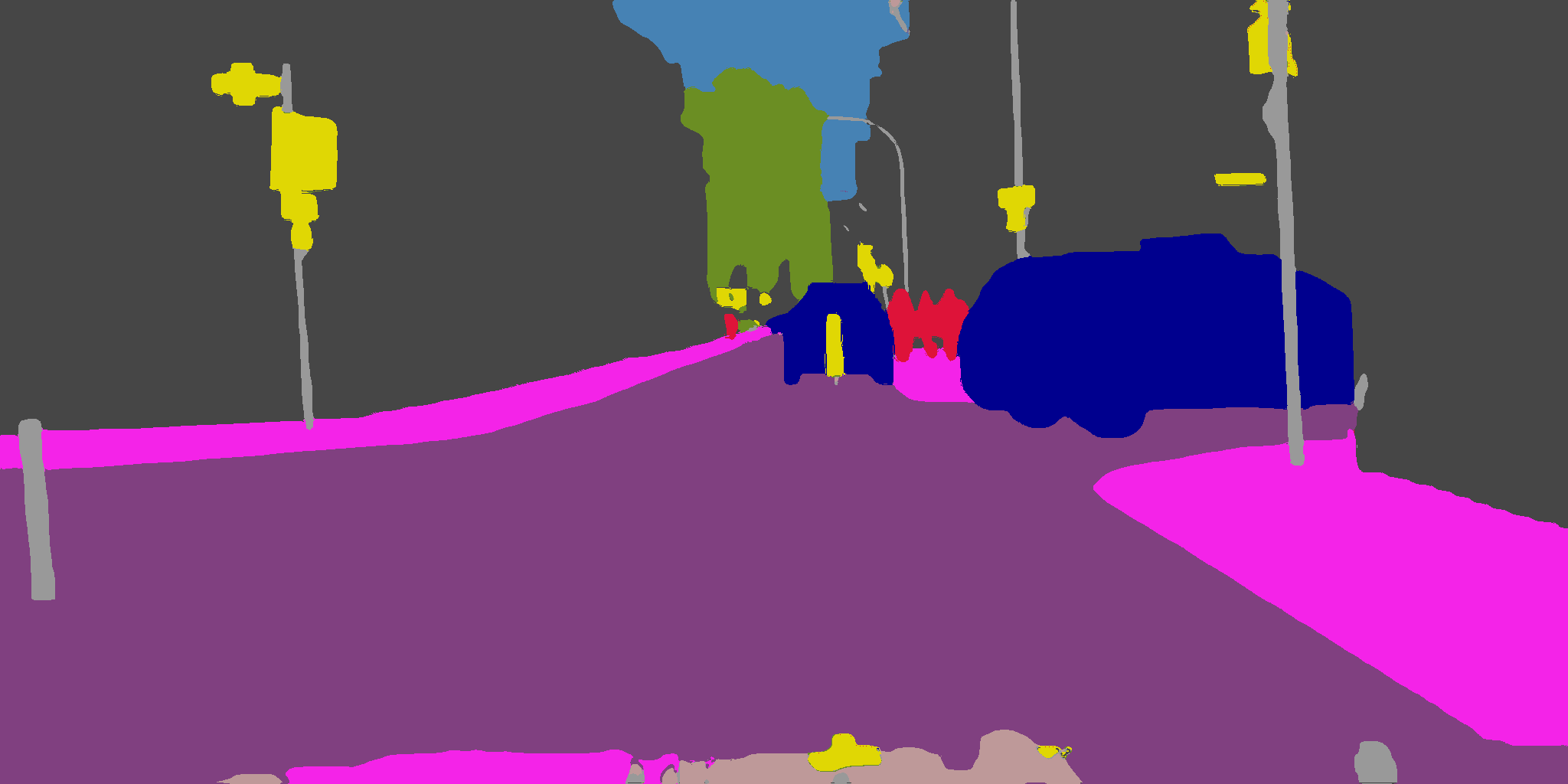}
	\includegraphics[height=\cityheight, width=\imwidth, align=b]{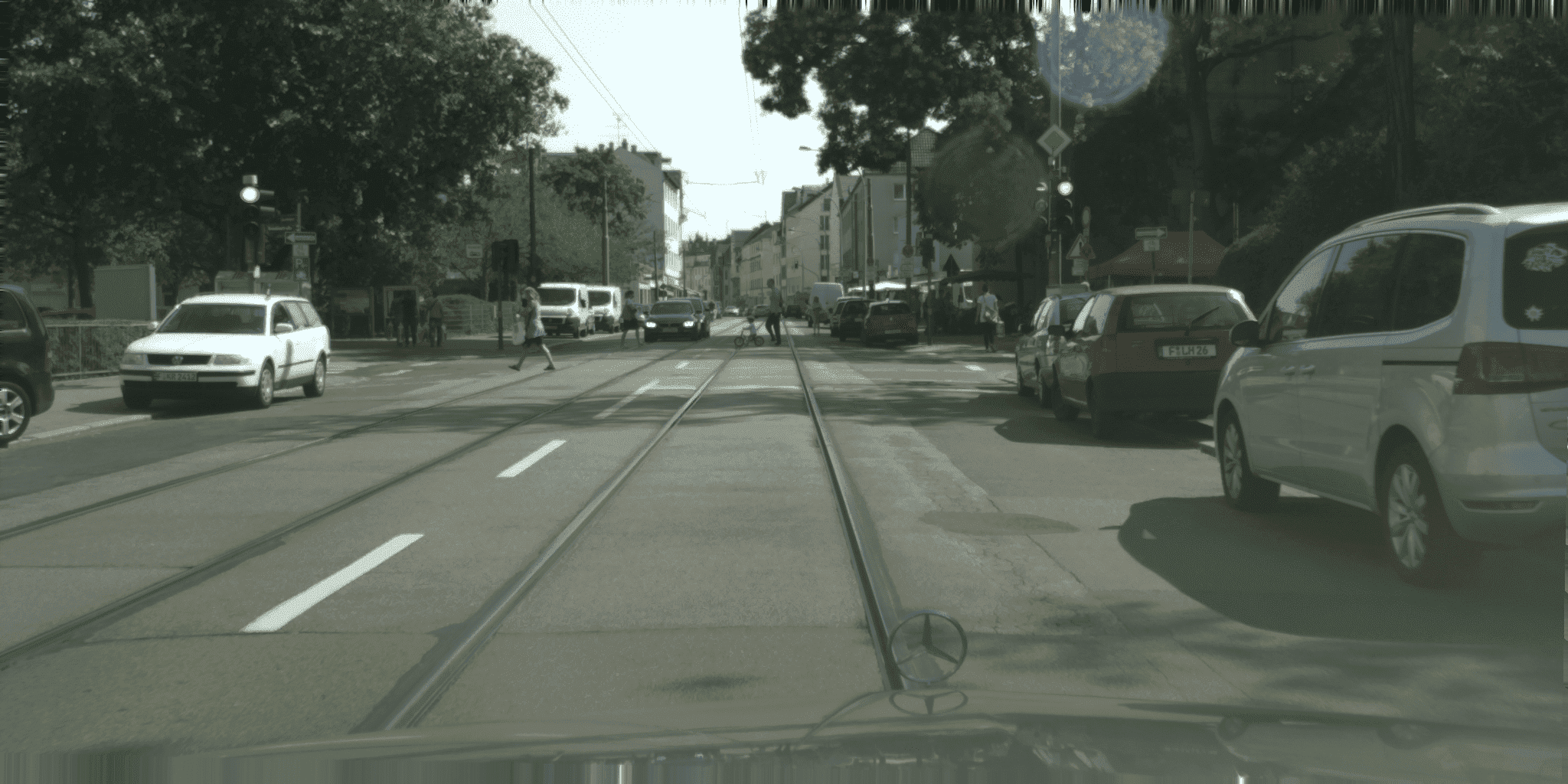}
	\includegraphics[height=\cityheight, width=\imwidth, align=b]{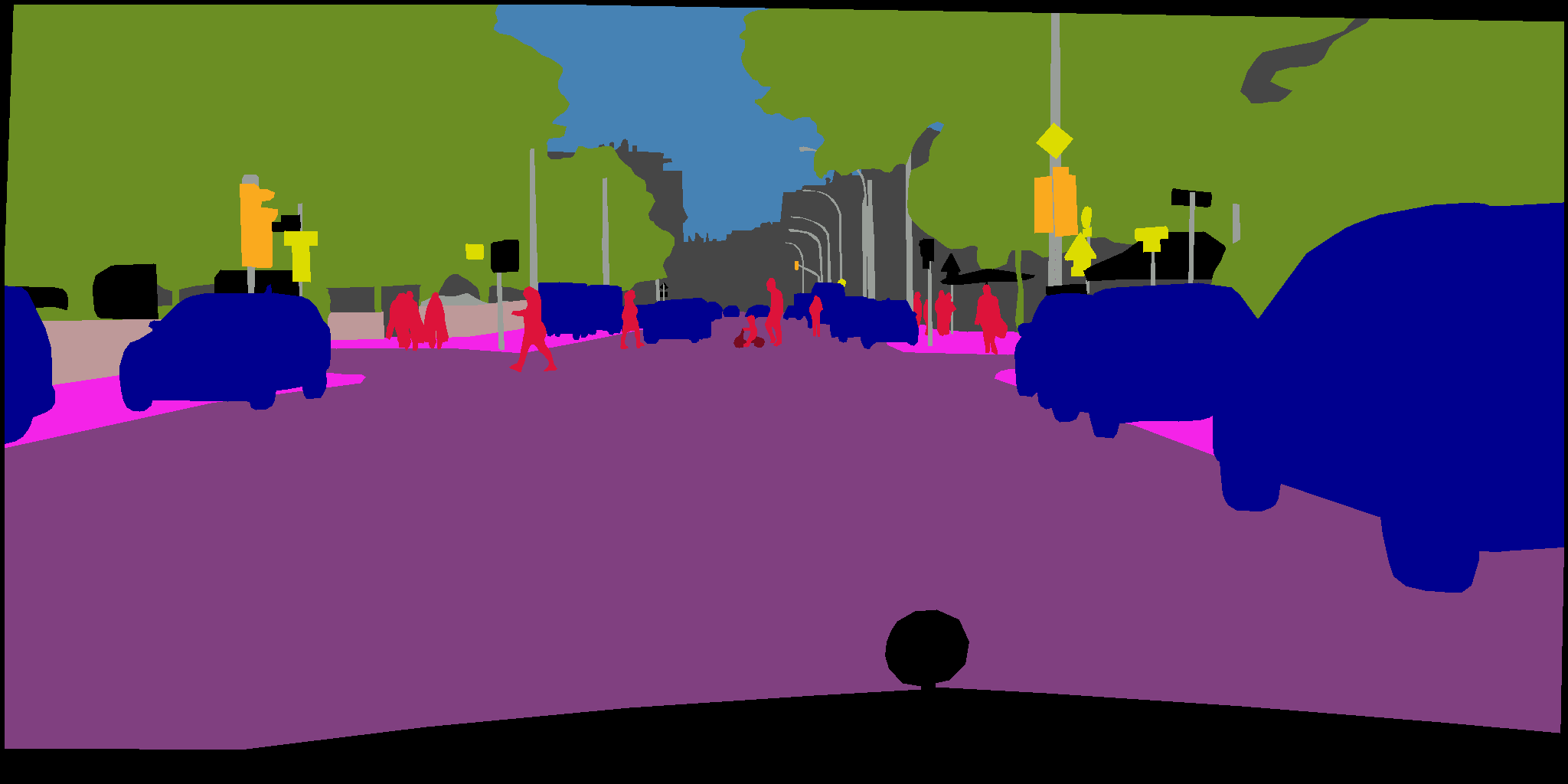}
	\includegraphics[height=\cityheight, width=\imwidth, align=b]{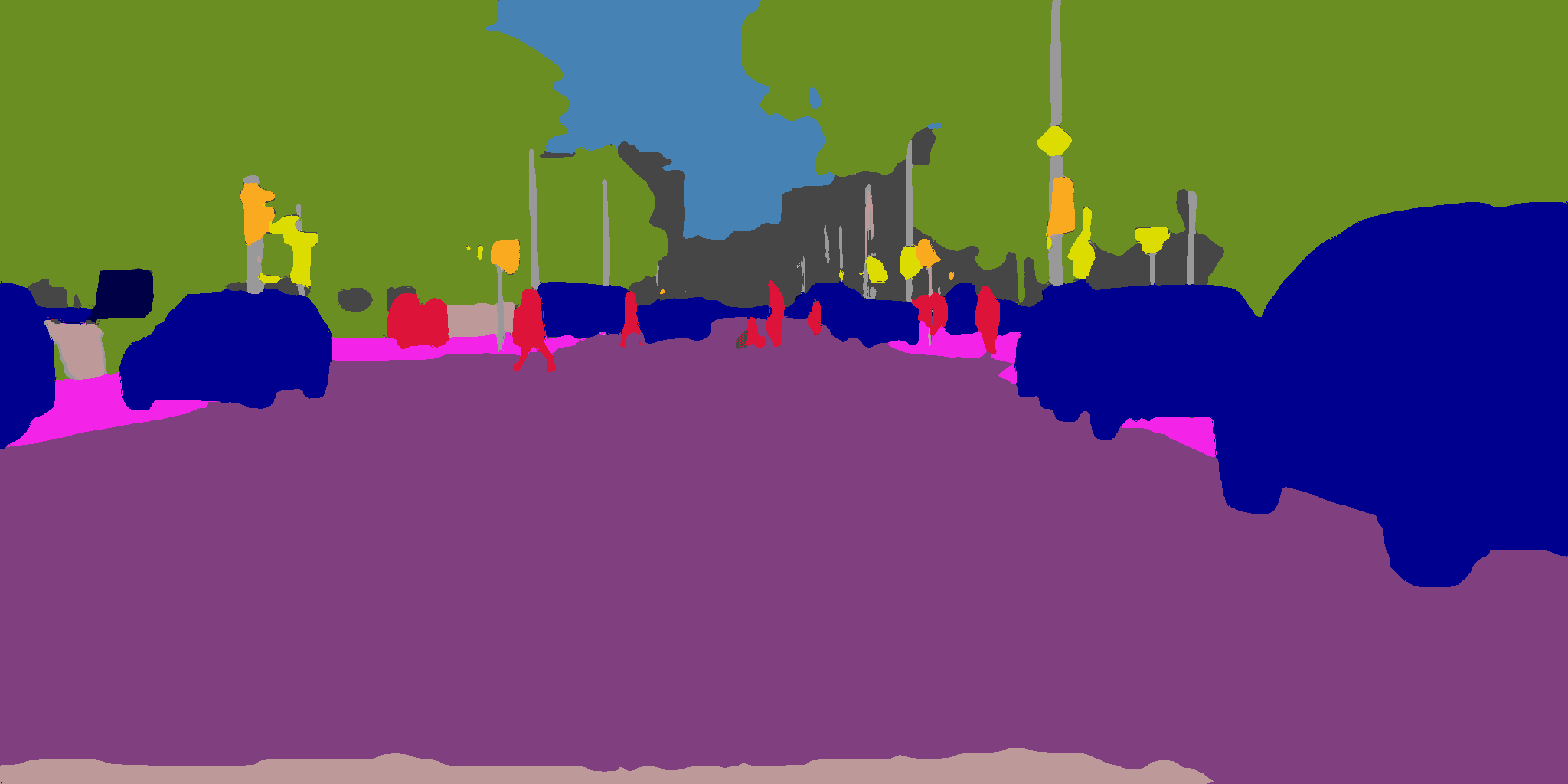}
	\\
	\includegraphics[height=\adeheight, width=\imwidth, align=b]{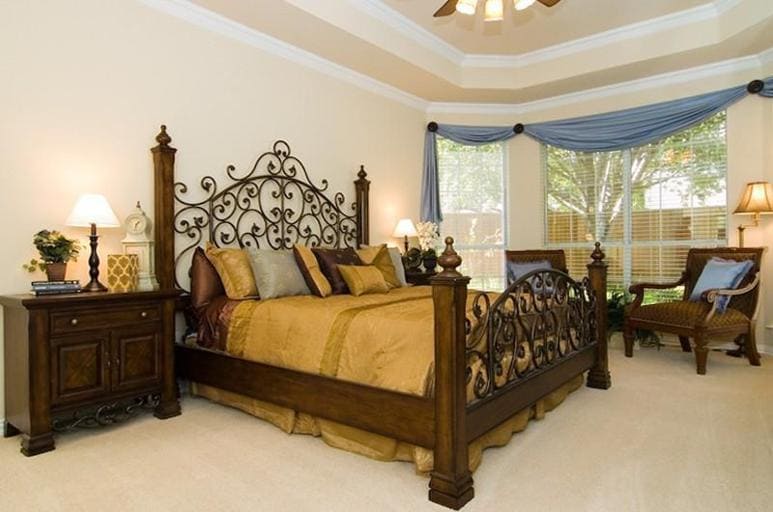}
	\includegraphics[height=\adeheight, width=\imwidth, align=b]{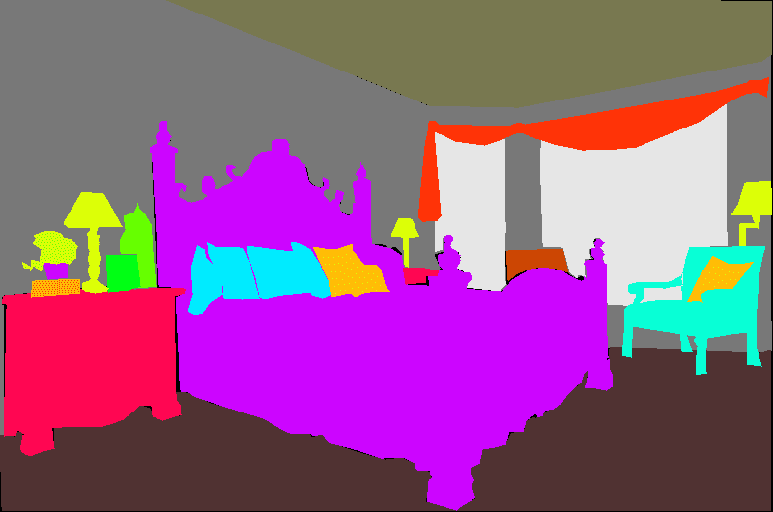}
	\includegraphics[height=\adeheight, width=\imwidth, align=b]{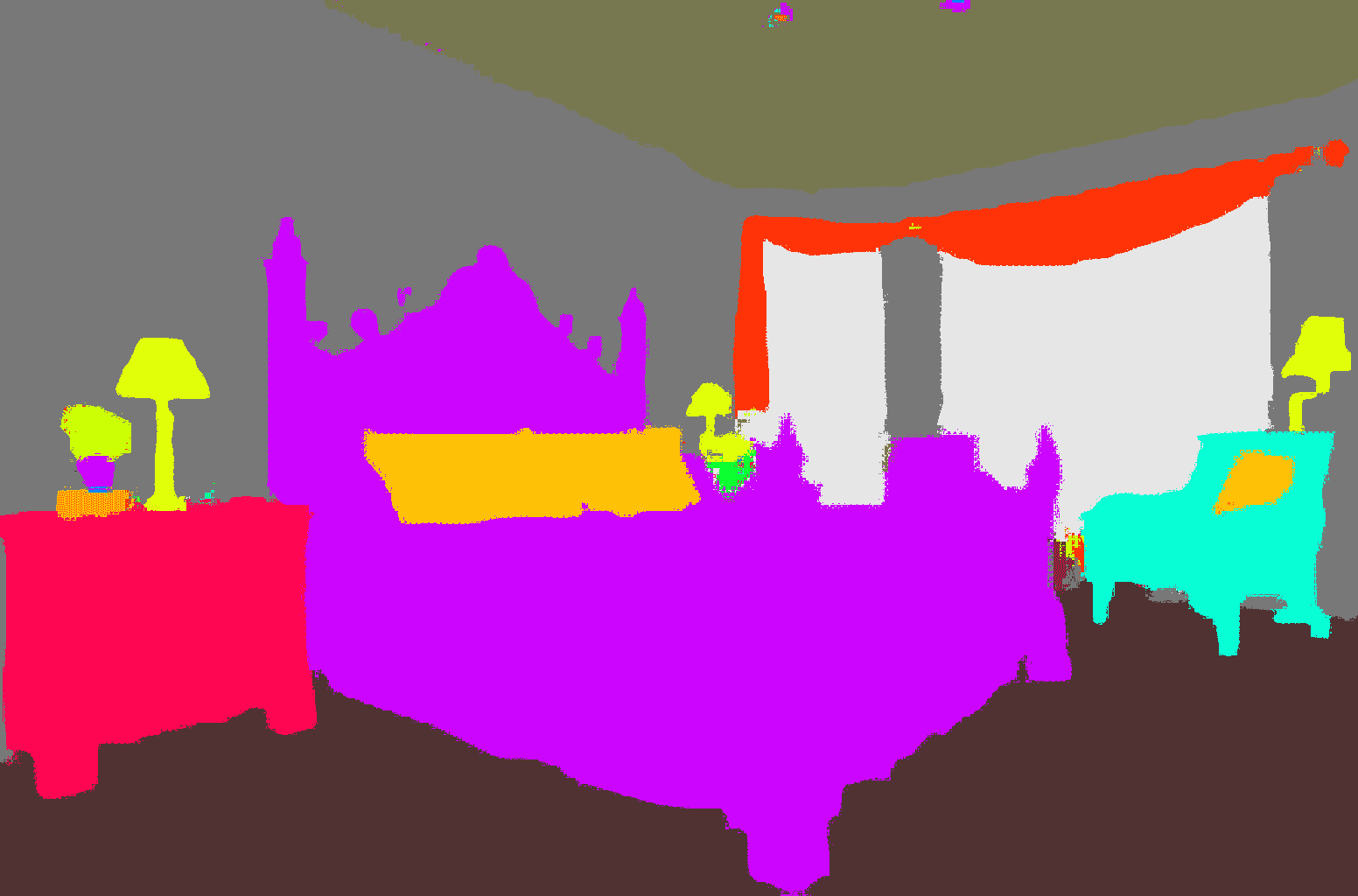}
	\includegraphics[height=\adeheight, width=\imwidth, align=b]{figures/stage2/image/ADE_val_00000118.jpg}
	\includegraphics[height=\adeheight, width=\imwidth, align=b]{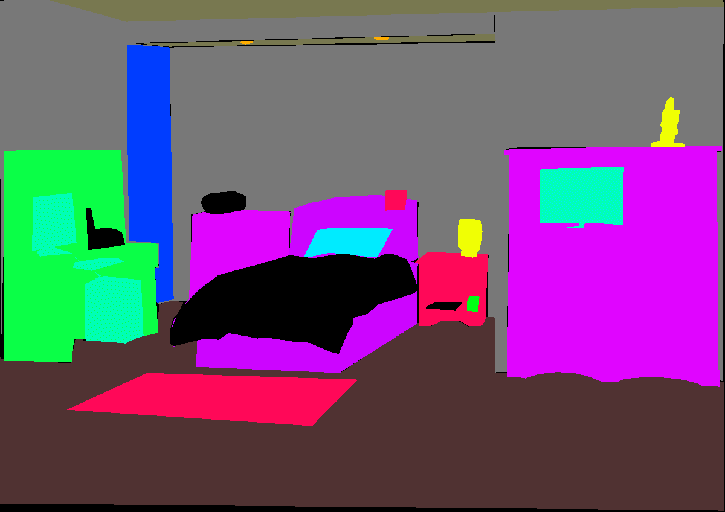}
	\includegraphics[height=\adeheight, width=\imwidth, align=b]{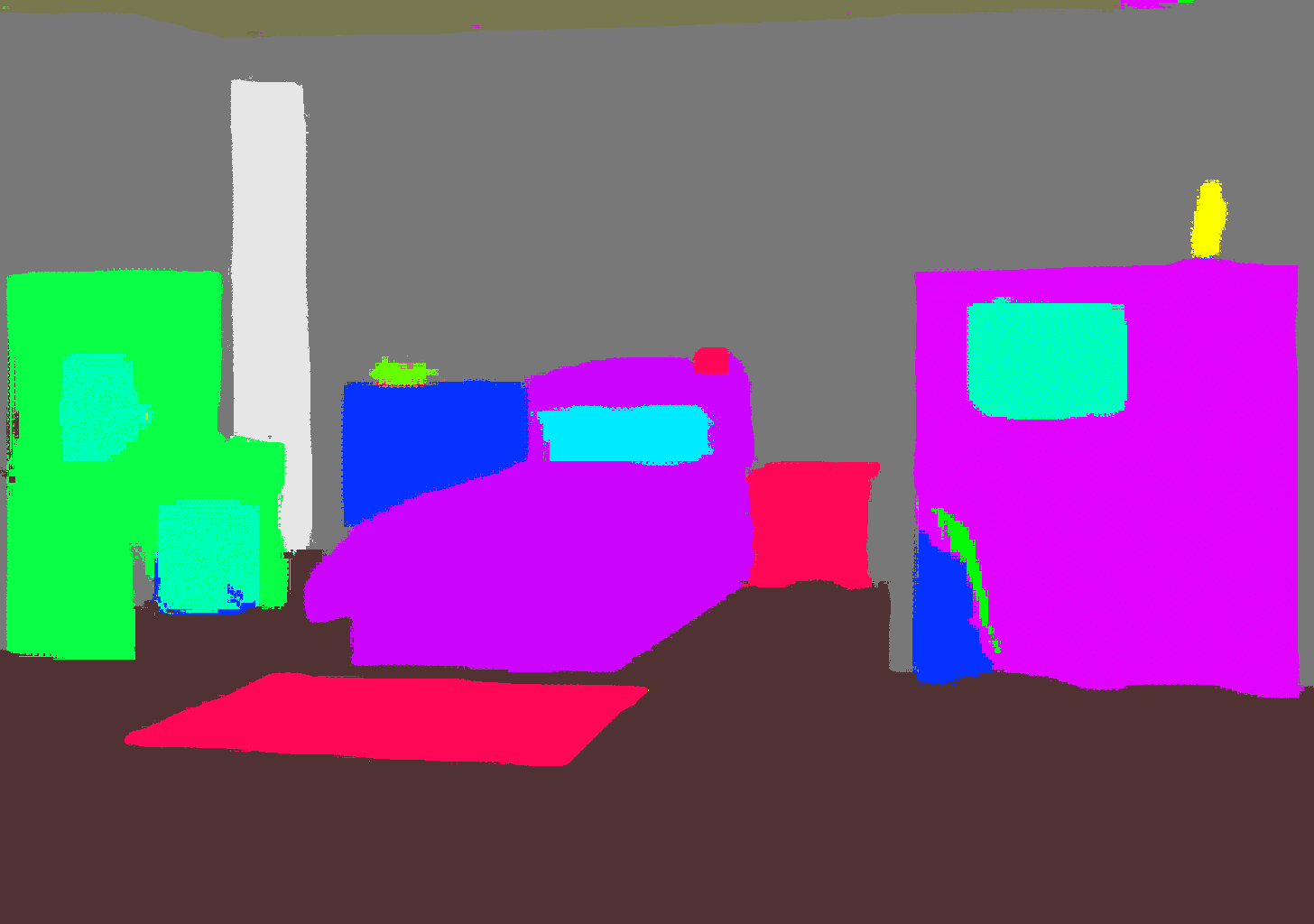}
	\\
	\rotatebox{0}{\textcolor{white}{-----------}Image}
	\rotatebox{0}{\textcolor{white}{------------}Ground Truth}
	\rotatebox{0}{\textcolor{white}{---------}Prediction}
	\rotatebox{0}{\textcolor{white}{-------------}Image}
	\rotatebox{0}{\textcolor{white}{------------}Ground Truth}
	\rotatebox{0}{\textcolor{white}{---------}Prediction\textcolor{white}{-------}}
\vspace{-0.5em}
	\caption{Qualitative results of semantic segmentation on  Cityscapes~\cite{cordts2016cityscapes} and ADE20K~\cite{zhou2019semantic} datasets.
	}
	\vspace{-0.8em}
	\label{fig:visualization_stage_2}
\end{figure*}
\subsection{Single-domain semantic segmentation}
We compare our \model{} with prior art discriminative methods and the latest generative model (UViM~\cite{kolesnikov2022uvim}, a replicated version for semantic segmentation task).
We report the results in Table~\ref{tab:cityscapes_val}
for Cityscapes~\cite{cordts2016cityscapes} and
Table~\ref{tab:ade20k_val} for ADE20K~\cite{zhou2019semantic}. 
\emph{\textbf{(i)} In comparison to discriminative methods}:
Our \model{} yields competitive performance with either Transformers (Swin) or CNNs (\eg\ ResNet-101).
For example, under the same setting, \model{} matches the result of Maskformer~\cite{cheng2021per}.
Also, {\model{}-\gssc{}-W} is competitive to the Transformer-based SETR~\cite{zheng2021rethinking} on both datasets.
\emph{\textbf{(ii)} In comparison to generative methods}: 
\model{}-\gssb{} surpasses UViM~\cite{kolesnikov2022uvim} by a large margin
whilst enjoying higher training efficiency.
Specifically, UViM takes 1,900 TPU-v3 hours for the first training stage and 900 TPU-v3 hours for the second stage.
While the first stage takes only 329.5 GPU hours with \model{}-\gssc{}-W, and zero time with \model{}-\gssb{}. The second stage of \model{}-\gssb{} requires approximately 680 GPU hours. 
This achievement is 
due to our \texttt{maskige} mechanism for enabling the use of pretrained data representation and a series of novel designs for joint probability distribution modeling.

\subsection{Cross-domain semantic segmentation}
We evaluate cross-domain zero-shot benchmark~\cite{lambert2020mseg}. 
We compare the proposed \model{} with MSeg~\cite{lambert2020mseg}, a domain generalization algorithm (CCSA)~\cite{motiian2017unified} and a multi-task learning algorithm (MGDA)~\cite{sener2018multi}.
We test both HRNet-W48~\cite{sun2019high} and Swin-Large~\cite{liu2021swin} as backbone. 
As shown in Table~\ref{tab:mseg},
our \model{} is superior to all competitors using either backbone.
This suggests that generative learning could achieve more domain-generic representation than conventional discriminative learning counterparts.

\paragraph{Domain generic \texttt{maskige}}
Being independent to the visual appearance of images, \texttt{maskige} is intrinsically domain generic. 
To evaluate this, we transfer the {\tt maskige} from MSeg to Cityscapes.
As shown in Table~\ref{tab:share_maskiage_between_cityscapes_and_mseg}, \model{} can still achieve 79.5 mIoU (1\% drop).
In comparison, image representation transfer would double the performance decrease.
\subsection{Qualitative evaluation}
We evaluate the first-stage reconstruction quality of our \model{}, UViM~\cite{kolesnikov2022uvim} and VQGAN~\cite{esser2021taming}. 
As shown in Figure~\ref{fig:visualization_stage_1}, \model{} produces almost error-free reconstruction with clear and precise edges, and UViM fails to recognize some small objects while yielding distorted segmentation. VQGAN~\cite{esser2021taming} achieves better classification accuracy but produces more ambiguous edge segmentation.
As shown in Figure~\ref{fig:visualization_stage_2}, \model{} produces fine edge segmentation for interior furniture divisions on ADE20K~\cite{zhou2019semantic} and accurately segments distant pedestrians and slender poles on Cityscapes~\cite{cordts2016cityscapes}.

\section{Conclusion}

In this paper, we have presented a {\em Generative Semantic Segmentation} (\model) approach.
Casting semantic segmentation as an image-conditioned mask generation problem, our formulation is drastically distinctive to conventional discriminative learning based alternatives.
This is established on a novel notion of maskige
and an efficient optimization algorithm in two stages:
(i) Learning the posterior distribution of the latent variables for segmentation mask reconstruction, and
(ii) minimizing the distance between posterior distribution and the prior distribution of latent variables
for enabling input images to be conditioned on.
Extensive experiments on standard benchmarks demonstrate that our \model{} achieves competitive performance in comparison to prior art discriminative counterparts, whilst establishing new state of the art in the more challenging cross-domain evaluation setting.

\paragraph{Acknowledgments}
This work was supported in part by National Natural Science Foundation of China (Grant No. 62106050), Lingang Laboratory (Grant No. LG-QS-202202-07) and Natural Science Foundation of Shanghai (Grant No. 22ZR1407500).

{\small
\bibliographystyle{ieee_fullname}
\bibliography{cited}}

\begin{thebibliography}{10}\itemsep=-1pt

\bibitem{brostow2009semantic}
Gabriel~J Brostow, Julien Fauqueur, and Roberto Cipolla.
\newblock Semantic object classes in video: A high-definition ground truth
  database.
\newblock {\em Pattern Recognition Letters}, 2009.

\bibitem{caesar2018coco}
Holger Caesar, Jasper Uijlings, and Vittorio Ferrari.
\newblock Coco-stuff: Thing and stuff classes in context.
\newblock In {\em CVPR}, 2018.

\bibitem{chen2014semantic}
Liang-Chieh Chen, George Papandreou, Iasonas Kokkinos, Kevin Murphy, and Alan~L
  Yuille.
\newblock Semantic image segmentation with deep convolutional nets and fully
  connected crfs.
\newblock {\em arXiv preprint}, 2014.

\bibitem{chen2017rethinking}
Liang-Chieh Chen, George Papandreou, Florian Schroff, and Hartwig Adam.
\newblock Rethinking atrous convolution for semantic image segmentation.
\newblock {\em arXiv preprint}, 2017.

\bibitem{chen2018encoder}
Liang-Chieh Chen, Yukun Zhu, George Papandreou, Florian Schroff, and Hartwig
  Adam.
\newblock Encoder-decoder with atrous separable convolution for semantic image
  segmentation.
\newblock In {\em ECCV}, 2018.

\bibitem{chen2022generalist}
Ting Chen, Lala Li, Saurabh Saxena, Geoffrey Hinton, and David~J Fleet.
\newblock A generalist framework for panoptic segmentation of images and
  videos.
\newblock {\em arXiv preprint}, 2022.

\bibitem{chen2021pix2seq}
Ting Chen, Saurabh Saxena, Lala Li, David~J Fleet, and Geoffrey Hinton.
\newblock Pix2seq: A language modeling framework for object detection.
\newblock In {\em ICLR}, 2021.

\bibitem{cheng2022masked}
Bowen Cheng, Ishan Misra, Alexander~G Schwing, Alexander Kirillov, and Rohit
  Girdhar.
\newblock Masked-attention mask transformer for universal image segmentation.
\newblock In {\em CVPR}, 2022.

\bibitem{cheng2021per}
Bowen Cheng, Alex Schwing, and Alexander Kirillov.
\newblock Per-pixel classification is not all you need for semantic
  segmentation.
\newblock In {\em NeurIPS}, 2021.

\bibitem{contributors2020openmmlab}
MMSegmentation Contributors.
\newblock Openmmlab semantic segmentation toolbox and benchmark.
\newblock \url{https://github.com/open-mmlab/mmsegmentation}, 2020.

\bibitem{cordts2016cityscapes}
Marius Cordts, Mohamed Omran, Sebastian Ramos, Timo Rehfeld, Markus Enzweiler,
  Rodrigo Benenson, Uwe Franke, Stefan Roth, and Bernt Schiele.
\newblock The cityscapes dataset for semantic urban scene understanding.
\newblock In {\em CVPR}, 2016.

\bibitem{dai2017scannet}
Angela Dai, Angel~X Chang, Manolis Savva, Maciej Halber, Thomas Funkhouser, and
  Matthias Nie{\ss}ner.
\newblock Scannet: Richly-annotated 3d reconstructions of indoor scenes.
\newblock In {\em CVPR}, 2017.

\bibitem{deng2009imagenet}
Jia Deng, Wei Dong, Richard Socher, Li-Jia Li, Kai Li, and Li Fei-Fei.
\newblock Imagenet: A large-scale hierarchical image database.
\newblock In {\em CVPR}, 2009.

\bibitem{esser2020disentangling}
Patrick Esser, Robin Rombach, and Bjorn Ommer.
\newblock A disentangling invertible interpretation network for explaining
  latent representations.
\newblock In {\em CVPR}, 2020.

\bibitem{esser2021taming}
Patrick Esser, Robin Rombach, and Bjorn Ommer.
\newblock Taming transformers for high-resolution image synthesis.
\newblock In {\em CVPR}, 2021.

\bibitem{everingham2010pascal}
Mark Everingham, Luc Van~Gool, Christopher~KI Williams, John Winn, and Andrew
  Zisserman.
\newblock The pascal visual object classes (voc) challenge.
\newblock {\em IJCV}, 2010.

\bibitem{fu2019dual}
Jun Fu, Jing Liu, Haijie Tian, Yong Li, Yongjun Bao, Zhiwei Fang, and Hanqing
  Lu.
\newblock Dual attention network for scene segmentation.
\newblock In {\em CVPR}, 2019.

\bibitem{geiger2013vision}
Andreas Geiger, Philip Lenz, Christoph Stiller, and Raquel Urtasun.
\newblock Vision meets robotics: The kitti dataset.
\newblock {\em The International Journal of Robotics Research}, 2013.

\bibitem{goodfellow2020generative}
Ian Goodfellow, Jean Pouget-Abadie, Mehdi Mirza, Bing Xu, David Warde-Farley,
  Sherjil Ozair, Aaron Courville, and Yoshua Bengio.
\newblock Generative adversarial networks.
\newblock {\em Communications of the ACM}, 2020.

\bibitem{he2016deep}
Kaiming He, Xiangyu Zhang, Shaoqing Ren, and Jian Sun.
\newblock Deep residual learning for image recognition.
\newblock In {\em CVPR}, 2016.

\bibitem{huang2019ccnet}
Zilong Huang, Xinggang Wang, Lichao Huang, Chang Huang, Yunchao Wei, and Wenyu
  Liu.
\newblock Ccnet: Criss-cross attention for semantic segmentation.
\newblock In {\em ICCV}, 2019.

\bibitem{isola2017image}
Phillip Isola, Jun-Yan Zhu, Tinghui Zhou, and Alexei~A Efros.
\newblock Image-to-image translation with conditional adversarial networks.
\newblock In {\em CVPR}, 2017.

\bibitem{ji2021fsnet}
Ge-Peng Ji, Keren Fu, Zhe Wu, Deng-Ping Fan, Jianbing Shen, and Ling Shao.
\newblock Full-duplex strategy for video object segmentation.
\newblock In {\em ICCV}, 2021.

\bibitem{kingma2013auto}
Diederik~P Kingma and Max Welling.
\newblock Auto-encoding variational bayes.
\newblock {\em arXiv preprint}, 2013.

\bibitem{kolesnikov2022uvim}
Alexander Kolesnikov, Andr{\'e}~Susano Pinto, Lucas Beyer, Xiaohua Zhai,
  Jeremiah Harmsen, and Neil Houlsby.
\newblock Uvim: A unified modeling approach for vision with learned guiding
  codes.
\newblock {\em arXiv preprint}, 2022.

\bibitem{lambert2020mseg}
John Lambert, Zhuang Liu, Ozan Sener, James Hays, and Vladlen Koltun.
\newblock Mseg: A composite dataset for multi-domain semantic segmentation.
\newblock In {\em CVPR}, 2020.

\bibitem{larsen2016autoencoding}
Anders Boesen~Lindbo Larsen, S{\o}ren~Kaae S{\o}nderby, Hugo Larochelle, and
  Ole Winther.
\newblock Autoencoding beyond pixels using a learned similarity metric.
\newblock In {\em ICML}, 2016.

\bibitem{li2021semantic}
Daiqing Li, Junlin Yang, Karsten Kreis, Antonio Torralba, and Sanja Fidler.
\newblock Semantic segmentation with generative models: Semi-supervised
  learning and strong out-of-domain generalization.
\newblock In {\em CVPR}, 2021.

\bibitem{liang2022gmmseg}
Chen Liang, Wenguan Wang, Jiaxu Miao, and Yi Yang.
\newblock Gmmseg: Gaussian mixture based generative semantic segmentation
  models.
\newblock In {\em NeurIPS}, 2022.

\bibitem{lin2014microsoft}
Tsung-Yi Lin, Michael Maire, Serge Belongie, James Hays, Pietro Perona, Deva
  Ramanan, Piotr Doll{\'a}r, and C~Lawrence Zitnick.
\newblock Microsoft coco: Common objects in context.
\newblock In {\em ECCV}, 2014.

\bibitem{liu2021swin}
Ze Liu, Yutong Lin, Yue Cao, Han Hu, Yixuan Wei, Zheng Zhang, Stephen Lin, and
  Baining Guo.
\newblock Swin transformer: Hierarchical vision transformer using shifted
  windows.
\newblock In {\em ICCV}, 2021.

\bibitem{long2015fully}
Jonathan Long, Evan Shelhamer, and Trevor Darrell.
\newblock Fully convolutional networks for semantic segmentation.
\newblock In {\em CVPR}, 2015.

\bibitem{lu2022unified}
Jiasen Lu, Christopher Clark, Rowan Zellers, Roozbeh Mottaghi, and Aniruddha
  Kembhavi.
\newblock Unified-io: A unified model for vision, language, and multi-modal
  tasks.
\newblock {\em arXiv preprint}, 2022.

\bibitem{lu2021soft}
Jiachen Lu, Jinghan Yao, Junge Zhang, Xiatian Zhu, Hang Xu, Weiguo Gao,
  Chunjing Xu, Tao Xiang, and Li Zhang.
\newblock Soft: softmax-free transformer with linear complexity.
\newblock In {\em NeurIPS}, 2021.

\bibitem{maddison2016concrete}
Chris~J Maddison, Andriy Mnih, and Yee~Whye Teh.
\newblock The concrete distribution: A continuous relaxation of discrete random
  variables.
\newblock {\em arXiv preprint}, 2016.

\bibitem{motiian2017unified}
Saeid Motiian, Marco Piccirilli, Donald~A Adjeroh, and Gianfranco Doretto.
\newblock Unified deep supervised domain adaptation and generalization.
\newblock In {\em ICCV}, 2017.

\bibitem{mottaghi2014role}
Roozbeh Mottaghi, Xianjie Chen, Xiaobai Liu, Nam-Gyu Cho, Seong-Whan Lee, Sanja
  Fidler, Raquel Urtasun, and Alan Yuille.
\newblock The role of context for object detection and semantic segmentation in
  the wild.
\newblock In {\em CVPR}, 2014.

\bibitem{murphy2012machine}
Kevin~P Murphy.
\newblock {\em Machine learning: a probabilistic perspective}.
\newblock MIT press, 2012.

\bibitem{neuhold2017mapillary}
Gerhard Neuhold, Tobias Ollmann, Samuel Rota~Bulo, and Peter Kontschieder.
\newblock The mapillary vistas dataset for semantic understanding of street
  scenes.
\newblock In {\em ICCV}, 2017.

\bibitem{ramesh2021zero}
Aditya Ramesh, Mikhail Pavlov, Gabriel Goh, Scott Gray, Chelsea Voss, Alec
  Radford, Mark Chen, and Ilya Sutskever.
\newblock Zero-shot text-to-image generation.
\newblock In {\em ICML}, 2021.

\bibitem{rombach2020making}
Robin Rombach, Patrick Esser, and Bj{\"o}rn Ommer.
\newblock Making sense of cnns: Interpreting deep representations and their
  invariances with inns.
\newblock In {\em ECCV}, 2020.

\bibitem{sener2018multi}
Ozan Sener and Vladlen Koltun.
\newblock Multi-task learning as multi-objective optimization.
\newblock {\em NeurIPS}, 2018.

\bibitem{song2015sun}
Shuran Song, Samuel~P Lichtenberg, and Jianxiong Xiao.
\newblock Sun rgb-d: A rgb-d scene understanding benchmark suite.
\newblock In {\em CVPR}, 2015.

\bibitem{sun2019high}
Ke Sun, Yang Zhao, Borui Jiang, Tianheng Cheng, Bin Xiao, Dong Liu, Yadong Mu,
  Xinggang Wang, Wenyu Liu, and Jingdong Wang.
\newblock High-resolution representations for labeling pixels and regions.
\newblock {\em arXiv preprint}, 2019.

\bibitem{van2017neural}
Aaron Van Den~Oord, Oriol Vinyals, et~al.
\newblock Neural discrete representation learning.
\newblock {\em NeurIPS}, 2017.

\bibitem{varma2019idd}
Girish Varma, Anbumani Subramanian, Anoop Namboodiri, Manmohan Chandraker, and
  CV Jawahar.
\newblock Idd: A dataset for exploring problems of autonomous navigation in
  unconstrained environments.
\newblock In {\em WACV}, 2019.

\bibitem{wanseaformer}
Qiang Wan, Zilong Huang, Jiachen Lu, YU Gang, and Li Zhang.
\newblock Seaformer: Squeeze-enhanced axial transformer for mobile semantic
  segmentation.
\newblock In {\em ICLR}.

\bibitem{wang2021pyramid}
Wenhai Wang, Enze Xie, Xiang Li, Deng-Ping Fan, Kaitao Song, Ding Liang, Tong
  Lu, Ping Luo, and Ling Shao.
\newblock Pyramid vision transformer: A versatile backbone for dense prediction
  without convolutions.
\newblock In {\em CVPR}, 2021.

\bibitem{wang2018nonlocal}
Xiaolong Wang, Ross Girshick, Abhinav Gupta, and Kaiming He.
\newblock Non-local neural networks.
\newblock In {\em CVPR}, 2018.

\bibitem{xiao2018unified}
Tete Xiao, Yingcheng Liu, Bolei Zhou, Yuning Jiang, and Jian Sun.
\newblock Unified perceptual parsing for scene understanding.
\newblock In {\em ECCV}, 2018.

\bibitem{xiao2019generative}
Zhisheng Xiao, Qing Yan, and Yali Amit.
\newblock Generative latent flow.
\newblock {\em arXiv preprint}, 2019.

\bibitem{xie2021segformer}
Enze Xie, Wenhai Wang, Zhiding Yu, Anima Anandkumar, Jose~M Alvarez, and Ping
  Luo.
\newblock Segformer: Simple and efficient design for semantic segmentation with
  transformers.
\newblock In {\em NeurIPS}, 2021.

\bibitem{yu2020bdd100k}
Fisher Yu, Haofeng Chen, Xin Wang, Wenqi Xian, Yingying Chen, Fangchen Liu,
  Vashisht Madhavan, and Trevor Darrell.
\newblock Bdd100k: A diverse driving dataset for heterogeneous multitask
  learning.
\newblock In {\em CVPR}, 2020.

\bibitem{yuan2019object}
Yuhui Yuan, Xilin Chen, and Jingdong Wang.
\newblock Object-contextual representations for semantic segmentation.
\newblock In {\em ECCV}, 2020.

\bibitem{zendel2018wilddash}
Oliver Zendel, Katrin Honauer, Markus Murschitz, Daniel Steininger, and
  Gustavo~Fernandez Dominguez.
\newblock Wilddash-creating hazard-aware benchmarks.
\newblock In {\em ECCV}, 2018.

\bibitem{zhang2020dynamic}
Li Zhang, Dan Xu, Anurag Arnab, and Philip~HS Torr.
\newblock Dynamic graph message passing networks.
\newblock In {\em CVPR}, 2020.

\bibitem{zhao2017pyramid}
Hengshuang Zhao, Jianping Shi, Xiaojuan Qi, Xiaogang Wang, and Jiaya Jia.
\newblock Pyramid scene parsing network.
\newblock In {\em CVPR}, 2017.

\bibitem{zheng2021rethinking}
Sixiao Zheng, Jiachen Lu, Hengshuang Zhao, Xiatian Zhu, Zekun Luo, Yabiao Wang,
  Yanwei Fu, Jianfeng Feng, Tao Xiang, Philip~HS Torr, et~al.
\newblock Rethinking semantic segmentation from a sequence-to-sequence
  perspective with transformers.
\newblock In {\em CVPR}, 2021.

\bibitem{zhou2019semantic}
Bolei Zhou, Hang Zhao, Xavier Puig, Tete Xiao, Sanja Fidler, Adela Barriuso,
  and Antonio Torralba.
\newblock Semantic understanding of scenes through the ade20k dataset.
\newblock {\em IJCV}, 2019.

\end{thebibliography}

\newpage
\appendices
\section{Proofs}
\subsection{Derivation of GSS ELBO}
We provide the proof of Eq.~\eqref{equ:elbo_sem} in the main paper here.
We rewrite the log-likelihood of semantic segmentation $\log{p(c|x)}$ by introducing  a discrete L-dimension latent distribution $q(z|c)$ (with $z\in \mathbb{Z}^L$).
\begin{align*}
    \notag\log{p(c|x)} &= \log{\int p(c,z|x)} \text{\ d}z\\
    \notag &= \log{\int p(c,z|x)} \frac{q(z|c)}{q(z|c)}\text{\ d}z\\
    \notag &=\log\mathbb{E}_{q(z|c)}\left[\frac{p(c,z|x)}{q(z|c)}\right]\\
    \notag &\geq \mathbb{E}_{q(z|c)}\left[\log \frac{p(c,z|x)}{q(z|c)} \right]\\
    \notag \left( \text{as} \right.-log(\cdot)& \text{ is convex, by Jensen's Inequality: }\\
    \notag f(\sum_i\lambda_i x_i)& \leq \sum_i \lambda_if(x_i)\text{, where } \lambda_i\geq 0, \sum_i\lambda_i=1 \left. \right)\\
    \notag  &= \mathbb{E}_{q(z|c)}\left[\log \frac{p(c|z)p(z|x)}{q(z|c)} \right]\\
    \notag  &= \mathbb{E}_{q(z|c)}\left[\log p(c|z) \right] + \mathbb{E}_{q(z|c)}\left[\log \frac{p(z|x)}{q(z|c)} \right]\\
    \notag  = \mathbb{E}_{q(z|c)}&\left[ \log p(c|z) \right] -D_{KL}(q(z|c), p(z|x)) \\
    \notag  = \mathbb{E}_{q_\phi(z|c)}&\left[ \log p_\theta(c|z) \right] -D_{KL}(q_\phi(z|c), p_\psi(z|x)).
\end{align*}

Different from ELBO~\cite{kingma2013auto} in VAE, the latent variable we introduce here is $q(z|c)$, rather than $q(z)$ to solve the conditioned mask generation problem.

\subsection{Derivation of latent posterior learning}
We provide the proof of Eq.~\eqref{eq:posterior} in the main paper here.
As stated in main paper, the first stage latent posterior training is conducted by a MSE loss
\begin{equation*}
    \min_{\theta, \phi}\sum_c \mathbb{E}_{q_\phi(z|c)}\|p_\theta(c|z)-c\|.
\end{equation*}

Let us denote $\hat{c}=p_\theta(c|z)$ is the reconstructed mask.
Then, we define a linear transform $x^{(c)}=\mathcal{X}_\beta(c)=c\beta$, where $\beta\in\mathbb{R}^{K\times 3}$ and an arbitrary inverse transform $\hat{c}=\mathcal{X}^{-1}_\gamma(\hat{x}^{(c)})$.
Noted that the parameter $\gamma$ can be non-linear.
$x^{(c)}$ is called \texttt{maskige} and $\hat{x}^{(c)}$ is the reconstructed \texttt{maskige} produced by the {\tt maskige} decoder $\hat{x}^{(c)}=\mathcal{D}_{{\theta}}(\hat{z})$. 
The transformed latent parameter $\hat{z}$ preserves the probability for the linear transformation,
\begin{figure}[tb]
	\begin{center}
\includegraphics[width=1.0\linewidth]{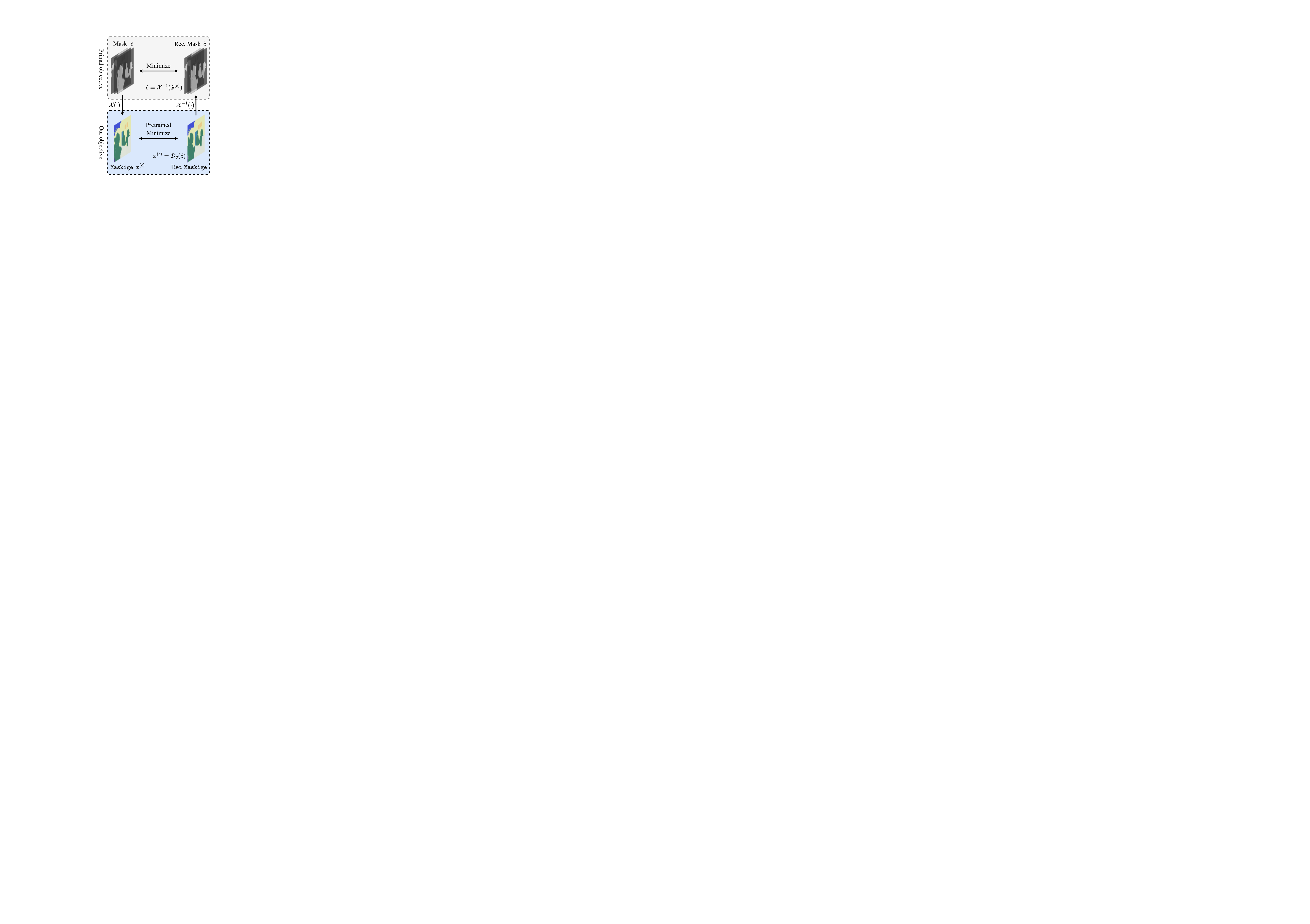}
	\end{center}
	\caption{
	\textbf{An illustration of our transformed objective.} \texttt{Rec.} stands for reconstruction.}
	\label{fig:sup_obj}
\end{figure}
\begin{equation}
\label{equ:qhatz}
    q_{\hat{\phi}}(\hat{z}|x^{(c)}) =q_{\hat{\phi}}(\hat{z}|c\beta)= q_\phi(z|c).
\end{equation}
Then, we have
\begin{align}
    \notag &\min_{\theta, \phi}\sum_c \mathbb{E}_{q_\phi(z|c)}2\|\hat{c}-c\|\\
    \notag =&\min_{\theta, \phi, \gamma}\sum_c \mathbb{E}_{q_\phi(z|c)}\left[\|\mathcal{X}^{-1}(\hat{x}^{(c)})-c\| + \|\hat{c}-c\|\right].
\end{align}
For the first term, since $\mathcal{X}^{-1}(\hat{x}^{(c)})=\mathcal{X}^{-1}(\mathcal{D}_{{\theta}}(\hat{z}))$ which is not related to $\theta$ and $\phi$.
Therefore, we have
\begin{align}
    \notag &\min_{\theta, \phi, \gamma}\sum_c \mathbb{E}_{q_\phi(z|c)}\|\mathcal{X}^{-1}(\hat{x}^{(c)})-c\|\\
     \notag=&\min_{\gamma}\sum_c \mathbb{E}_{q_\phi(z|c)}\|\mathcal{X}^{-1}(\mathcal{D}_{\theta}(\hat{z}))-c\|\\
    \label{equ:t1}
    =&\min_{\gamma,\beta}\sum_c \mathbb{E}_{q_{\hat{\phi}}(\hat{z}|x^{(c)})}\|\mathcal{X}^{-1}(\mathcal{D}_{{\theta}}(\hat{z}))-c\|\ (\text{Eq.~\eqref{equ:qhatz}}).
\end{align}
\begin{figure*}[tb]
	\begin{center}
		\includegraphics[width=0.9\linewidth]{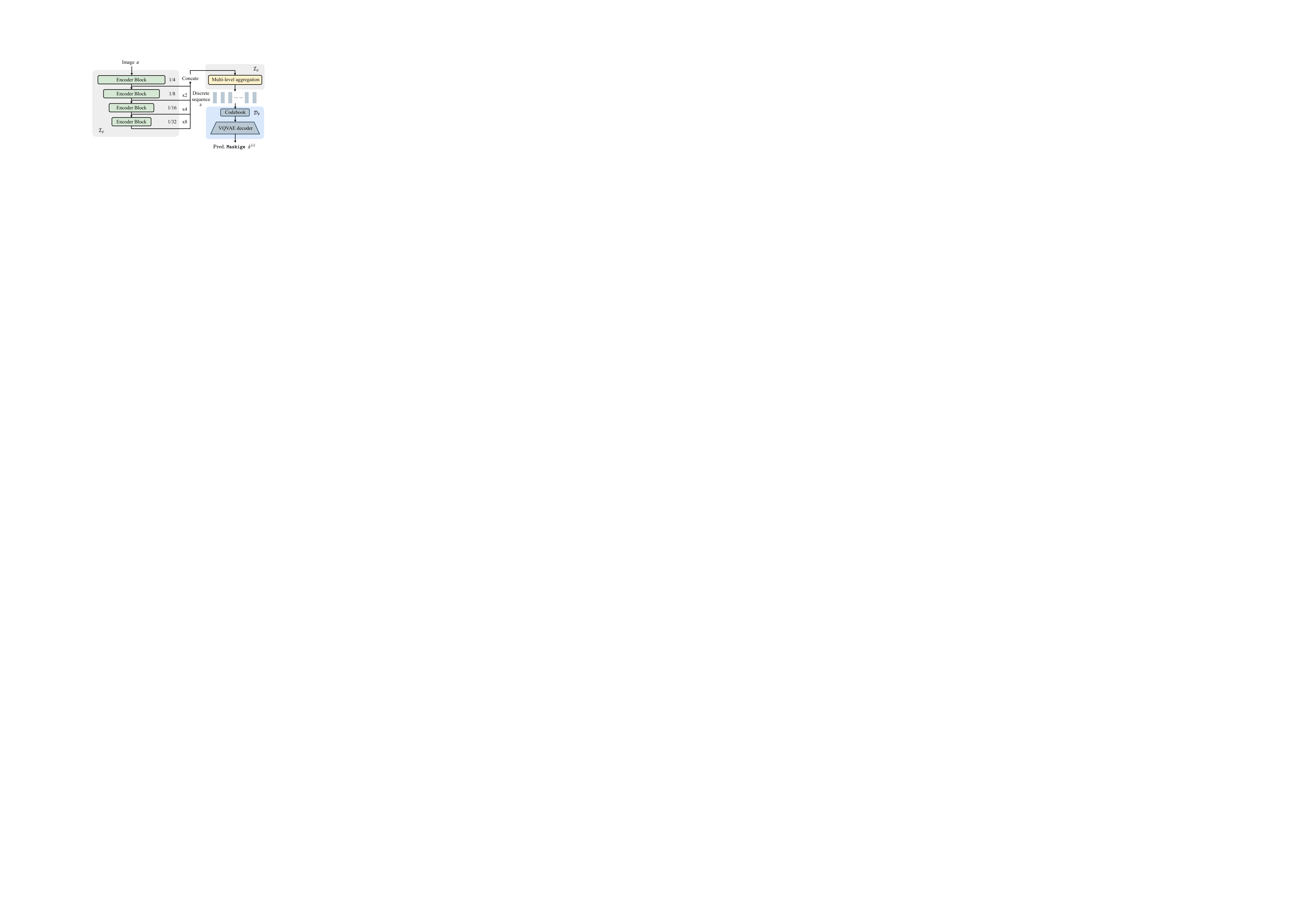}
	\end{center}
	\caption{
\textbf{Encoder-decoder style architecture of \model{}.} ``Pred." stands for prediction.}
	\label{fig:sup_arch}
\end{figure*}

For the second term, $\hat{c}=\mathcal{D}_\theta(z)$ is not related to $\gamma$, which can be rewritten as
\begin{align}
    \notag &\min_{\theta, \phi}\sum_c \mathbb{E}_{q_\phi(z|c)}\|\hat{c}-c\|\\
    \notag &=\min_{\theta, \phi, s.t. \|\beta\|=1}\sum_c \mathbb{E}_{q_\phi(z|c)}\|\hat{c}-c\|\|\beta\|\\
    \notag &=\min_{\theta, \phi, s.t. \|\beta\|=1}\sum_c \mathbb{E}_{q_\phi(z|c)}\|\hat{c}\beta-c\beta\|\\
    \notag &= \min_{\theta, \phi, s.t. \|\beta\|=1} \sum_c
    \mathbb{E}_{q_\phi(z|c)}\|(\hat{c}\beta-\hat{x}^{(c)}) \ (\text{equal to 0 by def.})\\
    \notag &\quad\quad\quad\quad\quad\quad\quad\quad+ (\hat{x}^{(c)} - x^{(c)})\\
    \notag&\quad\quad\quad\quad\quad\quad\quad\quad+ (x^{(c)} -c\beta)\|\ (\text{equal to 0 by def.})\\
    \notag &=\min_{\theta, \phi, s.t. \|\beta\|=1}\sum_{x^{(c)}} \mathbb{E}_{q_\phi(z|c)} \|\hat{x}^{(c)} - x^{(c)}\|\ (\text{not related to }\theta)\\
    \notag &=\min_{{\theta}, \phi, s.t. \|\beta\|=1}\sum_{x^{(c)}} \mathbb{E}_{q_\phi(z|c)} \|\mathcal{D}_{{\theta}}(\hat{z}) - x^{(c)}\|\\
    \notag &=\min_{{\theta}, \hat{\phi}, \beta, s.t. \|\beta\|=1}\sum_{x^{(c)}} \mathbb{E}_{q_{\hat{\phi}}(\hat{z}|x^{(c)})}
    \label{equ:t2}\|\mathcal{D}_{{\theta}}(\hat{z}) - x^{(c)}\|\ (\text{Eq.~\eqref{equ:qhatz}}).
\end{align}
Combining Eq.~\eqref{equ:t1} and Eq.~\eqref{equ:t2}, our final objective is 
\begin{align}
    \notag& \min_{\hat{\phi},{\theta}, \beta, s.t. \|\beta\|=1}\sum_{x^{(c)}} \mathbb{E}_{q_{\hat{\phi}}(\hat{z}|x^{(c)})}\|\mathcal{D}_{{\theta}}(\hat{z})-x^{(c)}\|\\
    + &\quad \min_{\gamma, \beta}\sum_c\mathbb{E}_{q_{\hat{\phi}}(\hat{z}|x^{(c)})}\|\mathcal{X}^{-1}(\mathcal{D}_{{\theta}}(\hat{z}))-c\|.
\end{align}
For the first term, it is a VQVAE~\cite{van2017neural} reconstruction objective for \texttt{maskige}.
Therefore, a VQVAE pretrained by DALL$\cdot$E~\cite{ramesh2021zero} with a large-scale OpenImage dataset can 
readily offer a good 
lower bound for the first term.

As such, only the second term is left for optimization.
We can optimize the $\gamma$ with gradient descent,
corresponding to {\model{}-\gssc{}\&\gsse{}}. 
Besides, we can solve this problem more efficiently by a {linear assumption}, \ie{} $\mathcal{X}^{-1}(\hat{x}^{(c)}) = \hat{x}^{(c)}\gamma$ where $\gamma\in\mathbb{R}^{3\times K}$.
We denote the $\hat{X}^{(c)}$ is a matrix with each row an reconstructed \texttt{maskige} and $C$ is a matrix with each row an input mask.
We solve the optimization with least square error
\begin{align}
    \notag&\|\mathcal{X}^{-1}(\hat{X}^{(c)})-C\|^2\\
    \notag=&\|\hat{X}^{(c)}\gamma-C\|^2\\
    \notag=&\|(\hat{X}^{(c)}-X^{(c)}+X^{(c)})\gamma-C\|^2\\
    \label{equ:opt_gamma_beta_mid}
    \leq&\left(\|\hat{X}^{(c)}-C\beta\|\|\gamma\|+\|X^{(c)}\gamma-C\|\right)^2.
\end{align}
The optimization over both $\beta$ and $\gamma$ is non-convex (as shown by the poor performance with {\model{}-\gsse{}}), so we optimize them sequentially in {\model{}-\gssb{}}\&{\gssc{}\&\gssd{}}.
For {\model{}-\gssb{}}, we use a hand-crafted optimized $\beta$.
\begin{align}
    \notag&\left(\|\hat{X}^{(c)}-C\beta\|\|\gamma\|+\|X^{(c)}\gamma-C\|\right)^2\\
    \notag\leq&\left(\tau\|\gamma\|+\|X^{(c)}\gamma-C\|\right)^2\\
    \notag=&\left(\tau\|\gamma\|+\|C\beta\gamma-C\|\right)^2\\
    \label{equ:opt_gamma_beta_mid2}
    \leq&\left(\tau\|\gamma\|+\|C\|\|\beta\gamma-\mathbb{1}\|\right)^2.
\end{align}
where $\tau=\|\hat{X}^{(c)}-C\beta\|$ is bounded and unrelated to $\gamma$ by provided VQVAE and $\beta$.
Our objective then changes to minimize the upper bound.
\begin{align}
    \notag \min_{\gamma, s.t. \|\gamma\|=1} \text{RSS}(\gamma)&=\min_{\gamma, s.t. \|\gamma\|=1} \|\beta\gamma-\mathbb{1}\|^2\\
    \label{equ:rss}
    &=\min_{\gamma, s.t. \|\gamma\|=1}(\beta\gamma-\mathbb{1})^\top(\beta\gamma-\mathbb{1}).
\end{align}
We take the derivative of Eq.~\eqref{equ:rss}, then
\begin{equation}
\label{equ:drss}
    \frac{\partial \text{RSS}}{\partial \gamma}=2\beta^\top(\beta\gamma-\mathbb{1})=0.
\end{equation}
The unique solution of Eq.~\eqref{equ:drss} is 
\begin{align}
    \notag\beta^\top\beta\gamma&=\beta^\top\\
    \notag(\Rightarrow)\quad (\beta^\top\beta)^{-1}(\beta^\top\beta)\gamma &= (\beta^\top\beta)^{-1}\beta^\top\\
    \label{equ:gamma_mse}
    (\Rightarrow)\quad\quad\quad\quad\quad\quad\quad\ \ \gamma &= (\beta^\top\beta)^{-1}\beta^\top.
\end{align}

For the special design {\model{}-\gssd{}}, we use a cascaded optimization to automatically optimize $\beta$ and $\gamma$.
\begin{align}
    \notag\beta_{t+1} &= \mathop{\arg\min}\limits_{\beta}\left(\|\hat{X}^{(c)}-C\beta\|\|\gamma_t\|+\|C\beta\gamma_t-C\|\right)^2\\
    \notag\gamma_{t+1} &= (\beta_{t+1}^\top\beta_{t+1})^{-1}\beta_{t+1}^\top.
\end{align}
The $\notag\beta_{t+1}$ is optimized by one mini-batch step of gradient descent.

\begin{table*}[htb]
\tablestyle{3.8pt}{1.05}
\centering
\renewcommand\tabcolsep{3.9pt}
\renewcommand\arraystretch{1.2}
\small
\begin{tabular}{llcccccc|c}
\hline

\hline

\hline

\hline
Method & Iteration &VOC~\cite{everingham2010pascal} & Context~\cite{mottaghi2014role} & CamVid~\cite{brostow2009semantic} & WildDash~\cite{zendel2018wilddash}   & KITTI~\cite{geiger2013vision}  & ScanNet~\cite{dai2017scannet} & \textit{h.\ mean} \\
\hline

\hline
\emph{- Discriminative modeling:}\\
\shline
CCSA~\cite{motiian2017unified} & 500k &48.9 & - & 52.4 & 36.0 & - & 27.0 & 39.7 \\
MGDA~\cite{sener2018multi} & 500k &69.4 & - & 57.5 & 39.9 & - & 33.5 & 46.1 \\
MSeg-w/o relabel~\cite{lambert2020mseg}  & 500k & 70.2 & 42.7 & 82.0 & 62.7 &  65.5 & 43.2 & 57.6\\
MSeg~\cite{lambert2020mseg}  &  500k & 70.7 & 42.7 & 83.3 & 62.0 & 67.0 & 48.2 & 59.2\\
MSeg-480p~\cite{lambert2020mseg} & 1,500k & 76.4 & 45.9 & 81.2 & 62.7 & \textbf{68.2} & 49.5 & 61.2 \\
MSeg-720p~\cite{lambert2020mseg} & 1,500k & 74.7 & 44.0 & 83.5 & 60.4 & 67.9 & 47.7 & 59.8 \\
MSeg-1080p~\cite{lambert2020mseg} & 1,500k & 72.0 & 44.0 & \textbf{84.5} & 59.9 & 66.5 & 49.5 & 59.8 \\
\hline

\hline
 \emph{- Generative modeling:}\\
 \shline
\rowcolor[gray]{.9}
\model-\gssb{}~(Ours) & 160k & 78.7 & 45.8 & 74.2 & 61.8 & 65.4 & 46.9 & 59.5\\
\rowcolor[gray]{.9}
\model-\gssc{}-W~(Ours) & 160k &
\textbf{79.5} & \textbf{47.7} & 75.9 & \textbf{65.3} & {68.0} &  \textbf{49.7} & \textbf{61.9}\\
\hline

\hline

\hline

\hline
\end{tabular}
\caption{\textbf{Additional cross-domain semantic segmentation performance on MSeg dataset {\tt test} split~\cite{lambert2020mseg}:}  
We add performance of Mseg-480p, Mseg-720p and Mseg-1080p~\cite{lambert2020mseg} results to Table~\ref{tab:mseg} of the main paper. \textbf{No improved versions of our methods are included.}
``480p", ``720p" and ``1080p" mean all test images are resized to 480p (the shorter side is 480 pixel), 720p and 1080p, respectively, when MSeg model inference.
}
\label{tab:supp_mseg}
\end{table*}

\section{{\tt Maskige} optimization designs}

As illustrated above, $\beta$ is a linear projection applied on the ground-truth mask, \ie{} $x^{(c)}=c\beta$.
As is shown in Eq.~\eqref{equ:opt_gamma_beta_mid} and Eq.~\eqref{equ:opt_gamma_beta_mid2}, the quality of $\|\hat{X}^{(c)}-C\beta\|$ directly affects the quality of the following $\gamma$ optimization, so the optimization of $\beta$ will affect the difficulty degree of our learning of $\mathcal{X}_\gamma^{-1}$.
The cascaded optimization in {\model{}-\gssd{}} (84.37\% reconstruction mIoU) provides an upper-bound for $\beta$ optimization. 
However, the gradient descent optimization costs another $5$ GPU hours according to Table~\ref{tab:vqvae} of main paper.
Therefore, we propose a hand-crafted optimization of $\beta$ in {\model-\gssb{}\&\gssc{}\&\gssc{}-W} that achieves satisfying performance without requiring extra training time, based on the \emph{maximal distance assumption}.

To understand this assumption, we can consider the linear projection parameter $\beta \in \mathbb{R}^{K \times 3}$ as a colorization process, where each category is assigned an \texttt{rgb} color. The idea behind the {\em maximal distance assumption} is to maximize the color difference between the encoding of each category. For instance, if two different categories are assigned similar colors, the model may struggle to differentiate between them. Therefore, by maximizing the distance between color embeddings, we can improve the model's ability to distinguish between categories.
We interpret the parameter $\beta$ as R, G, B color sequences $\mathcal{A}^r, \mathcal{A}^g, \mathcal{A}^b$ assigned to each category. To better satisfy the \emph{maximal distance assumption}, we will try different ways to construct these sequences, \ie., assigning colors to each category.

\emph{\textbf{(i)} Arithmetic sequence on R/G/B channels}: 
Designing three arithmetic sequences $\mathcal{A}^r, \mathcal{A}^g, \mathcal{A}^b$ for R/G/B channels respectively. Then we have
\begin{equation}
\mathcal{A}^m = \left \{a_1^m, a_2^m, \dots, a_i^m, \dots, a_n^m \right \}, m \in \left \{r, g, b\right \}.
\end{equation}
For the $i$-th color value,
\begin{equation}
      a_i^m = a_1^m + (i-1)\cdot k^m,k^m \in N^+, 
\end{equation}
where color channel $m \in \left \{r, g, b\right \}$, the interval of arithmetic sequence $k^m$ can be difference between channels, $a_1^m$ default is $0$.
The set of colors is the Cartesian product of these three series, 
\begin{equation}
    \mathcal{C} =  \mathcal{A}^r \times \mathcal{A}^g \times \mathcal{A}^b.
\end{equation}
\Eg, if the interval of R, G, B channel $k=45$, the color set $\mathcal{C}$ will be $\left \{(0, 0, 0)\right .$, $(0, 0, 45), \dots,$ $\left .(225, 225, 225) \right \}$.

\emph{\textbf{(ii)} Misalignment start points}: The original starting point of the arithmetic sequence is 0, 0, 0 for R/G/B respectively. In order to avoid duplication of values, we let R/G/B have different starting points, 
\begin{equation}
    a_1^{r}\ne a_1^{g} \ne a_1^{b}.
\end{equation}
In practice, we simply set to $a_1^r=0, a_1^g=1, a_1^b=2$.

\emph{\textbf{(iii)} Random additive factors}: Adding three independent random factors $t \in [0, T]$ on the R/G/B arithmetic sequence respectively, to avoid repetition of several same values,
\begin{equation}
    a_i^m = a_1^m + (i-1)\cdot k^m + t_i^m.
\end{equation}
\Eg, a color sequence with random additive factors: $\left \{(1, 7, 3)\right .$, $(4, 2, 45)$, $\dots$, $\left .(235, 215, 232)\right \}$. In practical terms, $T$ is set to $15$.

\emph{\textbf{(iv)} Category-specific refinement}: We equip the lower IoU categories with values where the R/G/B values vary at large degrees (\eg, we replace $(128, 128, 128)$ with $(0, 128, 255)$). In addition, we keep the color away from gray as possible, because gray is located in the center of the color space, thus being close to many categories and giving rise to a harder learning problem.
Such an category-specific refinement allows each category to
be possibly furthest from the others as possible.

\begin{table}[t]
\tablestyle{1.8pt}{1.05}

\renewcommand\tabcolsep{13.1pt}
\renewcommand\arraystretch{1.2}
\small
\begin{center}
 
\begin{tabular}{l|cc}
\hline

\hline

\hline

\hline
{Colorization technique} & {mIoU} & {aAcc} \\
\hline

\hline
Arithmetic sequence & 85.99 & 94.37  \\
~~~+ Misalignment start points & 86.12 & 94.45  \\
~~~+ Random additive factors & 87.42 & 95.08  \\
\rowcolor[gray]{.9}
~~~+ Category-specific refinement & \textbf{87.73} & \textbf{95.29}  \\

\hline

\hline

\hline

\hline
\end{tabular}
\end{center}
\caption{
\textbf{Ablation on \emph{maximal distance assumption}:} The \texttt{maskige} reconstruction performance (mIoU and aAcc) of {\model{}-\gssb} on ADE20K {\tt val} split under different Mask-to-\texttt{maskige} transformations $\fx$.
}
\label{tab:supp_colorization}
\end{table}

\paragraph{Results}

As shown in Table~\ref{tab:supp_colorization},
it is evident that the colorization design for \texttt{maskige} generation
presents a good amount of impact on the reconstruction performance.
In particular, the last design {category-specific refinement}
yields the best results, conforming our intuition and design consideration.

\begin{figure*}[htb]
	\begin{center}
		\includegraphics[width=12.86cm, height=22cm]{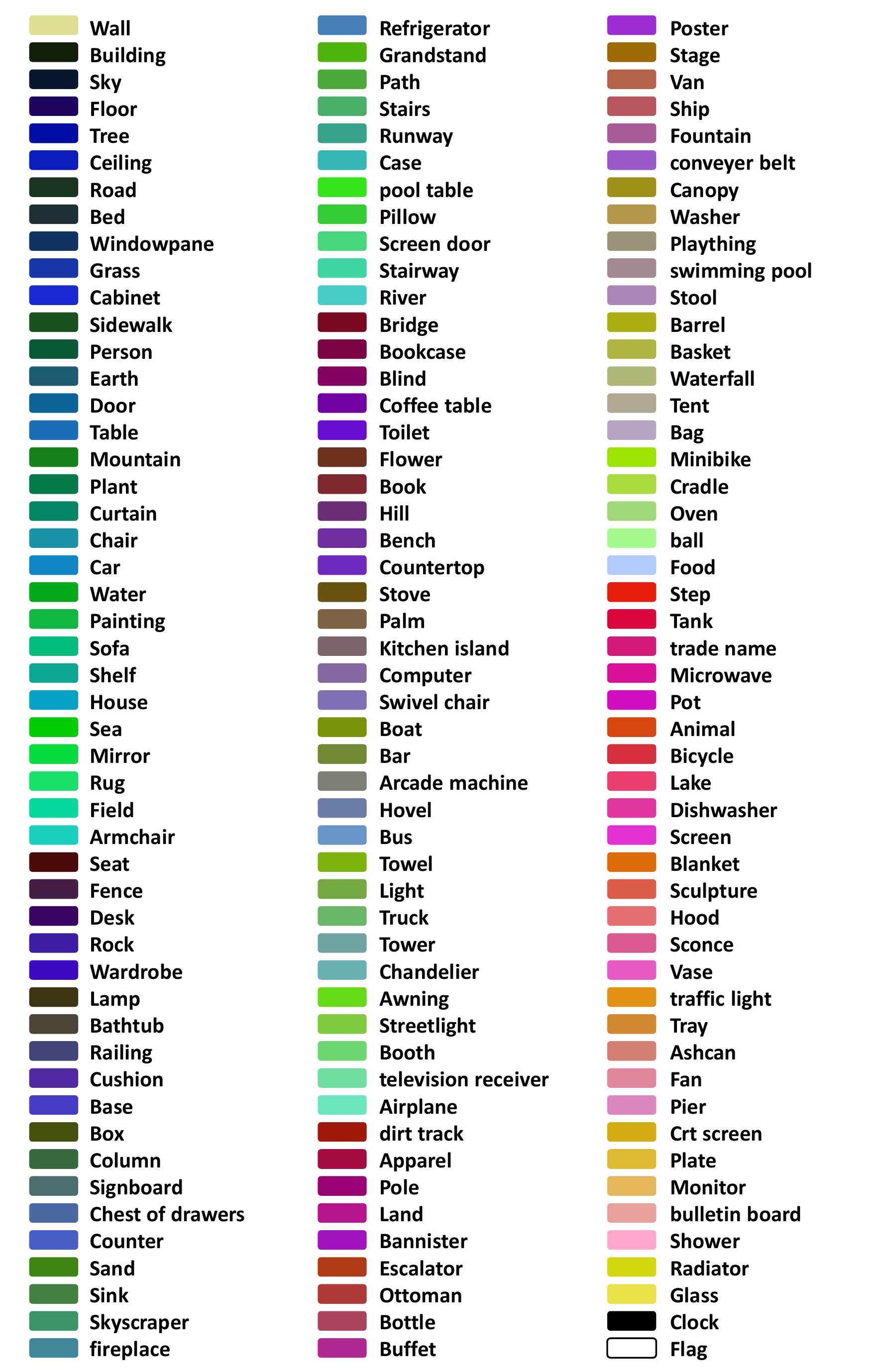}
	\end{center}
	\caption{
Visualization of \texttt{maskige} for each category in ADE20K~\cite{zhou2019semantic} dataset under \emph{maximal distance assumption}.
	}
	\label{fig:supp_color_ade_mda}
\end{figure*}

\begin{figure*}[htb]
	\begin{center}
		\includegraphics[width=12.86cm, height=22cm]{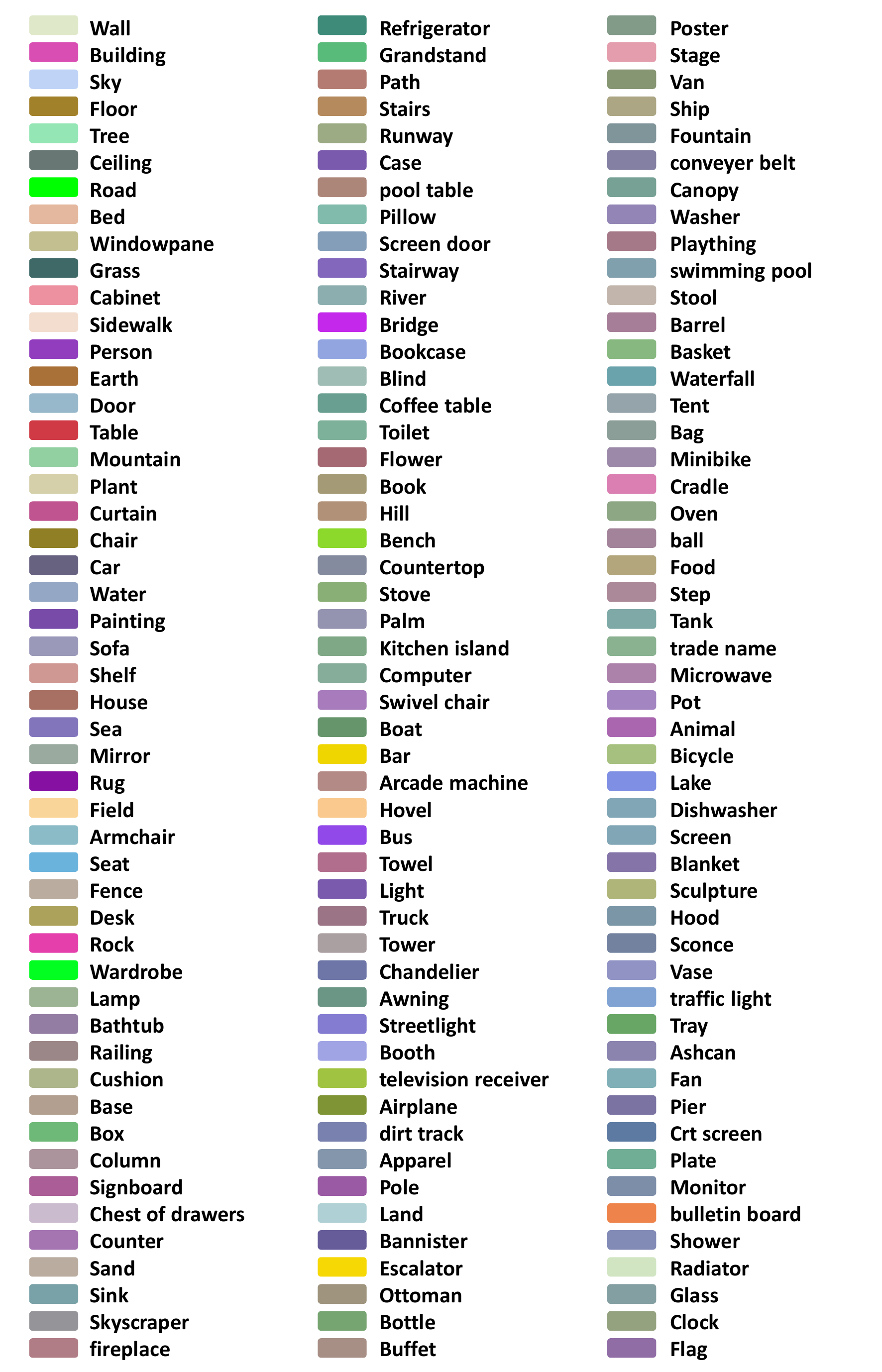}
	\end{center}
	\caption{
Visualization of \texttt{maskige} for each category in ADE20K~\cite{zhou2019semantic} dataset under gradient descent optimization.
	}
	\label{fig:supp_color_ade_opt}
\end{figure*}

\paragraph{Visualization}
For visual understanding,
in Figure~\ref{fig:supp_color_ade_mda} and Figure~\ref{fig:supp_color_ade_opt} we visualize the 150 colors corresponding to all the categories of ADE20K~\cite{zhou2019semantic} generated by 
the \emph{maximal distance assumption} (hand-designed) and gradient descent optimization (learned), respectively.
We observe that the hand-designed method
produces the colors with enhanced contrast and greater vibrancy.
Instead, the colors learned 
are vibrant for the more frequent categories and relatively dark for the less frequent categories.

\section{Overall architecture}
Following \cite{esser2021taming, ramesh2021zero}, the modeling of latent prior learning is formulated by an encoder-decoder architecture (See Figure~\ref{fig:sup_arch}).
For the image encoder $\mathcal{I}_\psi$, we take the advantage of hierarchical shifted window transformer~\cite{liu2021swin}
for extracting the multi-scale information~\cite{wang2021pyramid} and memory efficiency~\cite{lu2021soft}.
This is different from UViM \cite{kolesnikov2022uvim}, which uses a single-scale and full-range Transformer as the encoder.
To implement the image encoder, we use the Swin-Large architecture \cite{liu2021swin}, pre-trained on ImageNet-22K \cite{deng2009imagenet}, as the backbone.
As shown in Figure~\ref{fig:sup_arch}, we use four-scale feature maps ($1/4$, $1/8$, $1/16$, $1/32$) and upsample all the lower-resolution features to $1/4$ scale, then concatenate four features across the channel \cite{xie2021segformer}.
The multi-level aggregation consists of an MLP and $D$ layers of hierarchical shifted-window Transformer \cite{liu2021swin}, with the swin window size set to 7, the number of attention heads to 16, the embedding dimension to 512, and the FFN dimension to 1024. 
For the implementation version with resnet as the backbone, $D=6$. However, for models with strong Swin Transformer backbones, fewer MLA layers are needed, and thus $D=2$.
We implement the {\tt maskige} decoder $\mathcal{D}_\theta$ as a fixed VQVAE decoder \cite{van2017neural}.

\section{More training details}
\noindent \emph{\textbf{(i)} Latent posterior learning}: 
As illustrated before, the latent posterior learning is simplified as:
\begin{equation}
    \min_{\fxpi}\mathbb{E}_{q_{\hat{\phi}}(\hat{z}|\mathcal{X}(c))}\|\mathcal{X}^{-1}(\hat{x}^{(c)})-c\|.
\end{equation} 
The target can be interpreted as minimizing the distance between a ground-truth segmentation mask and the predicted mask.
Following \cite{kolesnikov2022uvim, chen2022generalist}, we use cross-entropy loss instead of euclidean distance for a better minimization between segmentation masks.

\noindent \emph{\textbf{(ii)} Latent posterior learning for $\fx$}: For GSS variants whose $\mathcal{X}$ dose not require training, such as {\model{}-\gssb{}}\&{\gssc{}}\&{\gssc{}-W}, we assign a 3-channel encoding to each category directly based on the {\em maximum distance assumption}. 
For the GSS variants that require training, including {\model{}-\gssd{}}\&{\gssd{}}\&, we freeze the parameters of the VQVAE of the DALL-E pretrain and train $\fx$ for 4,000 iterations using \texttt{SGD} optimizer with a batch size of 16. 
By the way, the training process of \model{}-\gsse{} also optimizes the $\fxpi$ function.

\noindent \emph{\textbf{(iii)}~Latent posterior learning for $\fxpi$}: 
We propose a method for training an $\fxpi$ that is more robust to noise (used in {\model{}-\gssc{}}\&{\gssc{}-W}). We found that training $\fxpi$ with \textit{noisy \texttt{maskige}} helps it learn to be more robust. To provide noisy {\tt maskige}, we can use the trained $\mathcal{I}_\psi$, DALL$\cdot$E pre-trained $\mathcal{D}_\theta$, and $\fxpi$ to directly predict noisy {\tt maskige} predictions. In practice, we use $\mathcal{I}_\psi$ that trained up to the middle checkpoint (\eg, 32,000 iterations) or final checkpoint in latent prior learning. $\fxpi$ is trained using cross-entropy loss and optimized with \texttt{AdamW}, with a batch size of 16. We trained \model{}-{\gssc{}-W} for 40,000 iterations, while \model{}-\gssc{} was trained for only 3,000 iterations due to its fast convergence.
The non-linear function $\fxpi$ is implemented using either a convolutional or Swin block structure. 
Specifically, for \model{}-\gssc{}, the structure comprises two conv $1 \times 1$ layers enclosing a conv $3 \times 3$ layer. 
However, this approach is superseded by \model{}-\gssc{}-W, the final model, which employs a group of Swin blocks with a number of heads of 4, a Swin window size of 7, and an embedding channel of 128 to realize $\fxpi$. 
Regardless of the specific implementation, $\fxpi$ relies on local RGB information in the predicted mask to deduce the category of each pixel.

\noindent \emph{\textbf{(iv)}~Latent prior learning}:
For the optimization of~~$\mathcal{I}_\psi$, we use the AdamW optimizer and implement a polynomial learning rate decay schedule\cite{zhao2017pyramid} with a minimum learning rate of $0.0$. We set the initial learning rate to $1.5 \times 10^{-3}$ for Cityscapes and $1.2 \times 10^{-4}$ for ADE20K and Mseg.

\section{Domain generic \texttt{maskige} and image encoder}
We did two tests by deriving a {\bf \em general maskige} on MSeg~\cite{lambert2020mseg}.

\noindent \text{\textbf{(i)}}
As shown in Table~\ref{tab:share_maskiage_between_cityscapes_and_mseg}, we applied our general \texttt{maskige} to the Cityscapes dataset and achieved a mIoU score of 79.5, which is only slightly lower than the mIoU score of 80.5 obtained using the Cityspaces specific \texttt{maskige}. This result demonstrates the versatility of our \texttt{maskige} across different datasets.

\noindent \text{\textbf{(ii)}} To further evaluate the effectiveness of our domain-generic approach, we shared the image encoder $\mathcal{I}_\psi$ between MSeg and Cityscapes and trained our model on the training split of MSeg. We then evaluated the model on the {\bf \em zero-shot} test split consisting of 6 unseen datasets. As shown in Table~\ref{tab:mseg}, our GSS outperforms other state-of-the-art methods on the MSeg dataset. These experiments demonstrate that our \texttt{maskige} is domain-generic and has the potential for open-world settings.

\section{Additional quantitative results}
We additionally compare the improved versions of MSeg~\cite{lambert2020mseg} with 1,500k longer training on the cross-domain benchmark.
As shown in Table~\ref{tab:supp_mseg},
despite using short training, our model still achieves 
better performance.
This verifies the advantage of our method
in terms of training efficiency, in addition to the accuracy.

\section{Additional qualitative results}
For further qualitative evaluation,
we visualize the prediction results of our GSS on both single-domain segmentation datasets~\cite{cordts2016cityscapes,zhou2019semantic} and cross-domain segmentation dataset~\cite{lambert2020mseg}.

\begin{figure*}[th]
	\def \imwidth {5.6cm}
	\def \cityheight {3cm}
	\centering
	\includegraphics[height=\cityheight, width=\imwidth, align=b]{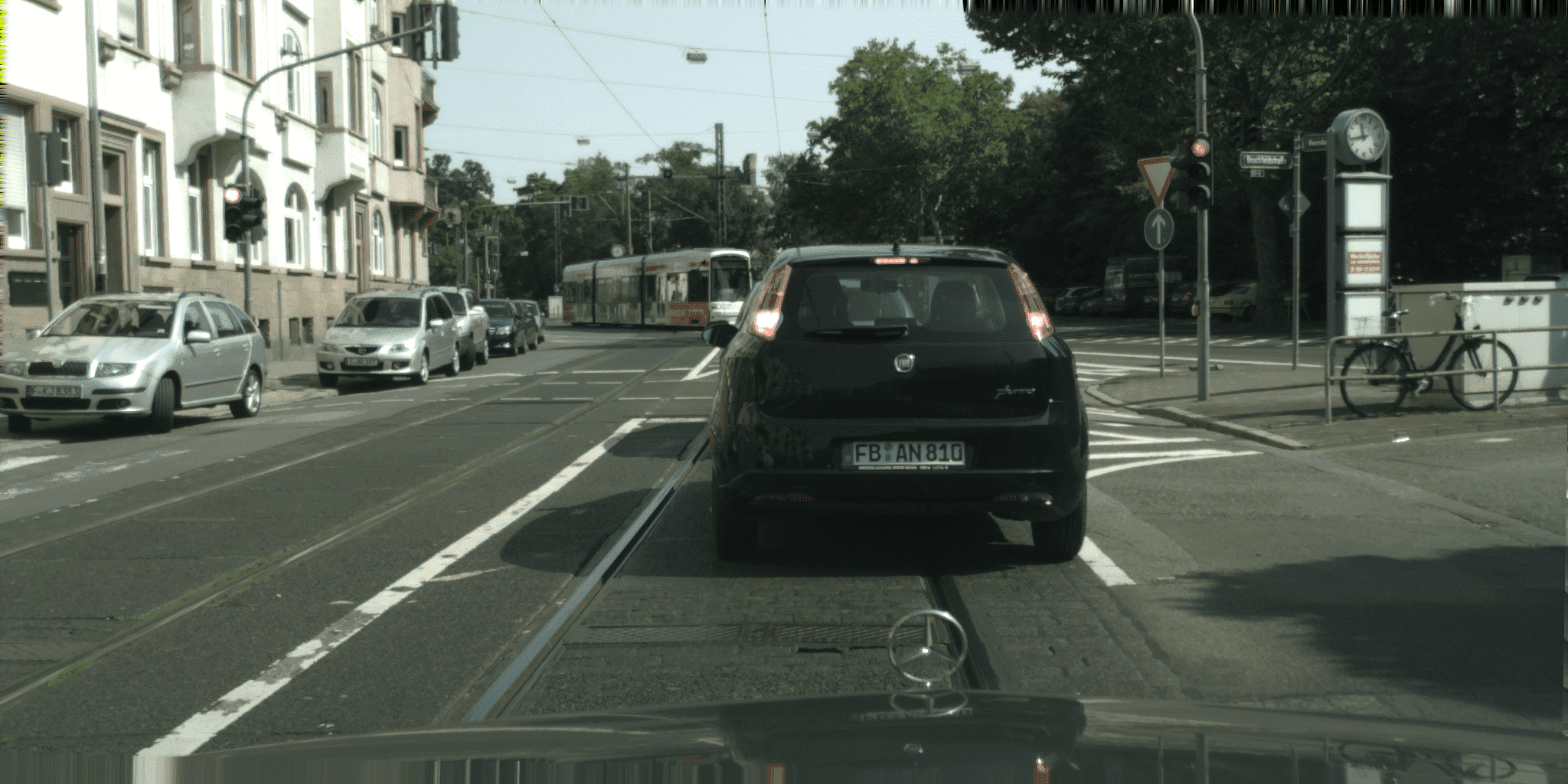}
	\includegraphics[height=\cityheight, width=\imwidth, align=b]{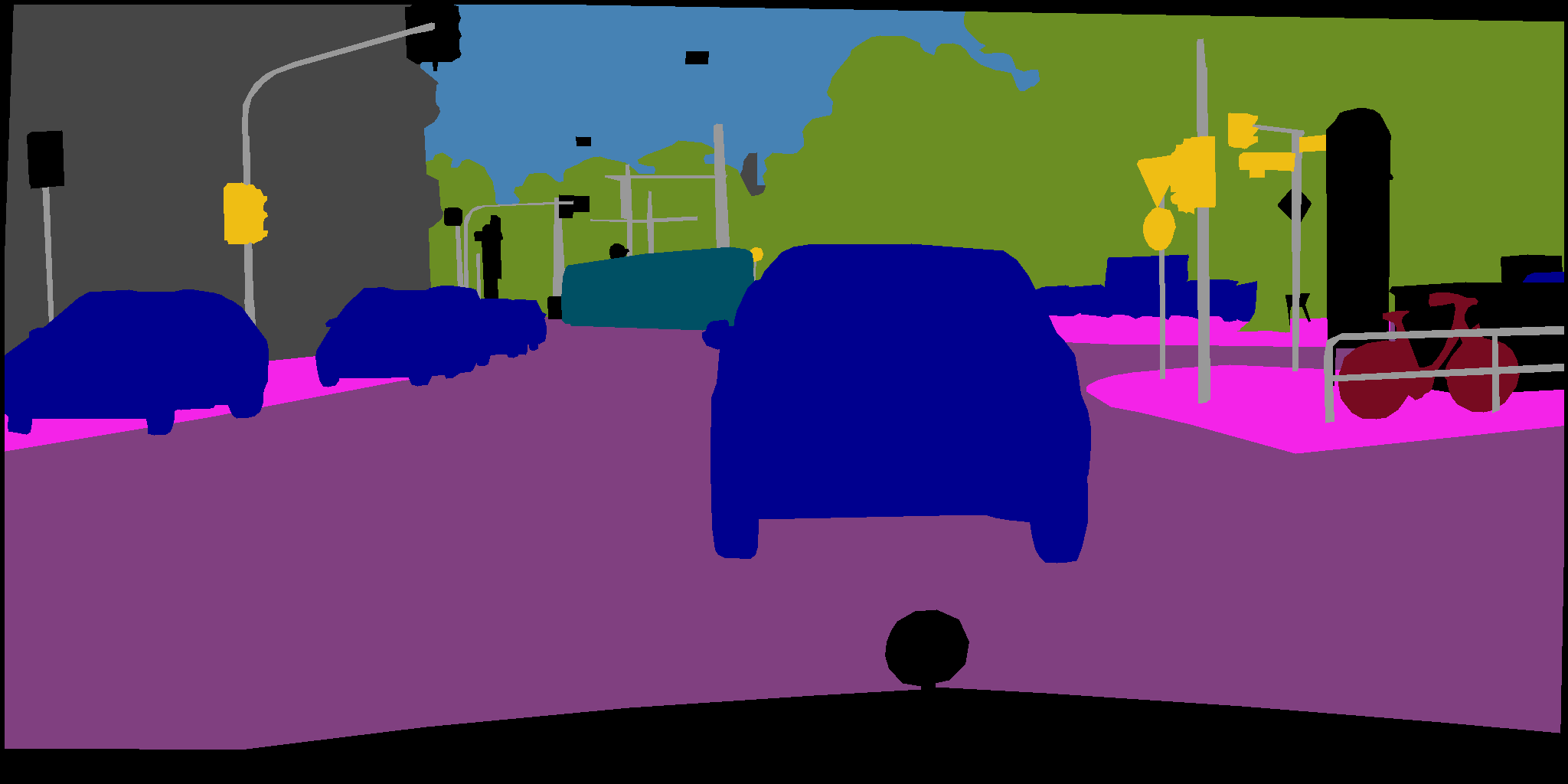}
	\includegraphics[height=\cityheight, width=\imwidth, align=b]{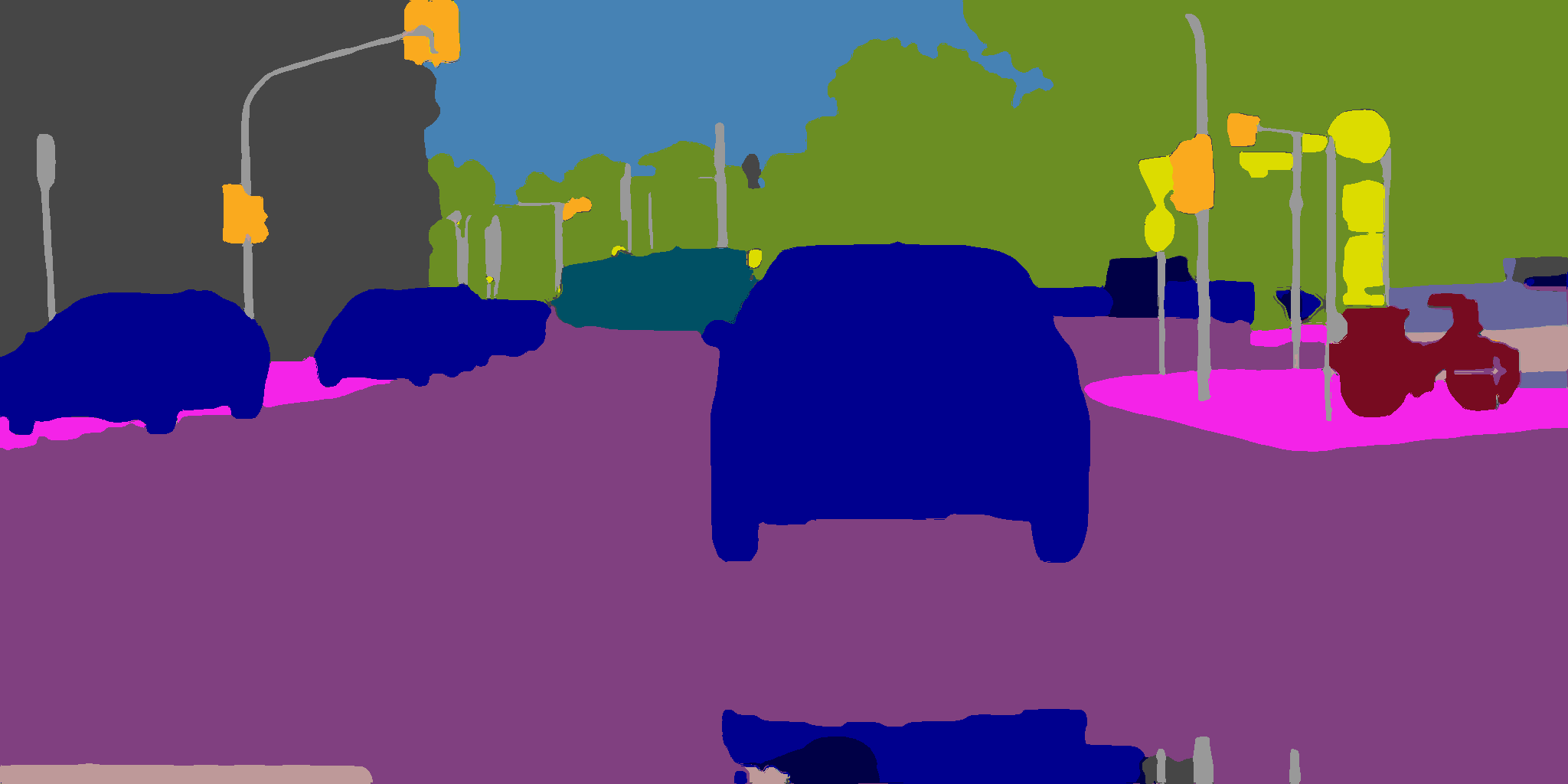}\\
	\includegraphics[height=\cityheight, width=\imwidth, align=b]{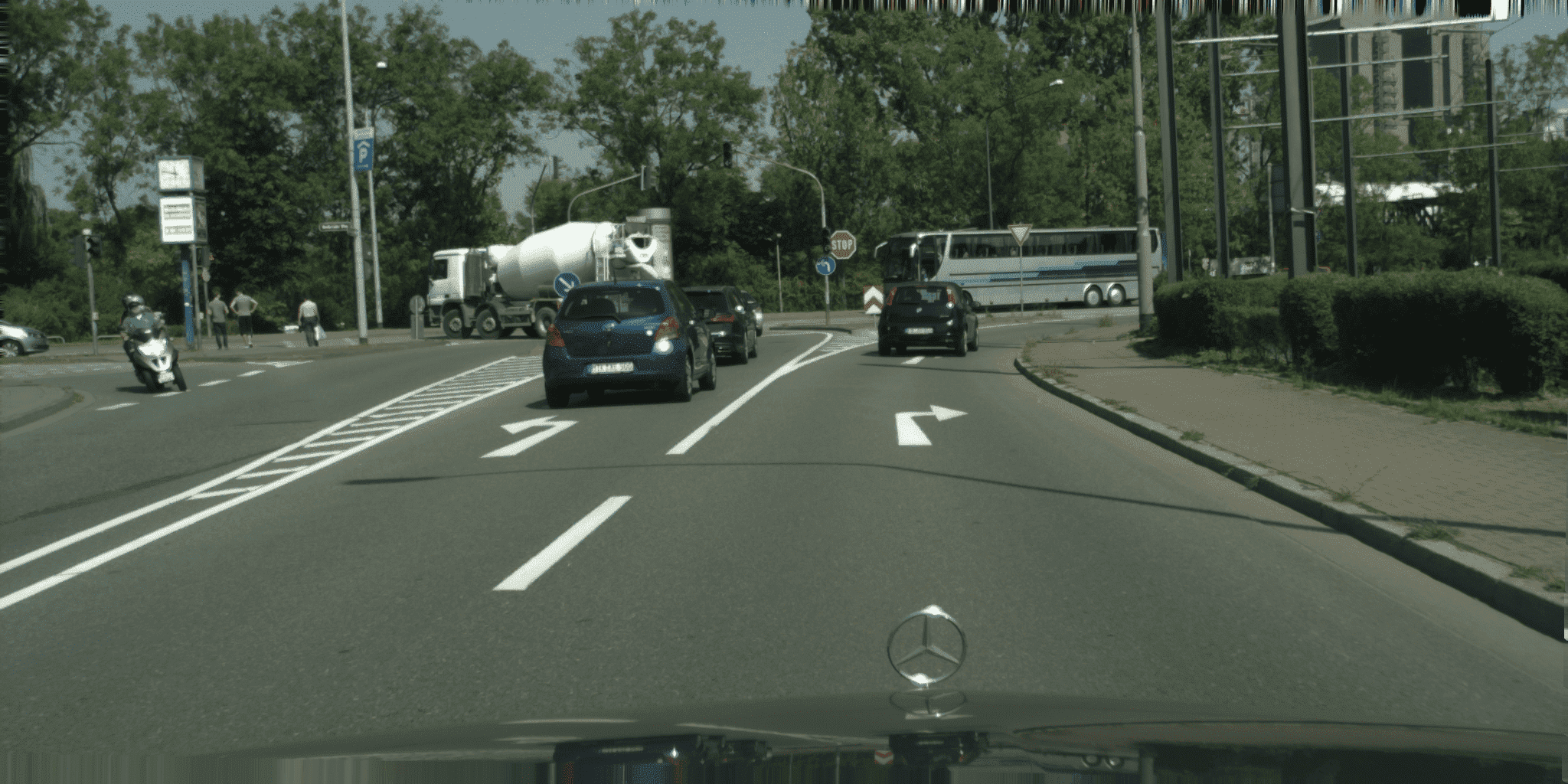}
	\includegraphics[height=\cityheight, width=\imwidth, align=b]{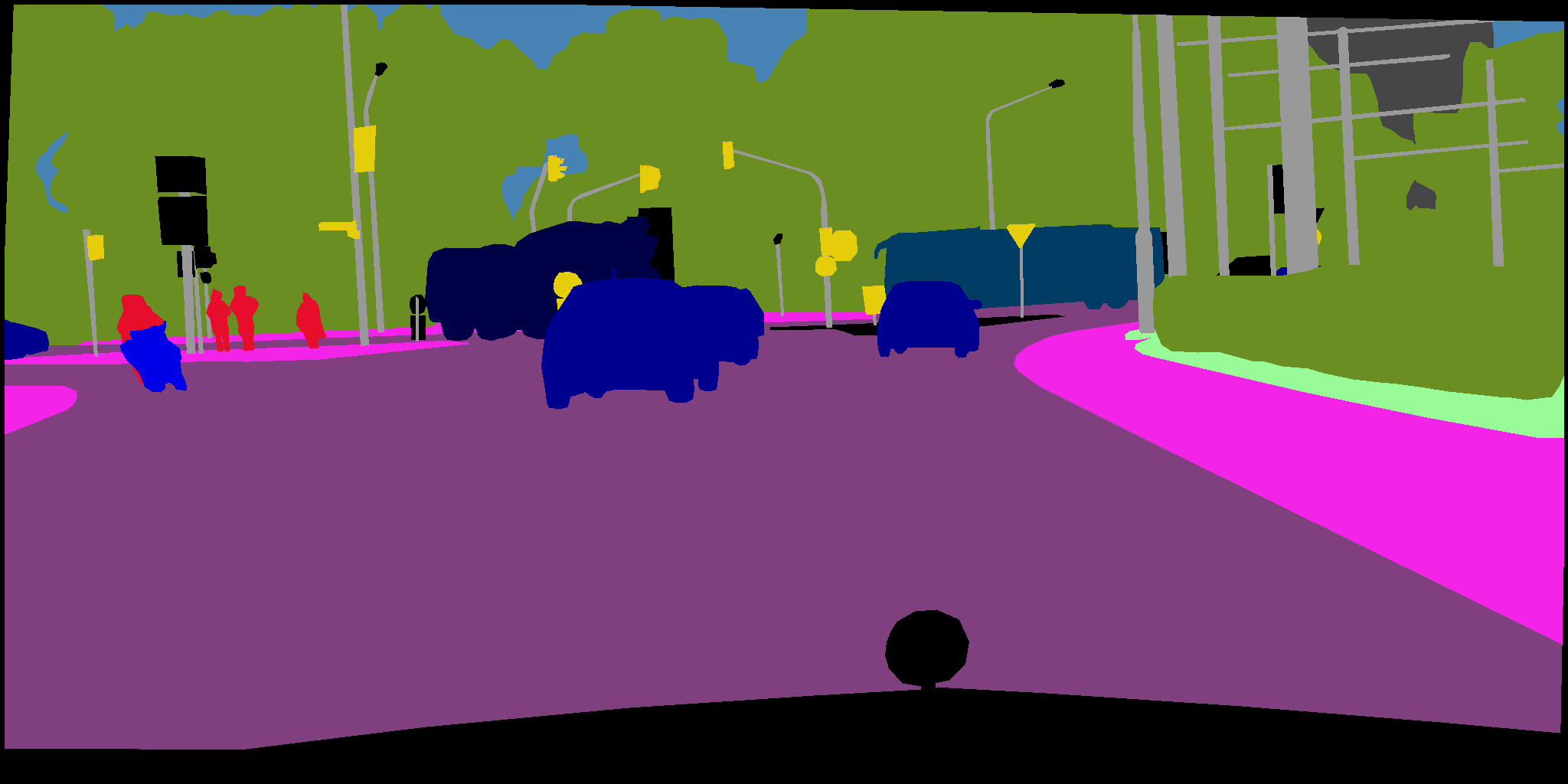}
	\includegraphics[height=\cityheight, width=\imwidth, align=b]{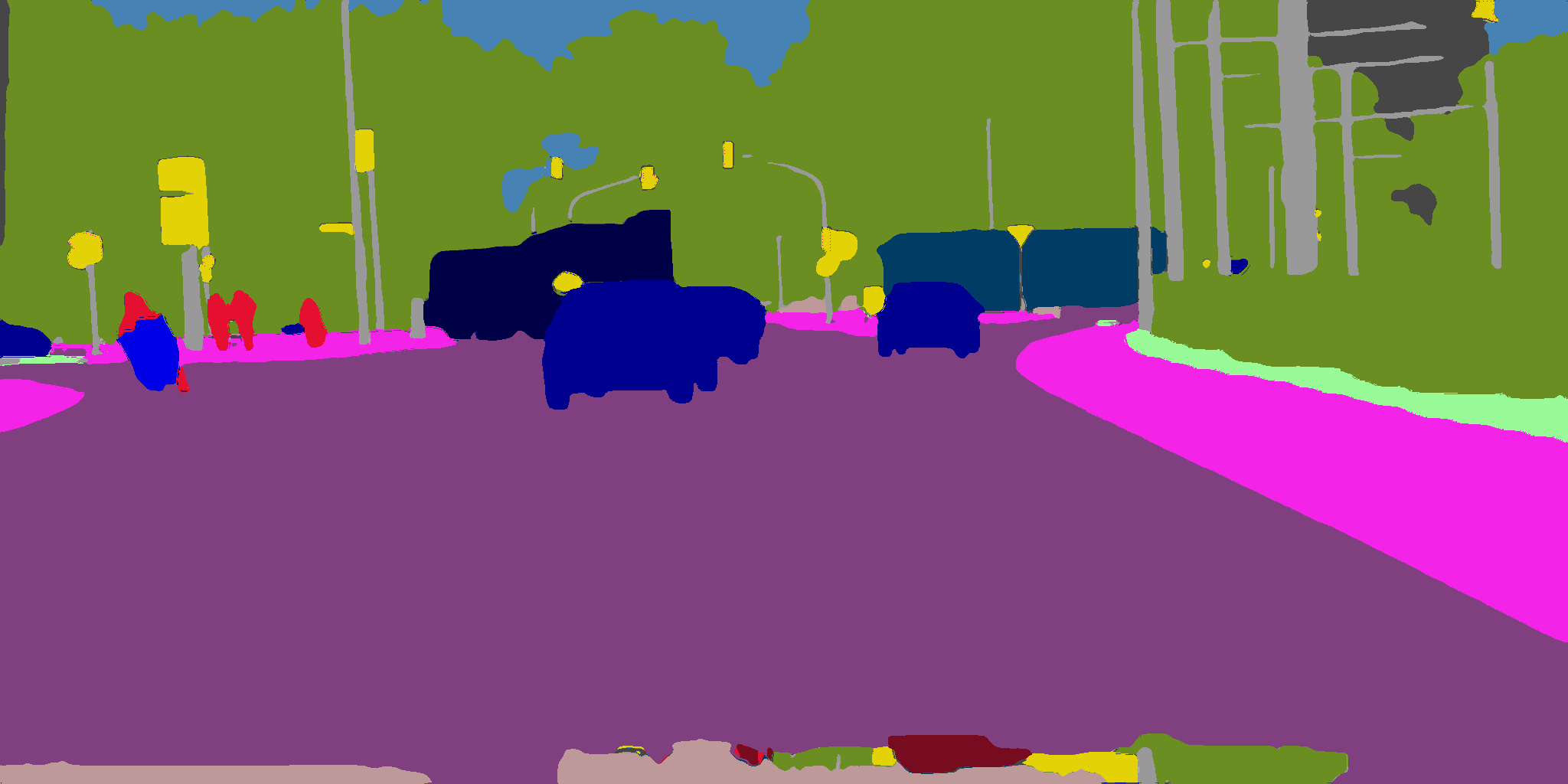}
	\\
 	\includegraphics[height=\cityheight, width=\imwidth, align=b]{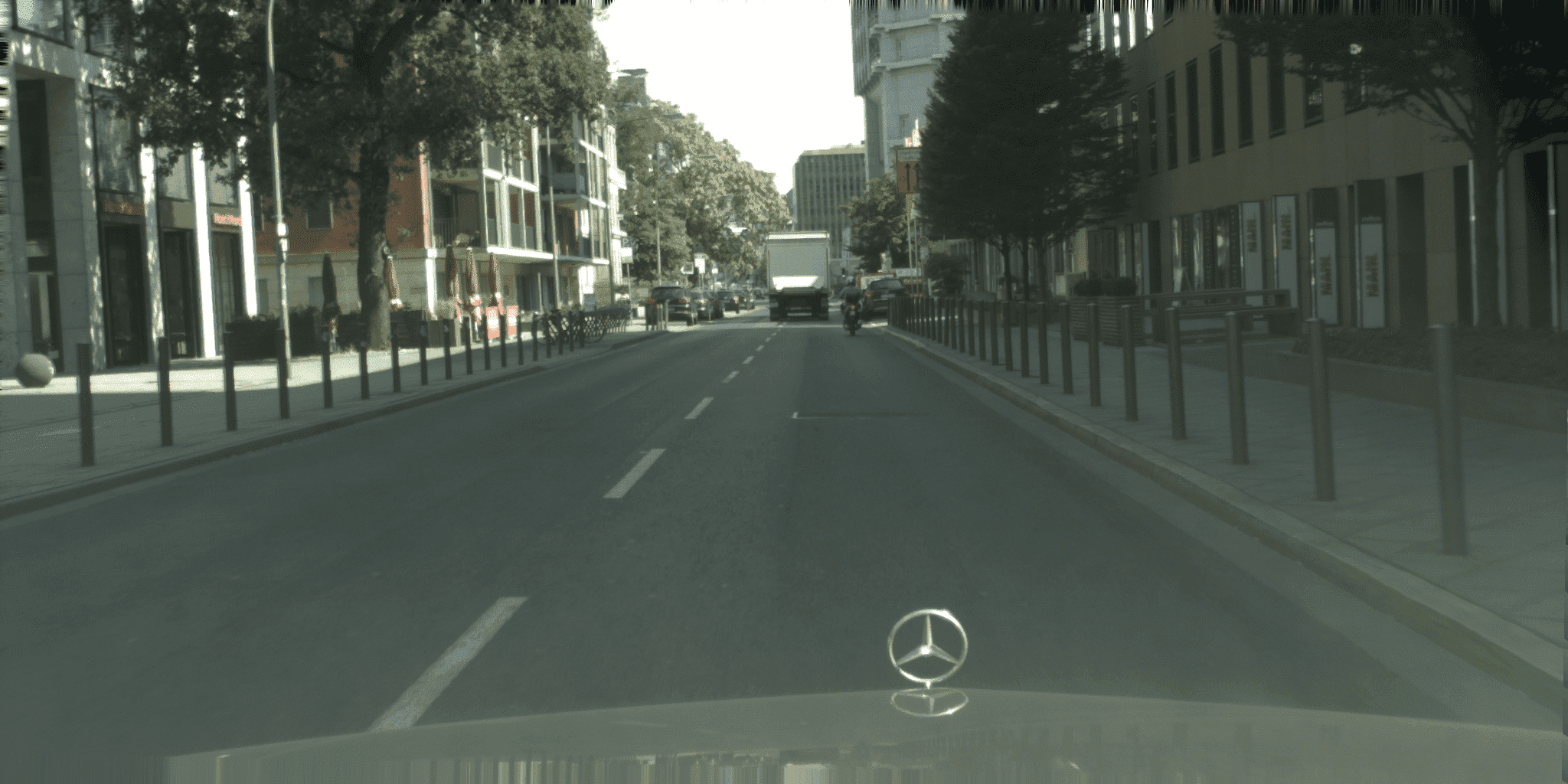}
	\includegraphics[height=\cityheight, width=\imwidth, align=b]{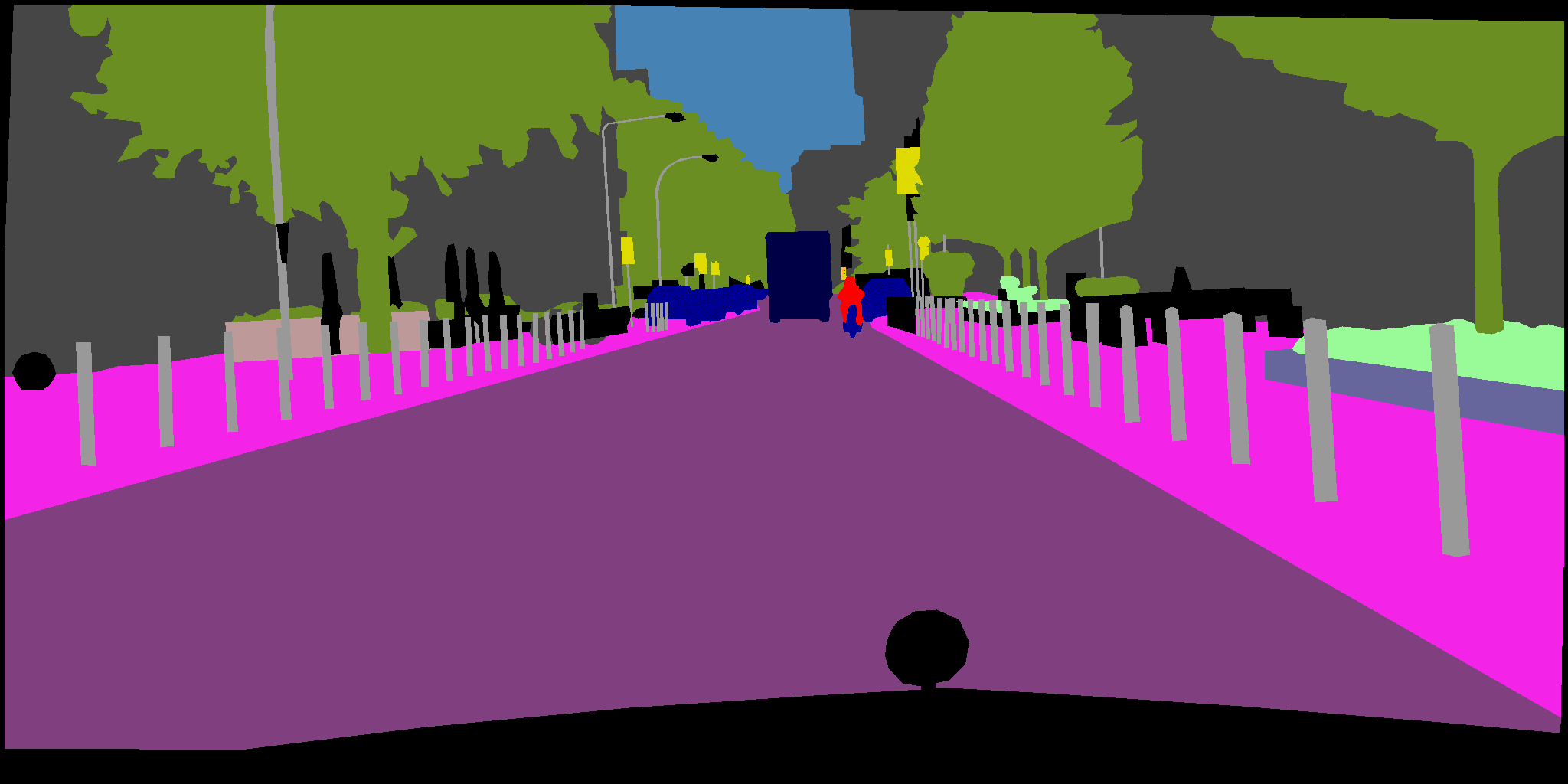}
	\includegraphics[height=\cityheight, width=\imwidth, align=b]{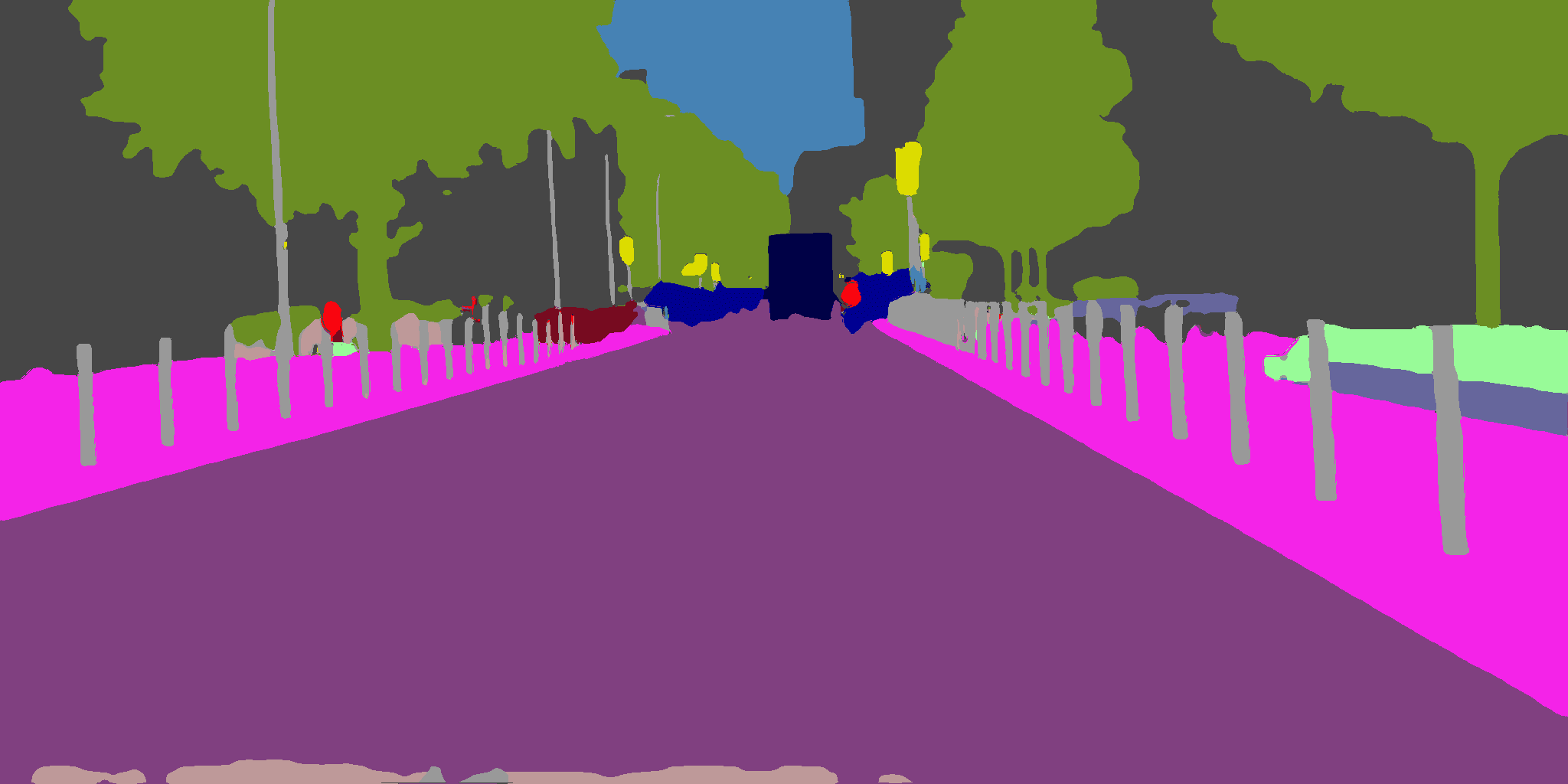}
	\\
  	\includegraphics[height=\cityheight, width=\imwidth, align=b]{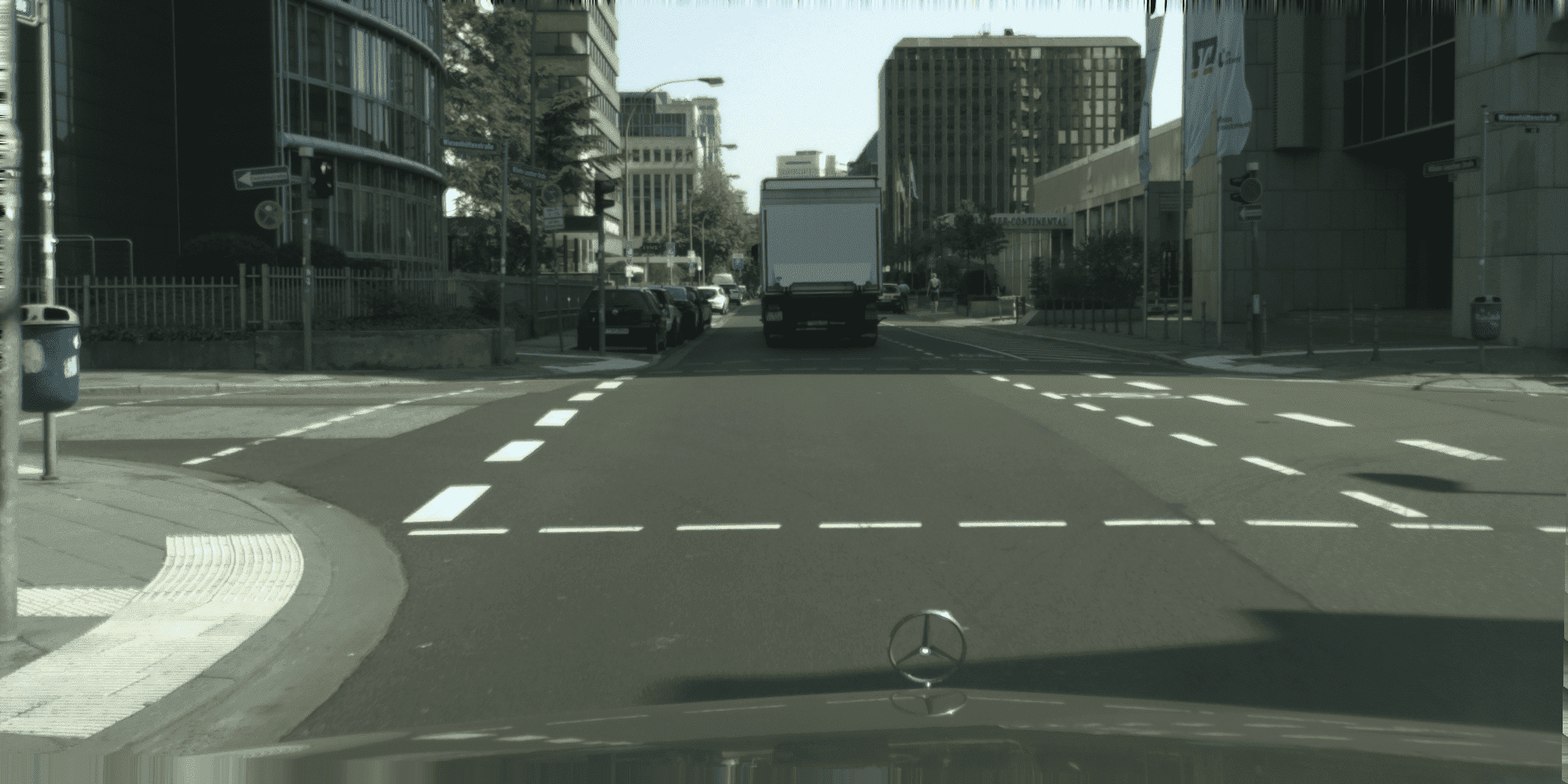}
	\includegraphics[height=\cityheight, width=\imwidth, align=b]{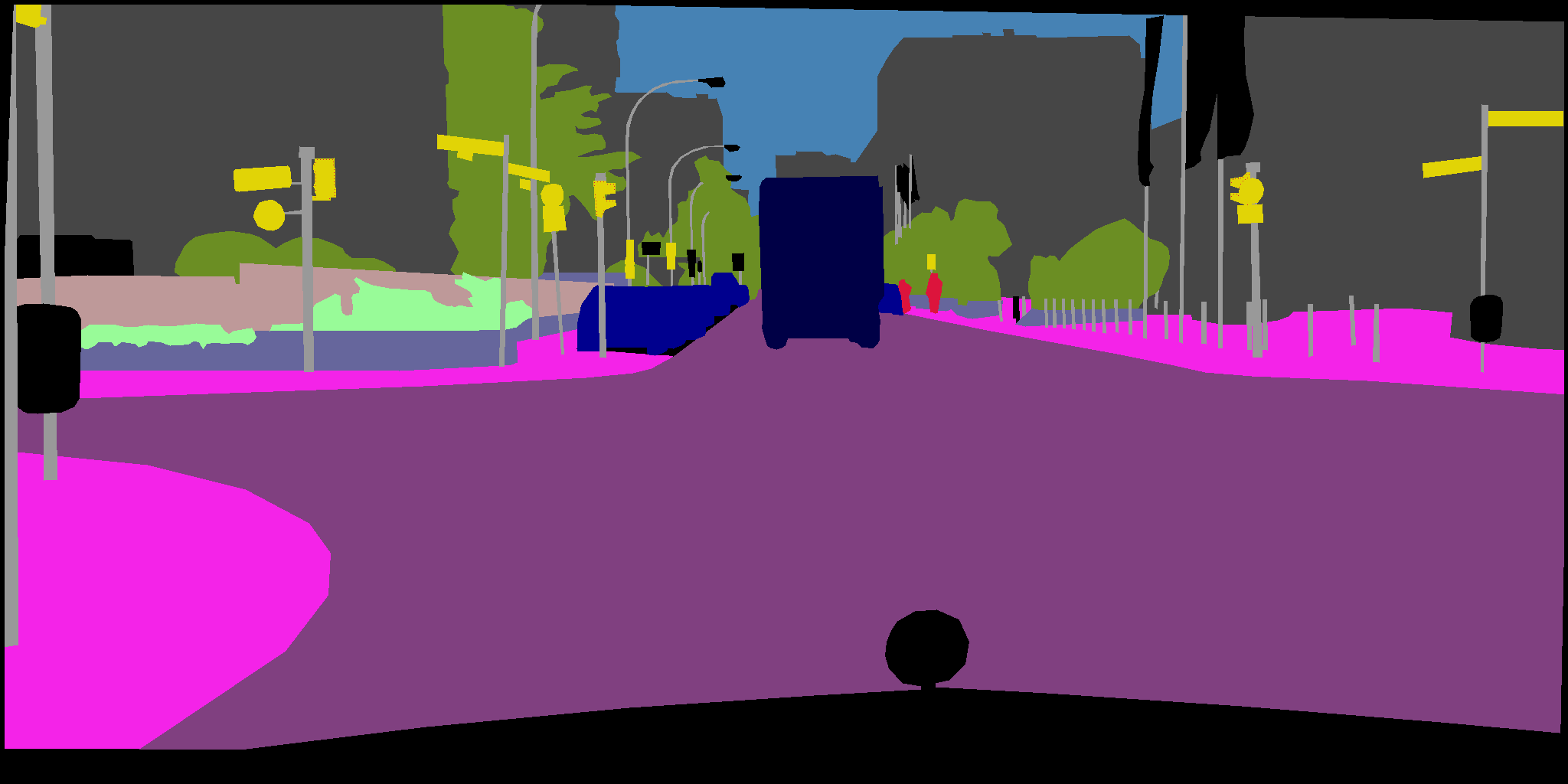}
	\includegraphics[height=\cityheight, width=\imwidth, align=b]{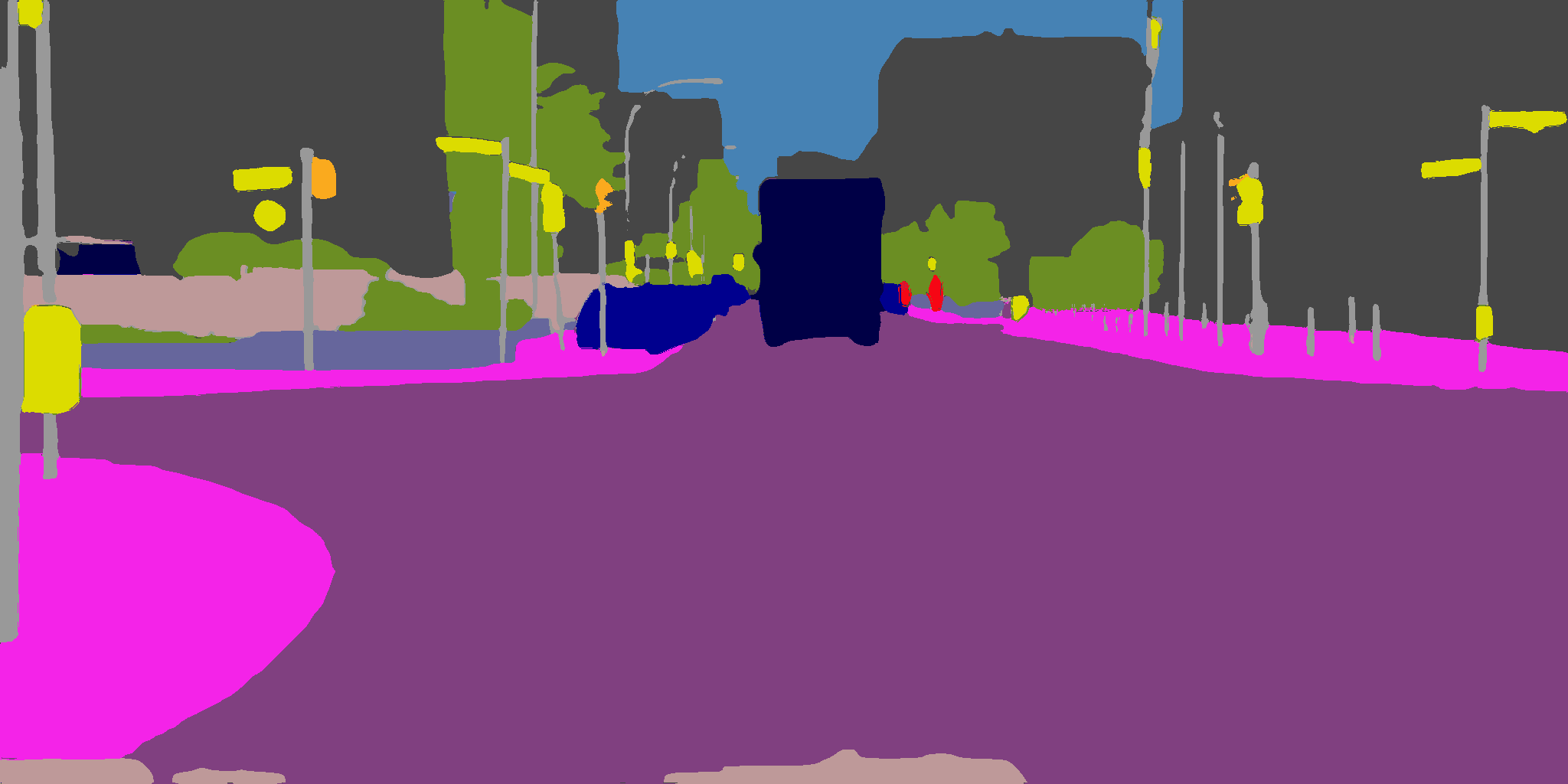}
	\\
  	\includegraphics[height=\cityheight, width=\imwidth, align=b]{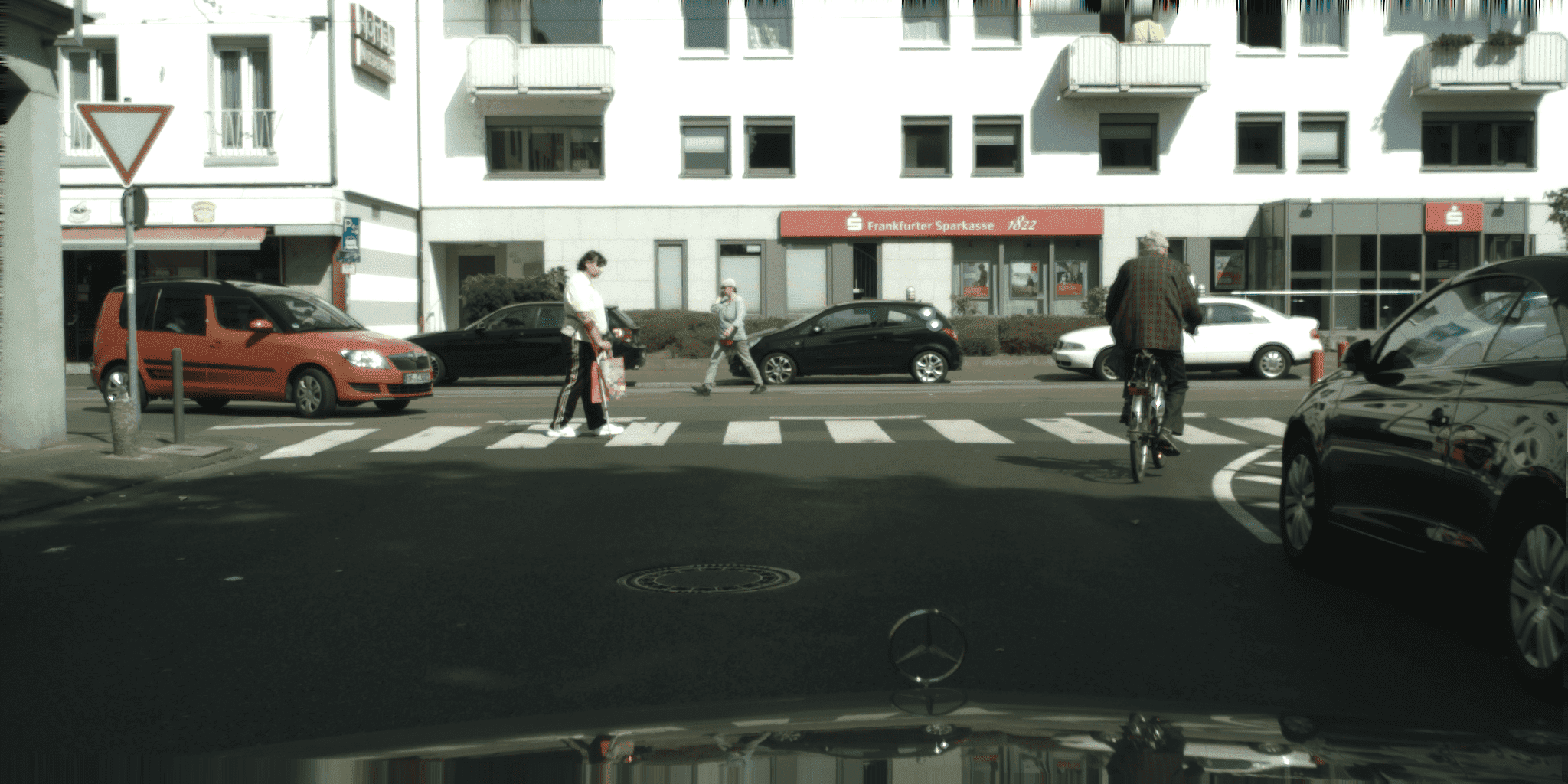}
	\includegraphics[height=\cityheight, width=\imwidth, align=b]{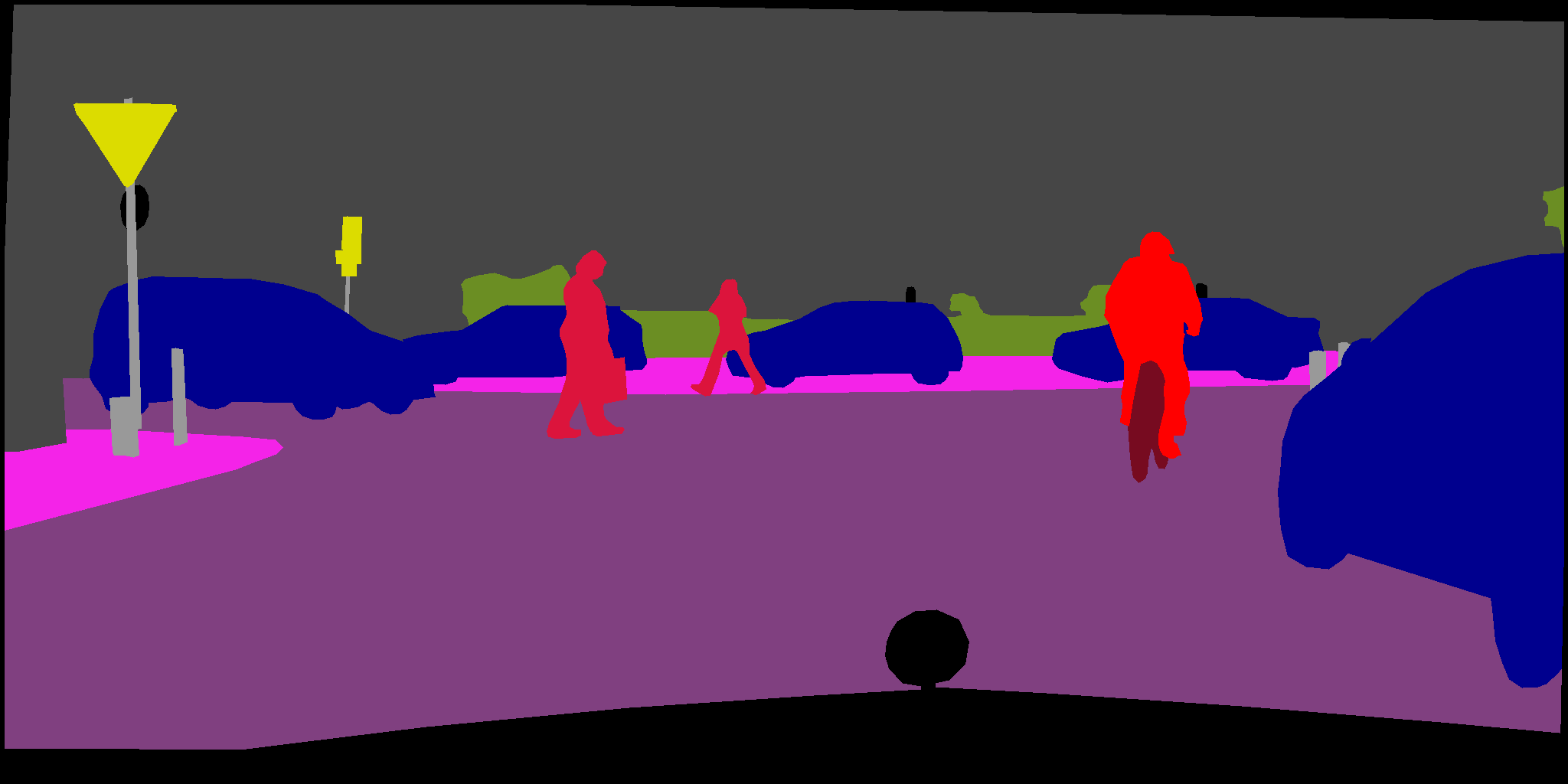}
	\includegraphics[height=\cityheight, width=\imwidth, align=b]{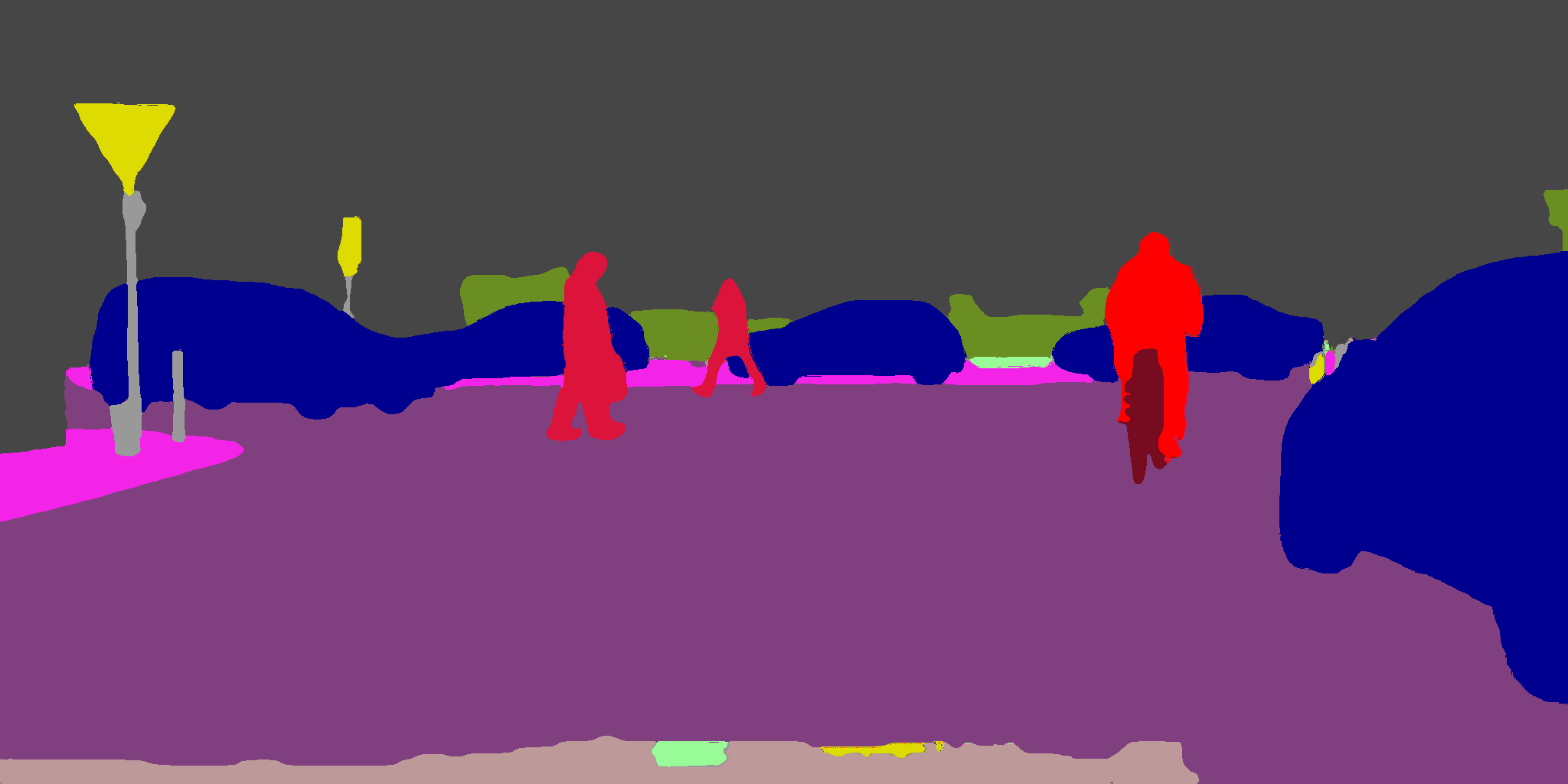}
	\\
  	\includegraphics[height=\cityheight, width=\imwidth, align=b]{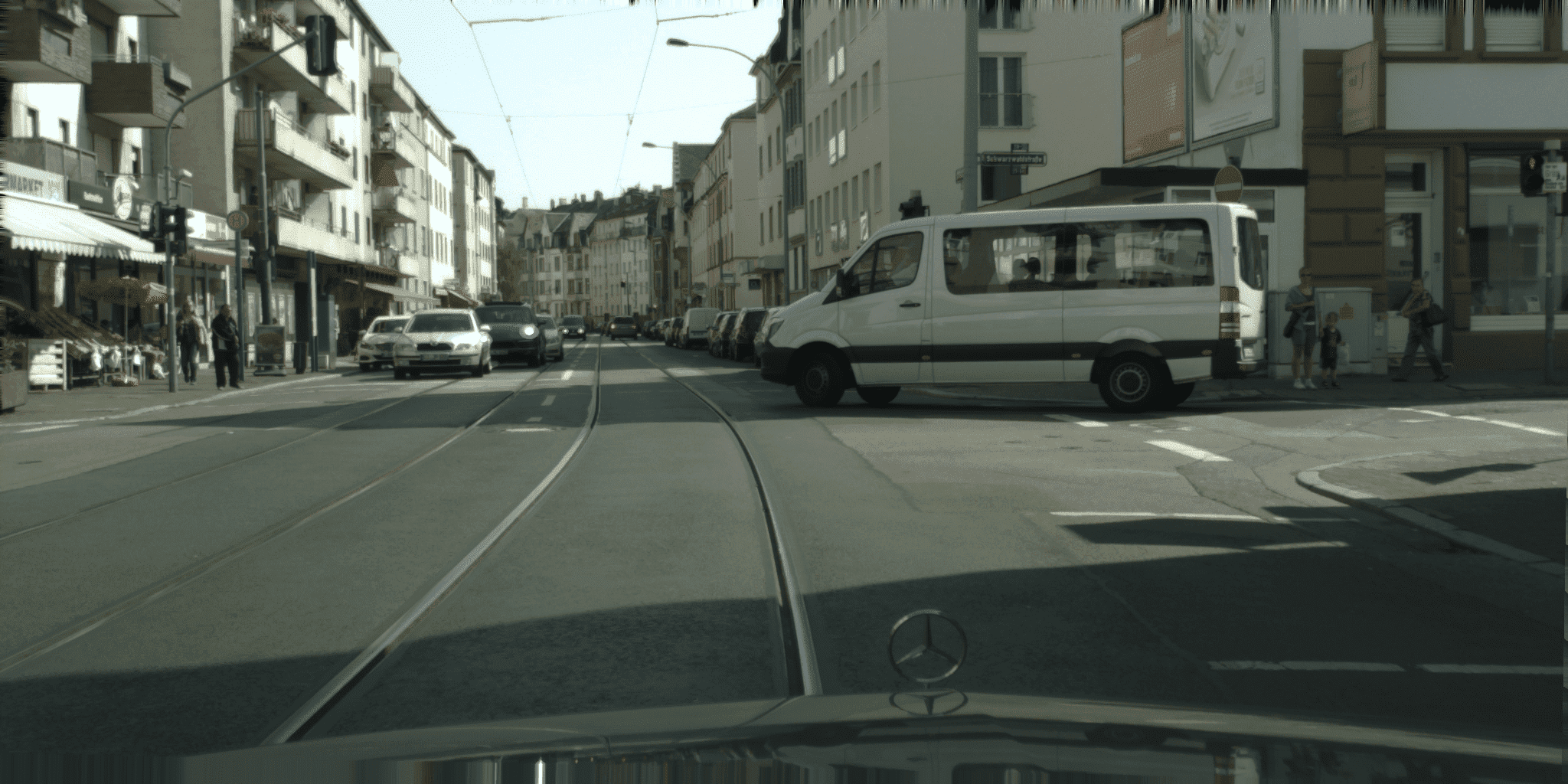}
	\includegraphics[height=\cityheight, width=\imwidth, align=b]{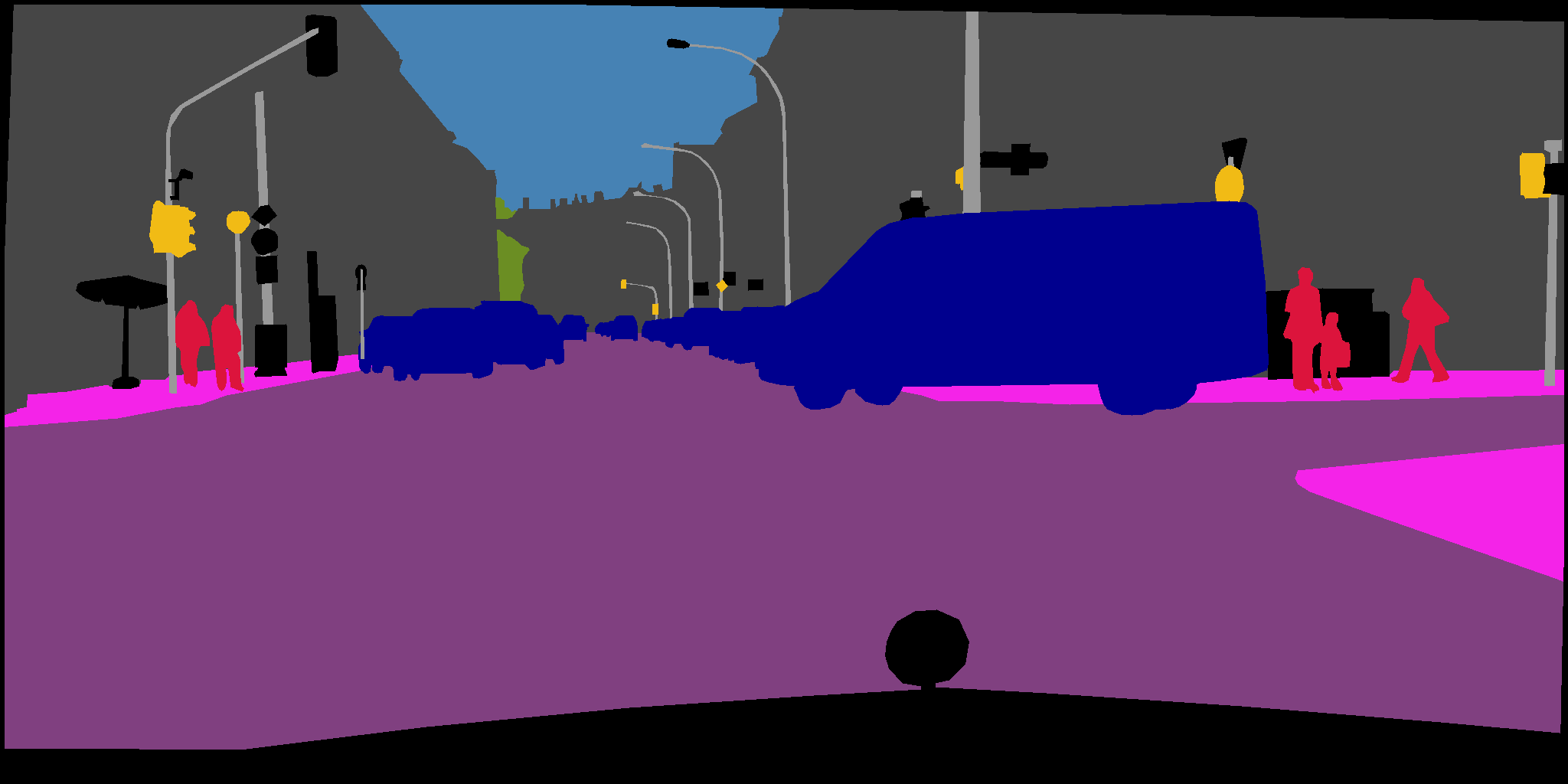}
	\includegraphics[height=\cityheight, width=\imwidth, align=b]{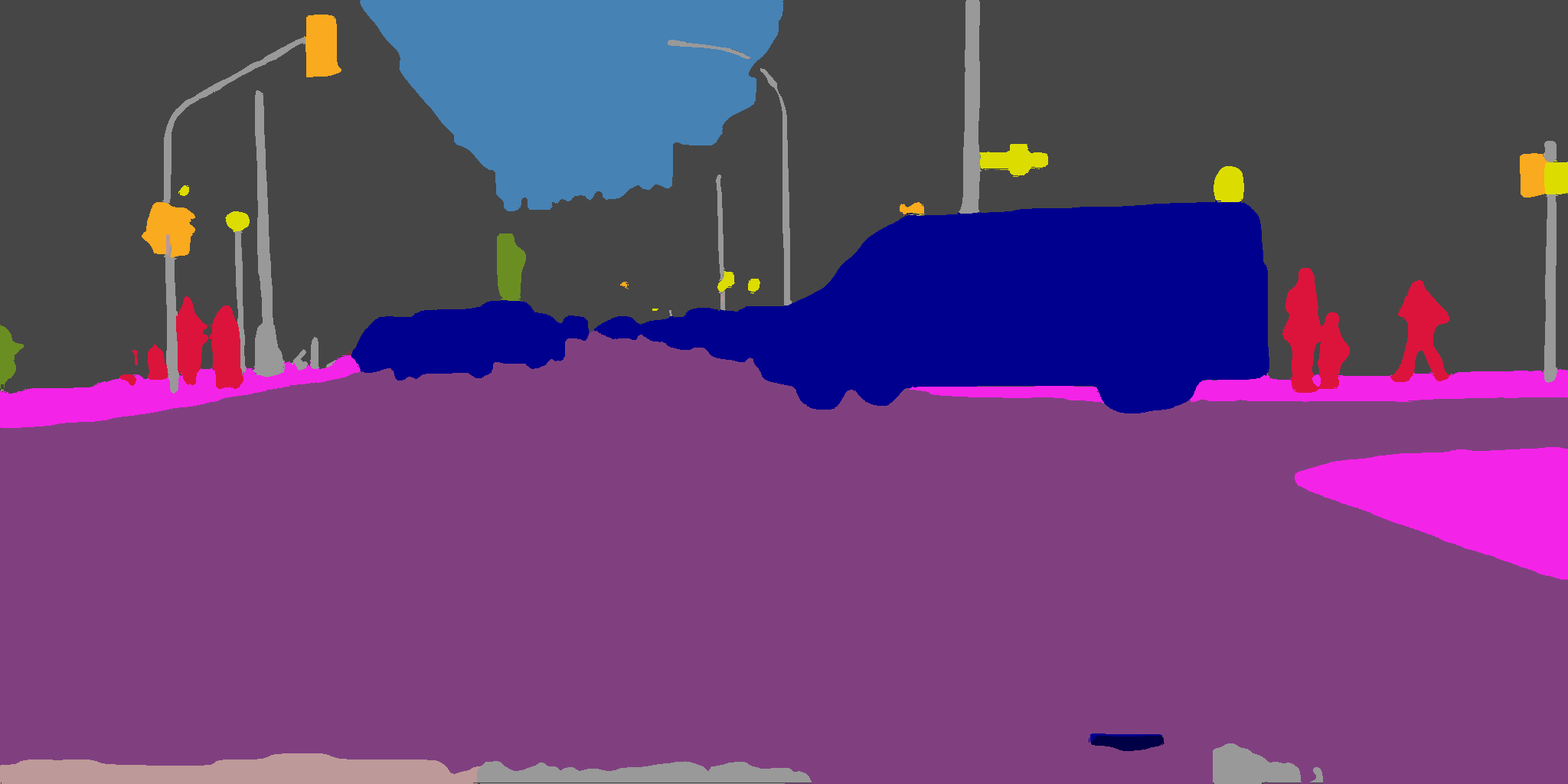}
	\\
	\rotatebox{0}{\textcolor{white}{---}Image}
	\rotatebox{0}{\textcolor{white}{------------------------------------}Ground Truth}
	\rotatebox{0}{\textcolor{white}{----------------------------------}Prediction}
	\caption{Qualitative results of semantic segmentation on  Cityscapes {\tt val} split~\cite{cordts2016cityscapes}.
	}
	\label{fig:supp_visualization_stage_2_cityscapes}
\end{figure*}

\begin{figure*}[th]
	\def \imwidth {5.6cm}
	\def \adeheight {3.6cm}
	\centering

	\includegraphics[height=\adeheight, width=\imwidth, align=b]{figures/supp_visualization_ade20k/img/ADE_val_00000086.jpg}
	\includegraphics[height=\adeheight, width=\imwidth, align=b]{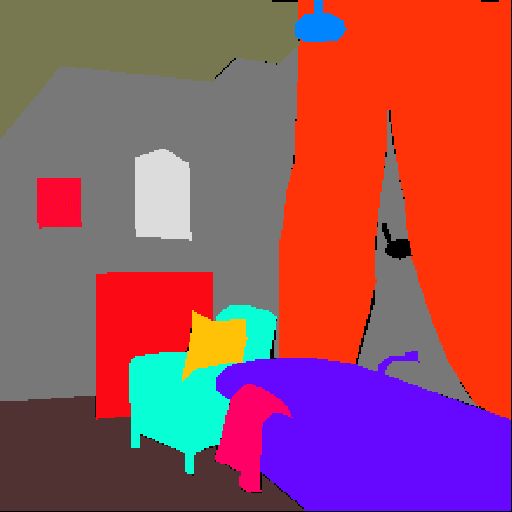}
	\includegraphics[height=\adeheight, width=\imwidth, align=b]{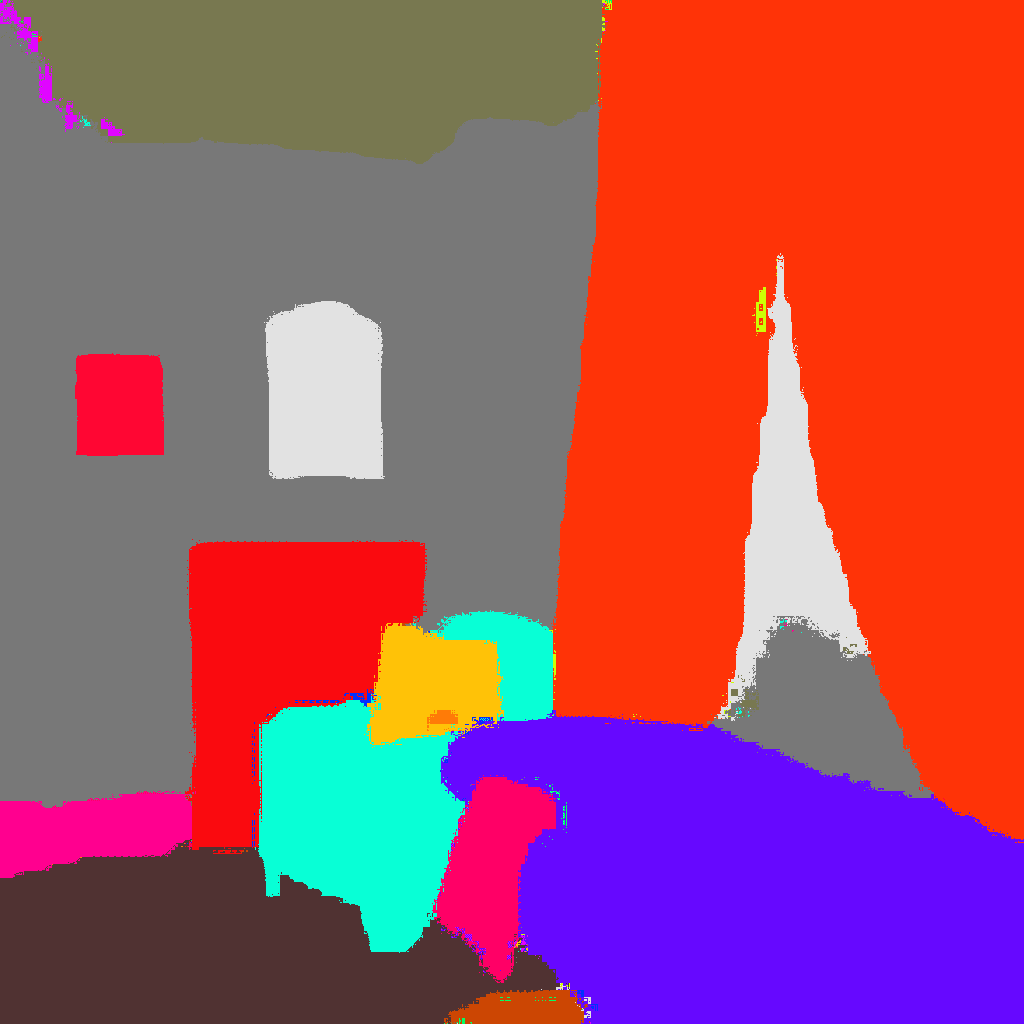}\\
	\includegraphics[height=\adeheight, width=\imwidth, align=b]{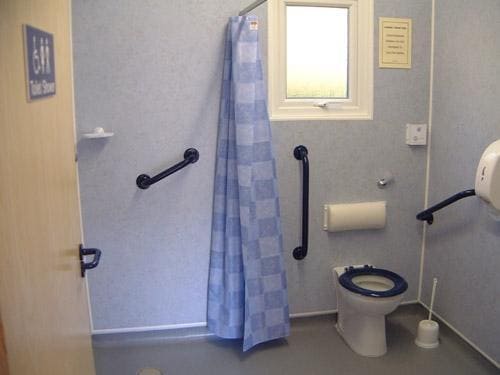}
	\includegraphics[height=\adeheight, width=\imwidth, align=b]{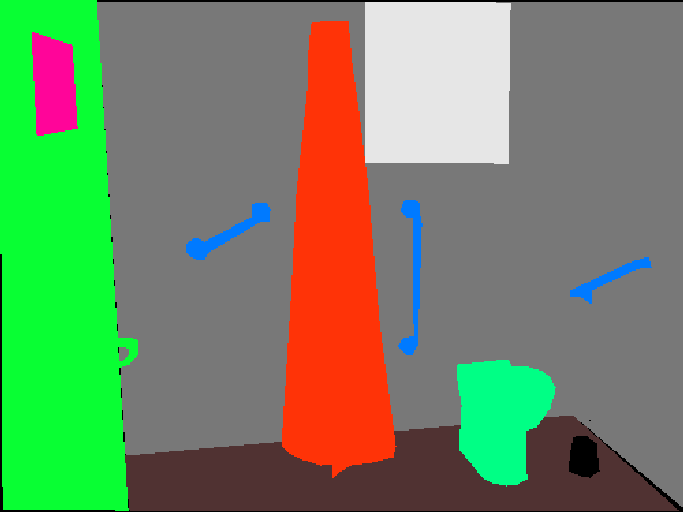}
	\includegraphics[height=\adeheight, width=\imwidth, align=b]{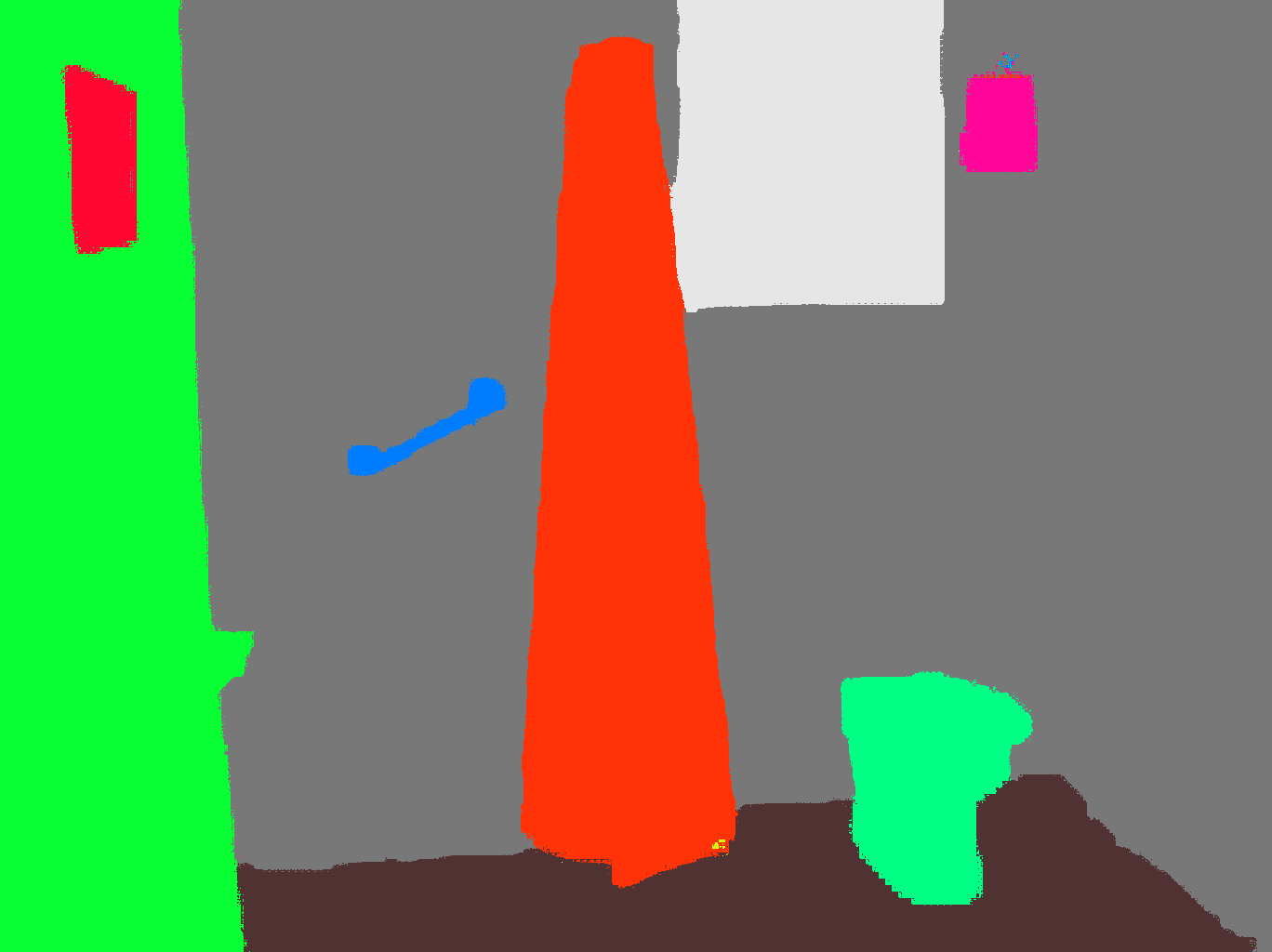}
	\\
 	\includegraphics[height=\adeheight, width=\imwidth, align=b]{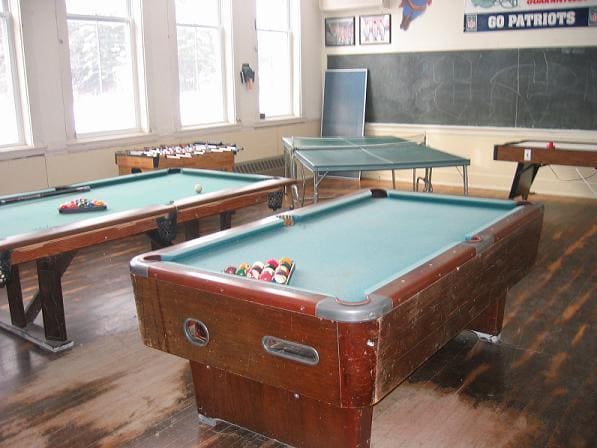}
	\includegraphics[height=\adeheight, width=\imwidth, align=b]{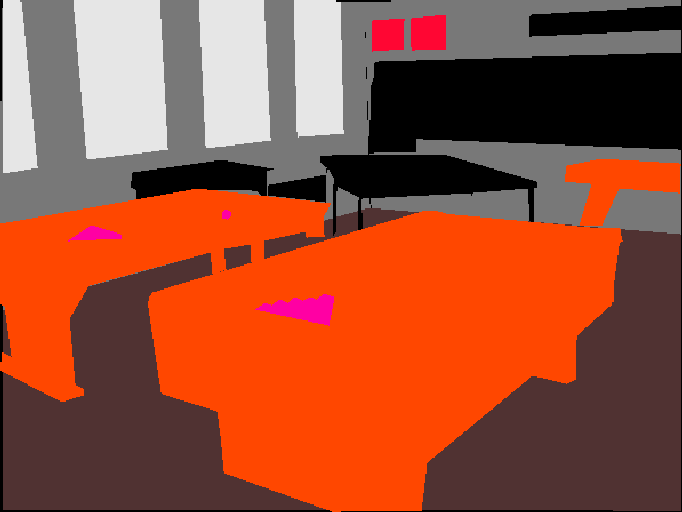}
	\includegraphics[height=\adeheight, width=\imwidth, align=b]{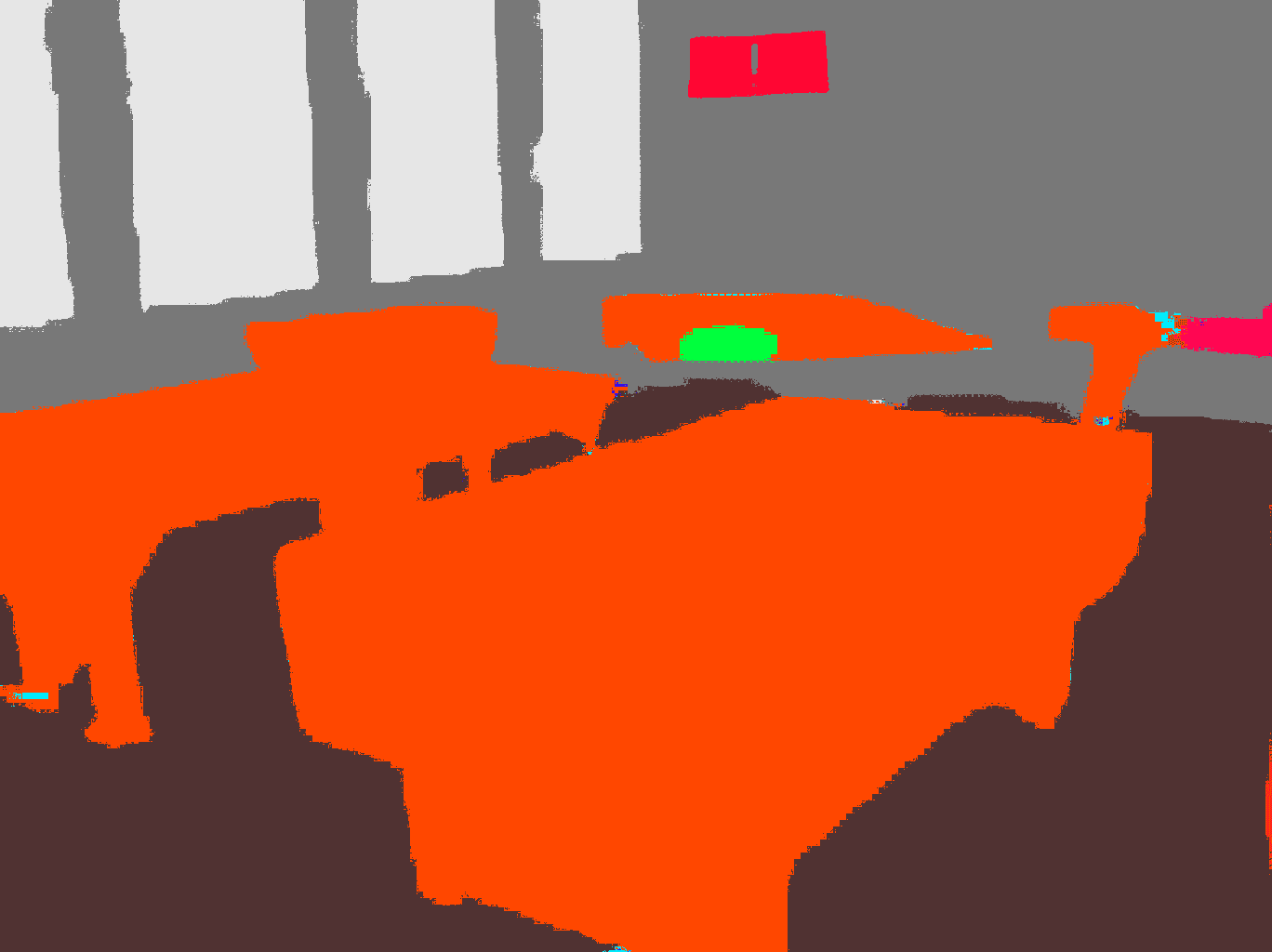}\\
	\includegraphics[height=\adeheight, width=\imwidth, align=b]{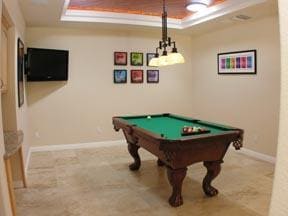}
	\includegraphics[height=\adeheight, width=\imwidth, align=b]{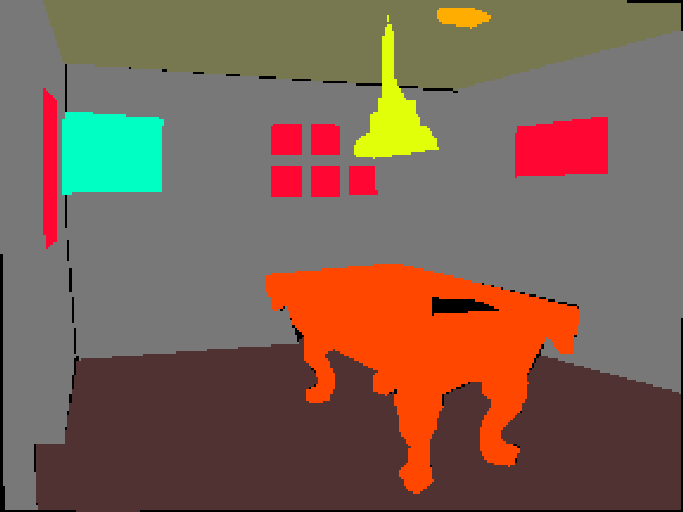}
	\includegraphics[height=\adeheight, width=\imwidth, align=b]{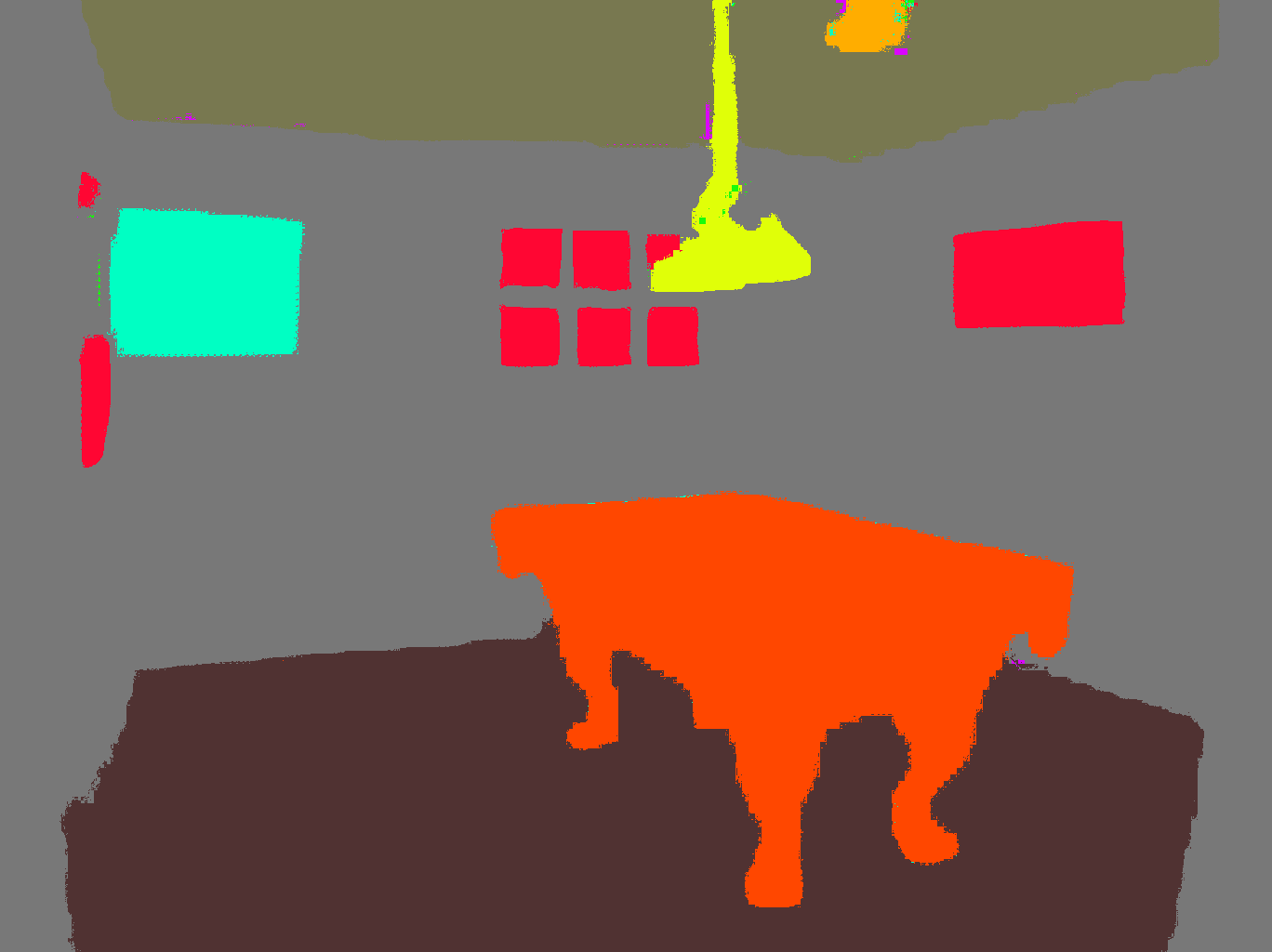}
	\\
	\includegraphics[height=\adeheight, width=\imwidth, align=b]{figures/supp_visualization_ade20k/img/ADE_val_00001135.jpg}
	\includegraphics[height=\adeheight, width=\imwidth, align=b]{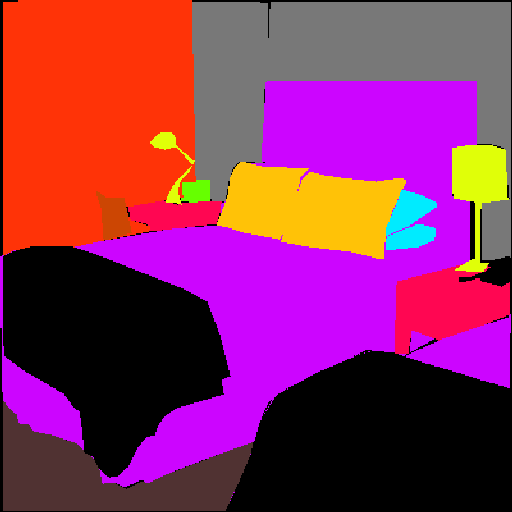}
	\includegraphics[height=\adeheight, width=\imwidth, align=b]{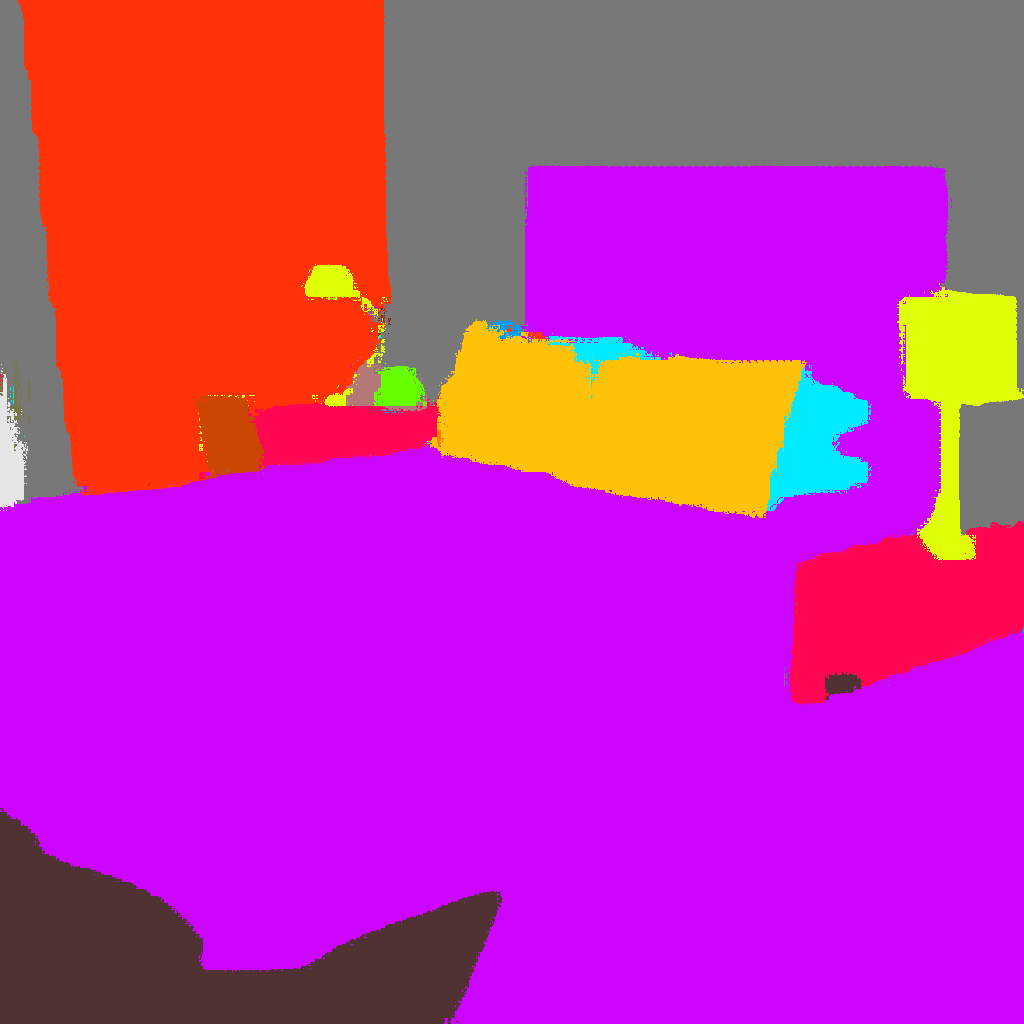}\\
	\includegraphics[height=\adeheight, width=\imwidth, align=b]{figures/supp_visualization_ade20k/img/ADE_val_00001282.jpg}
	\includegraphics[height=\adeheight, width=\imwidth, align=b]{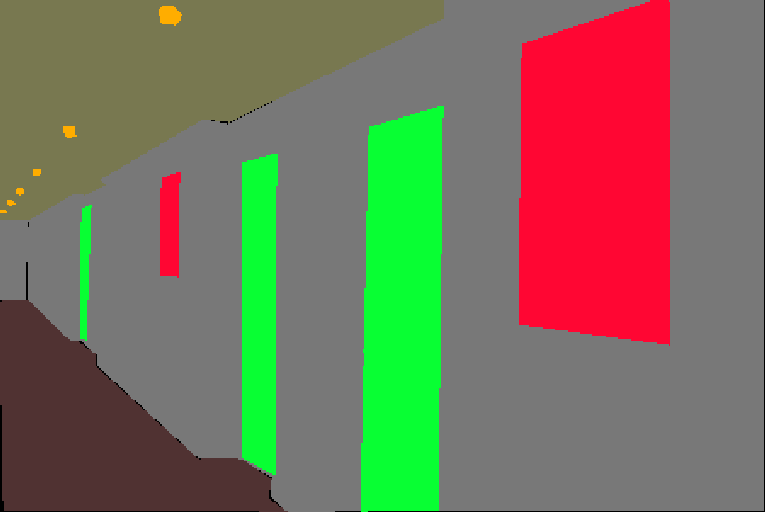}
	\includegraphics[height=\adeheight, width=\imwidth, align=b]{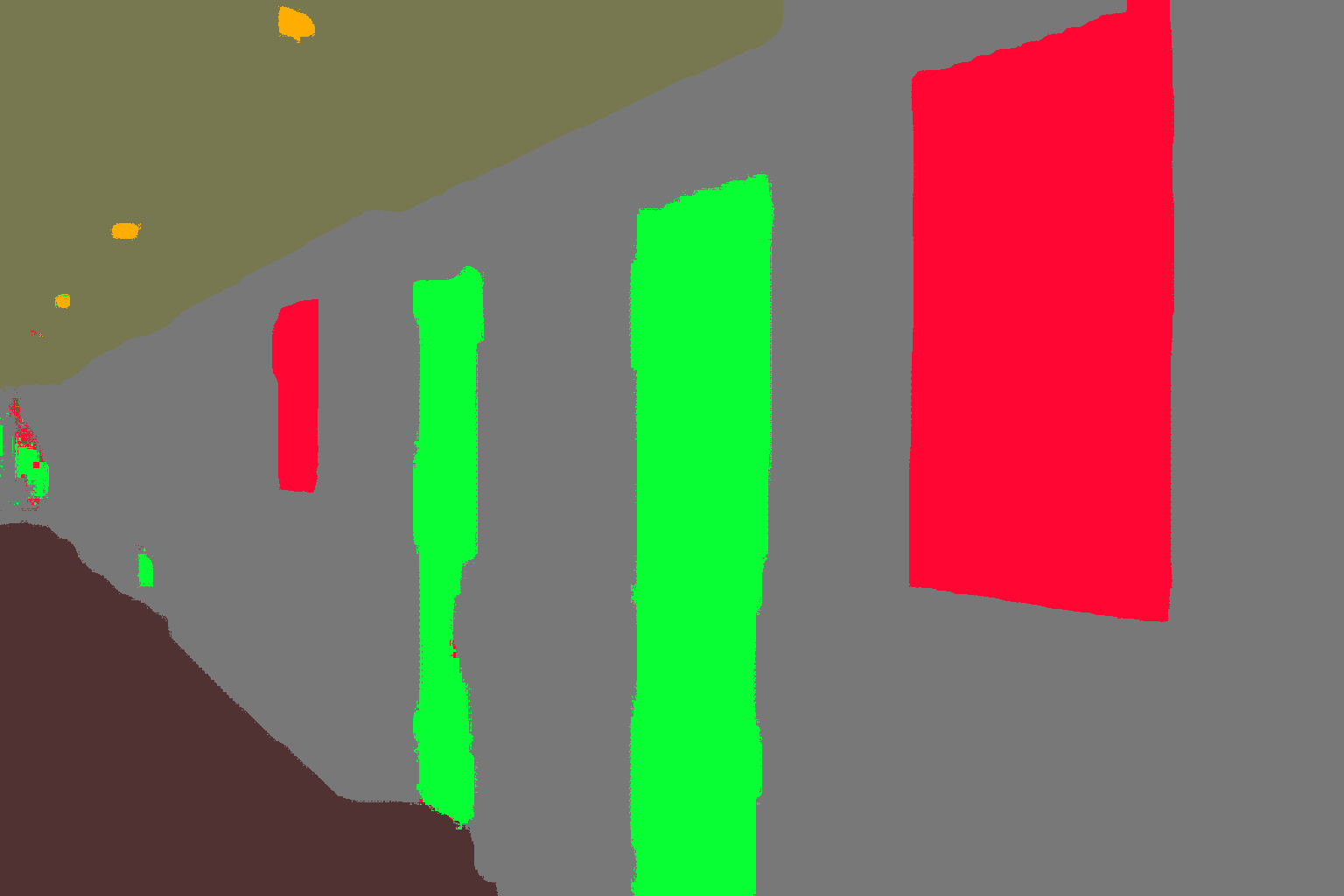}
	\rotatebox{0}{\textcolor{white}{------}Image}
	\rotatebox{0}{\textcolor{white}{--------------------------------------}Ground Truth}
	\rotatebox{0}{\textcolor{white}{--------------------------------}Prediction}
	\caption{Qualitative results of semantic segmentation on  ADE20K {\tt val} split~\cite{zhou2019semantic}.
	}
	\label{fig:supp_visualization_stage_2_ade20k}
\end{figure*}

\begin{figure*}[th]
	\def \imwidth {2.8cm}
    \def \vochight {2.0cm}
	\def \adeheight {2.0cm}
	\def \cityheight {1.5cm}
    \def \contexthight {2.0cm}
	\centering
	\includegraphics[height=\vochight, width=\imwidth, align=b]{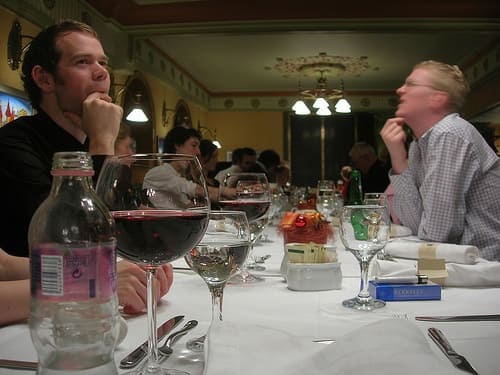}
	\includegraphics[height=\vochight, width=\imwidth, align=b]{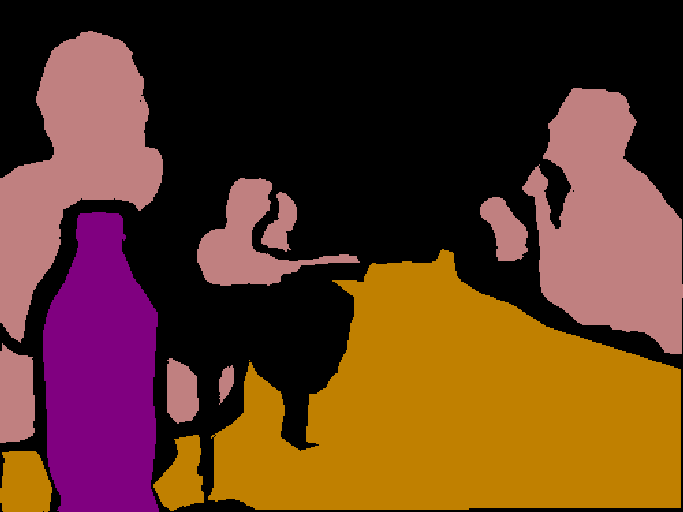}
	\includegraphics[height=\vochight, width=\imwidth, align=b]{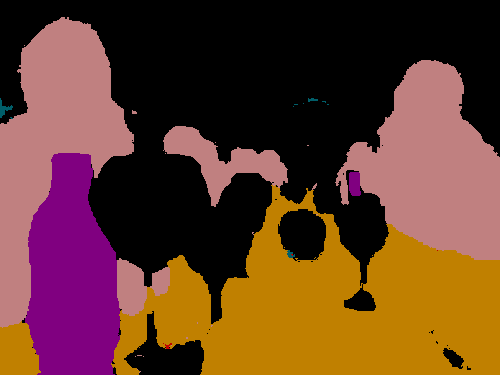}
	\includegraphics[height=\vochight, width=\imwidth, align=b]{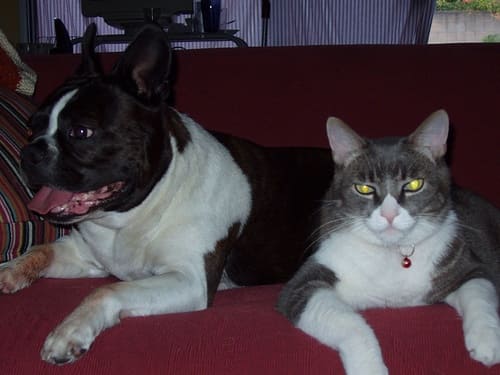}
	\includegraphics[height=\vochight, width=\imwidth, align=b]{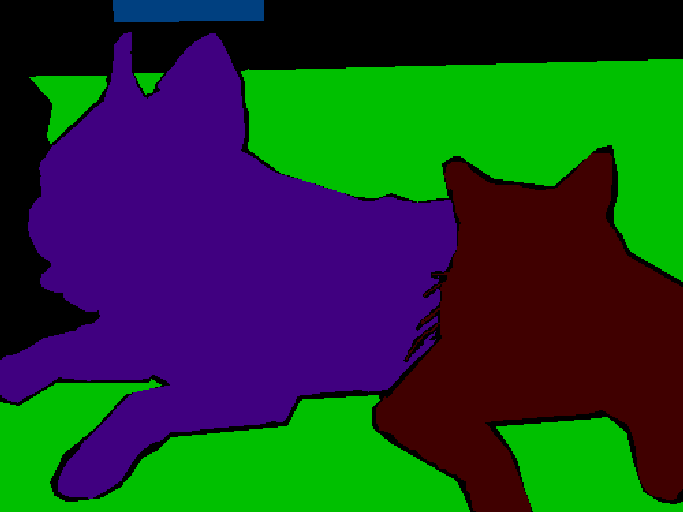}
	\includegraphics[height=\vochight, width=\imwidth, align=b]{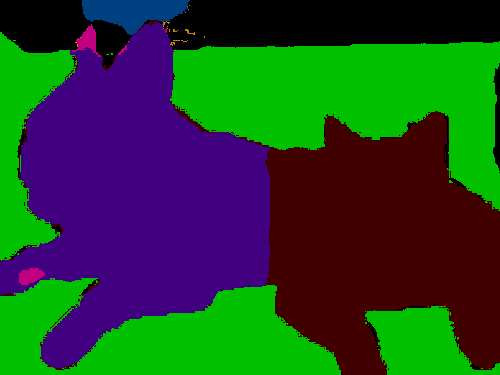}
	\\
  	\includegraphics[height=\contexthight, width=\imwidth, align=b]{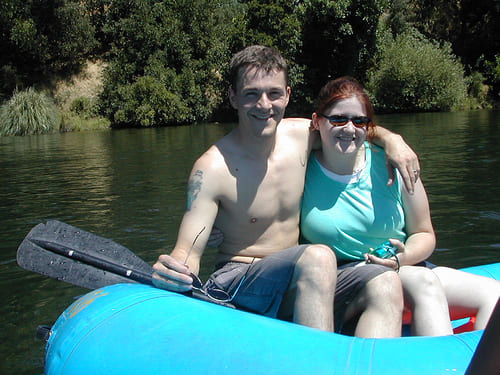}
	\includegraphics[height=\contexthight, width=\imwidth, align=b]{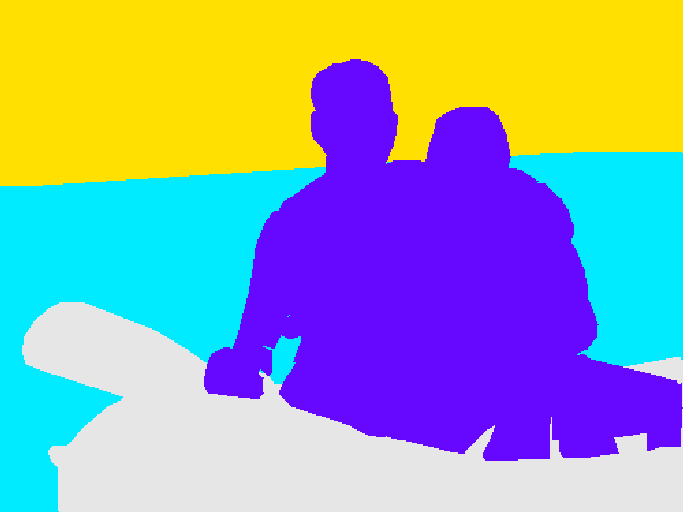}
	\includegraphics[height=\contexthight, width=\imwidth, align=b]{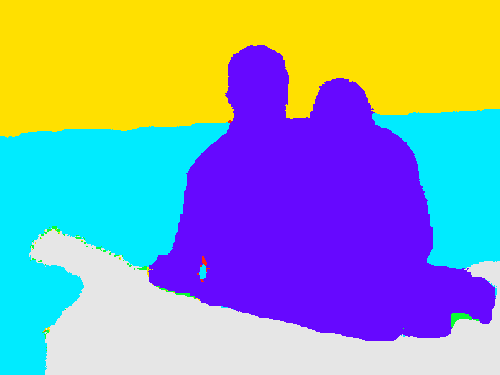}
 	\includegraphics[height=\contexthight, width=\imwidth, align=b]{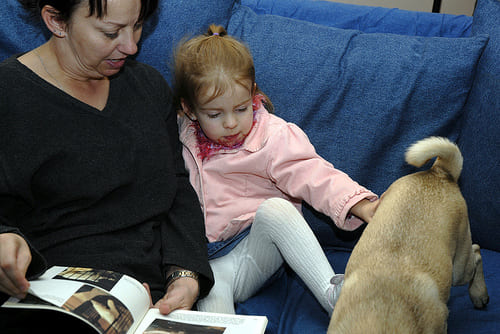}
	\includegraphics[height=\contexthight, width=\imwidth, align=b]{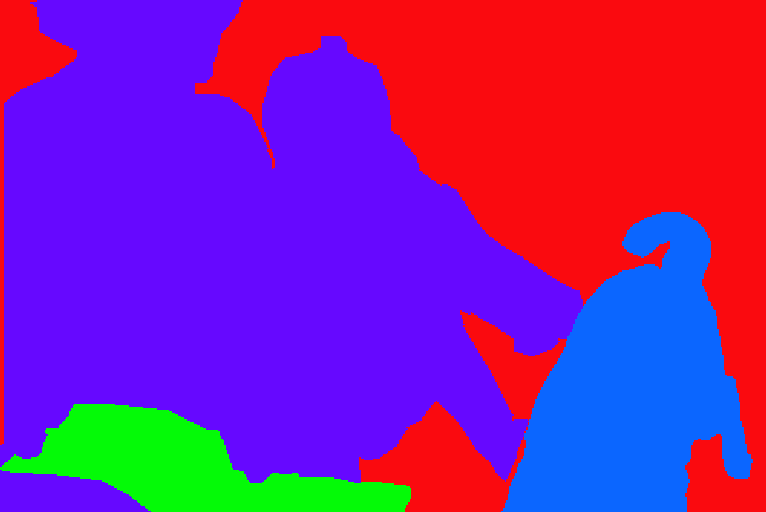}
	\includegraphics[height=\contexthight, width=\imwidth, align=b]{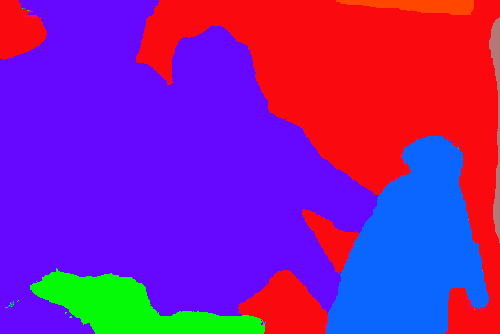}
	\\
	\includegraphics[height=\adeheight, width=\imwidth, align=b]{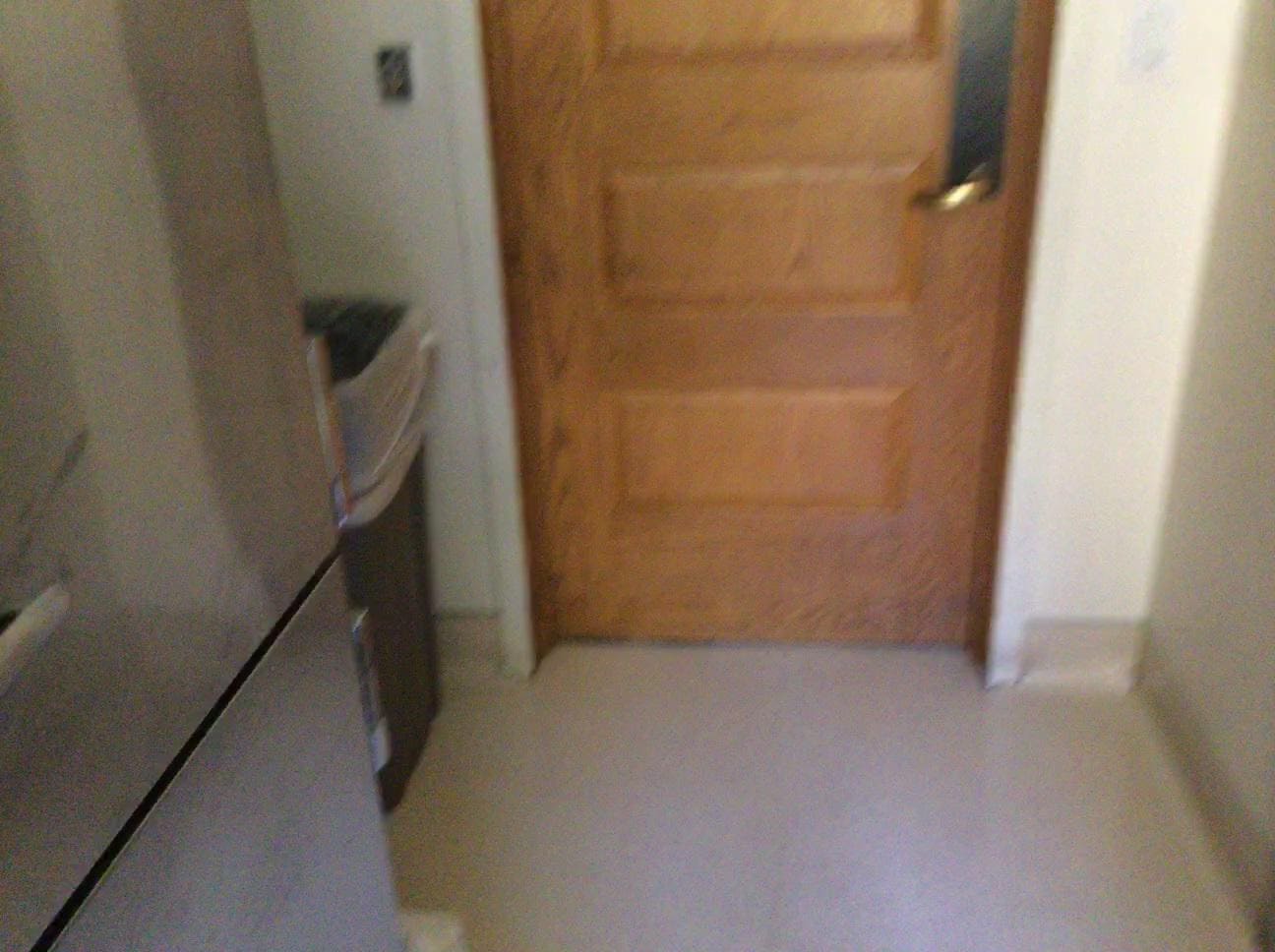}
	\includegraphics[height=\adeheight, width=\imwidth, align=b]{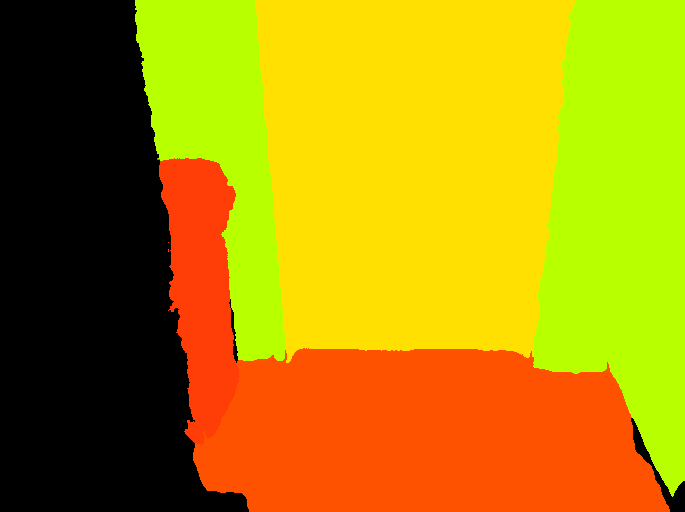}
	\includegraphics[height=\adeheight, width=\imwidth, align=b]{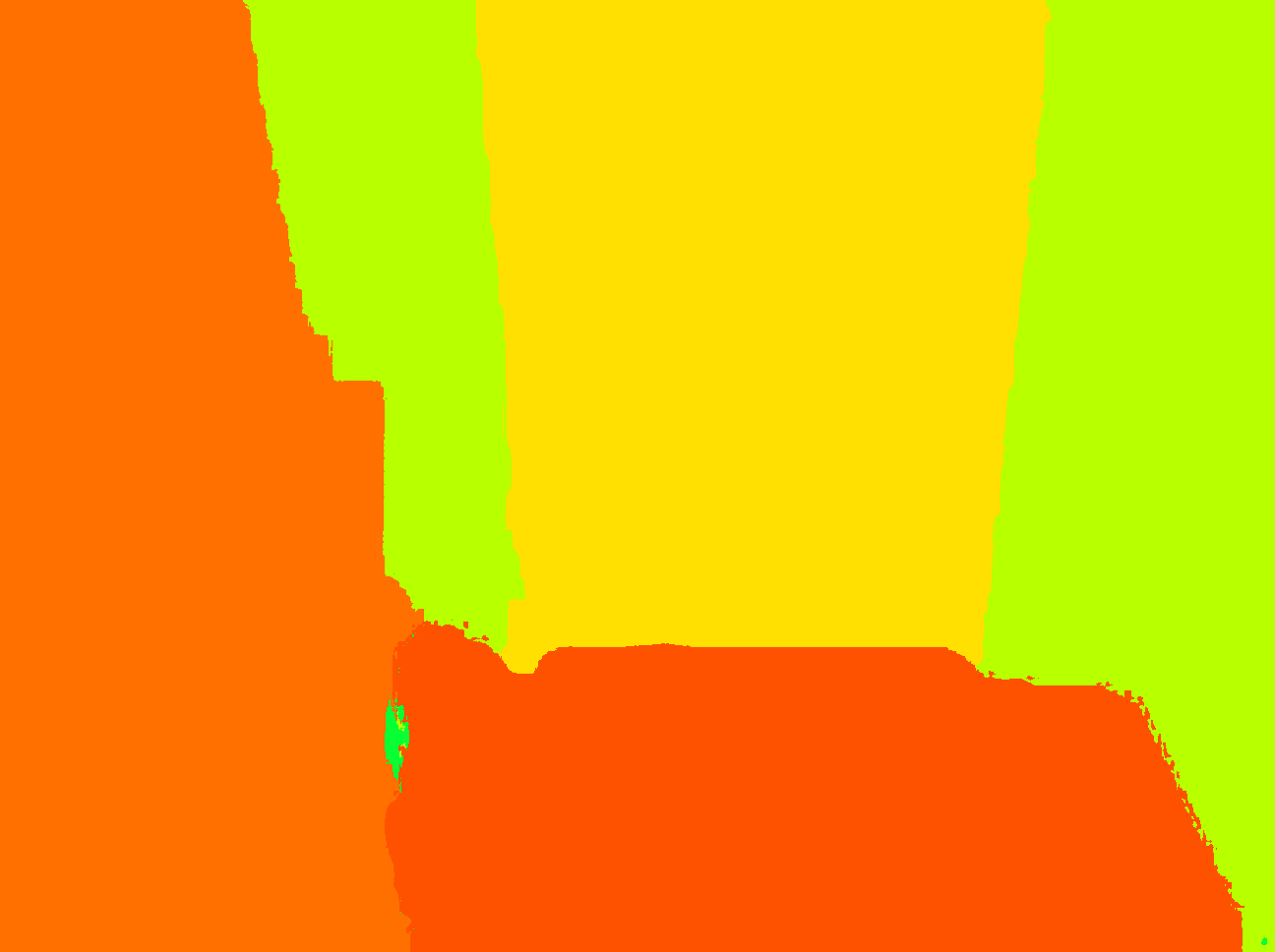}
	\includegraphics[height=\adeheight, width=\imwidth, align=b]{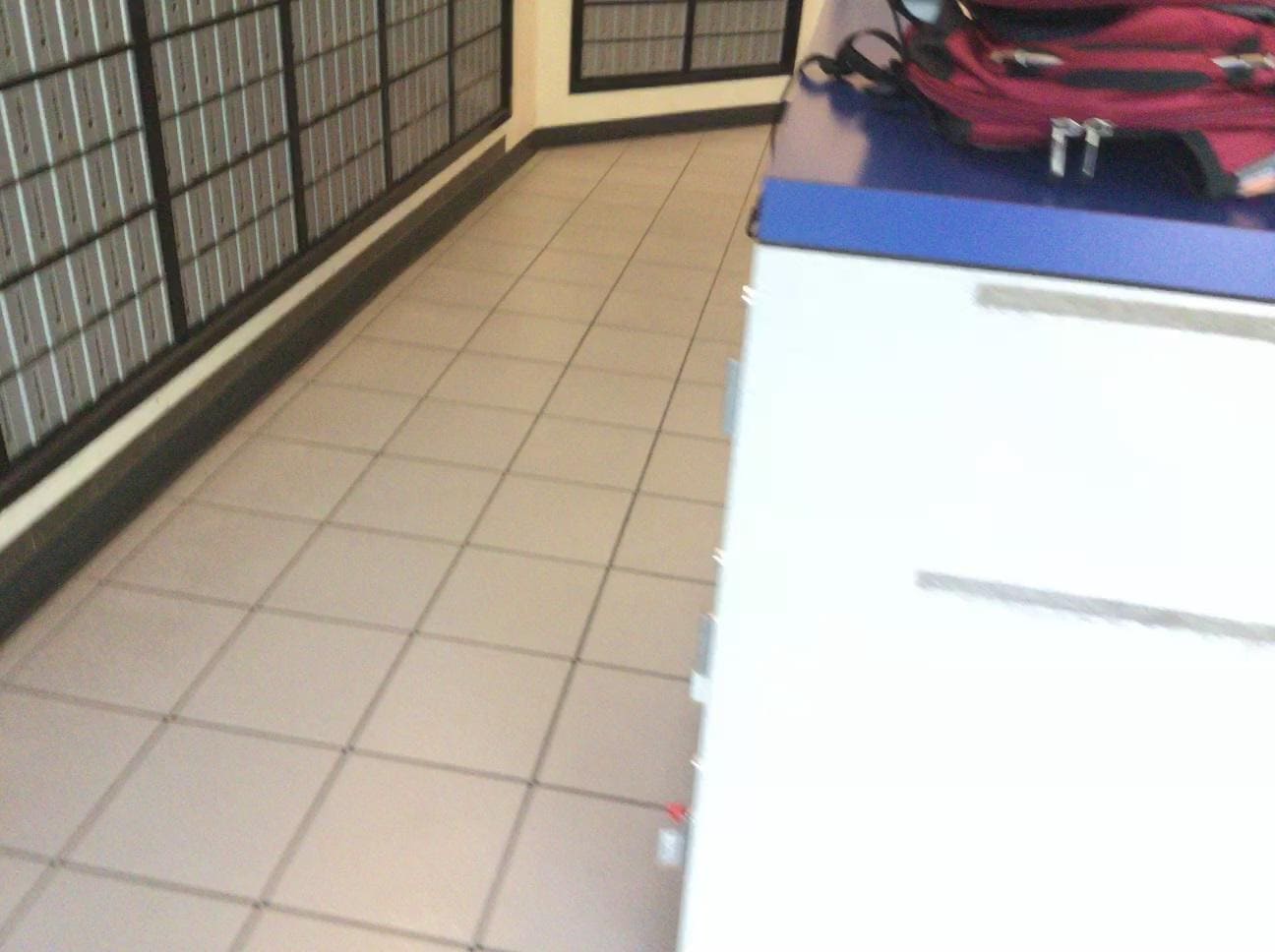}
	\includegraphics[height=\adeheight, width=\imwidth, align=b]{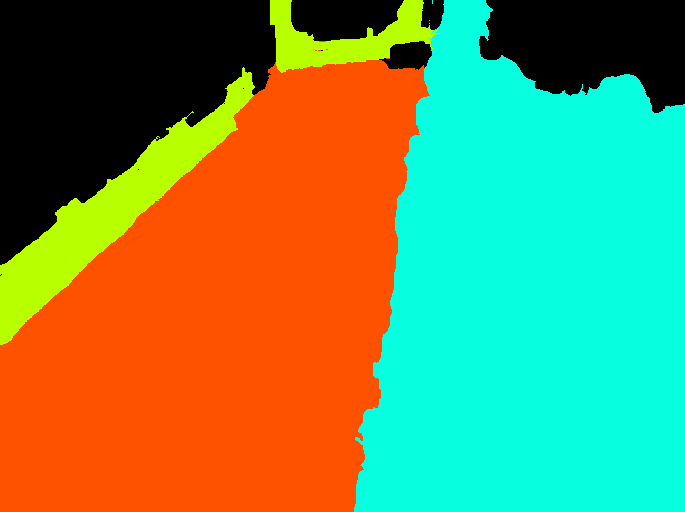}
	\includegraphics[height=\adeheight, width=\imwidth, align=b]{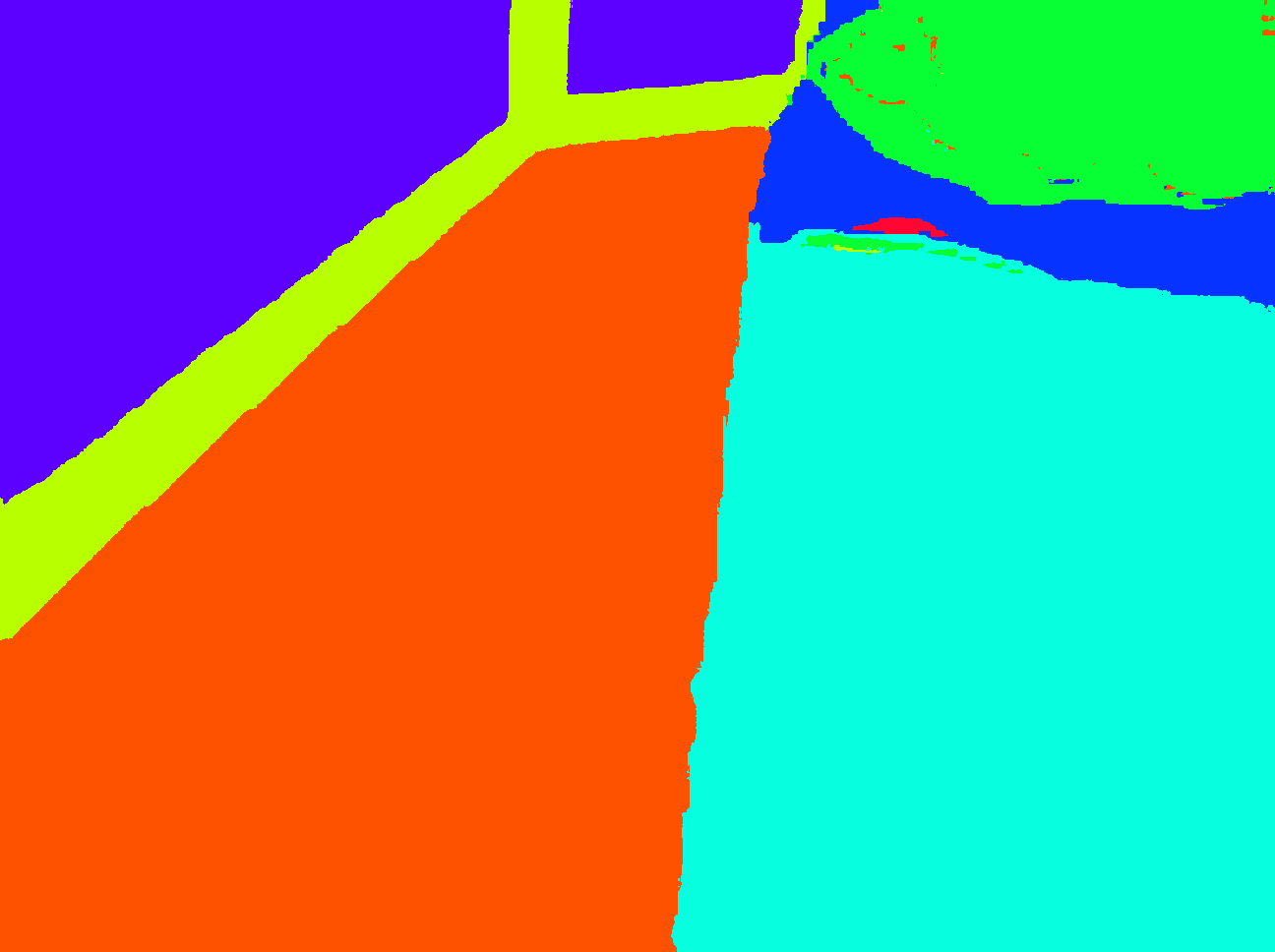}
	\\
 	\includegraphics[height=\cityheight, width=\imwidth, align=b]{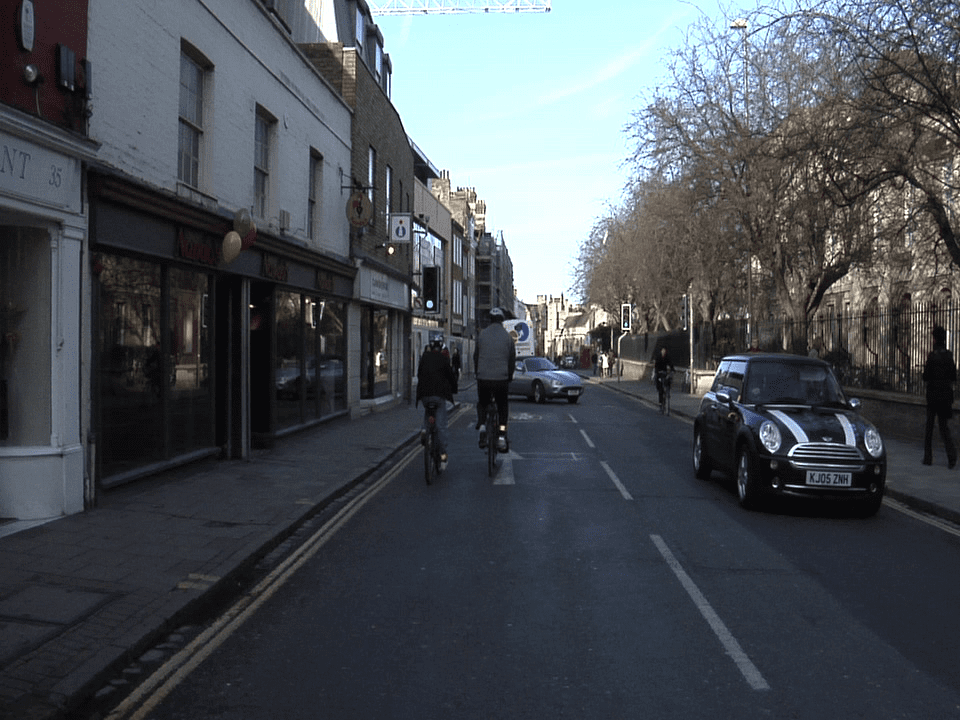}
	\includegraphics[height=\cityheight, width=\imwidth, align=b]{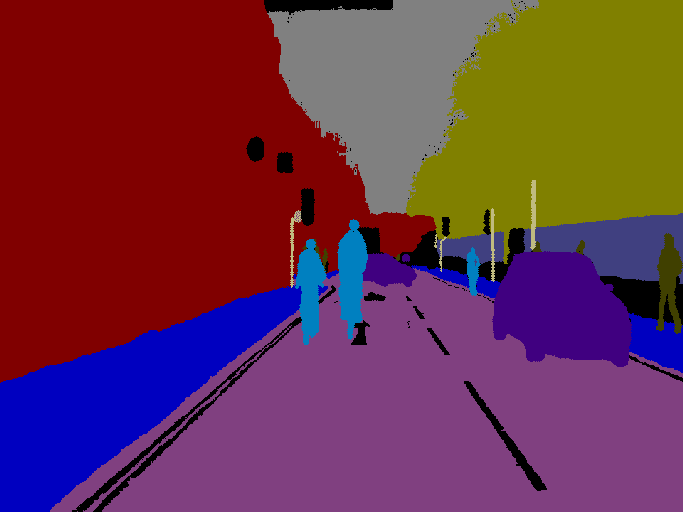}
	\includegraphics[height=\cityheight, width=\imwidth, align=b]{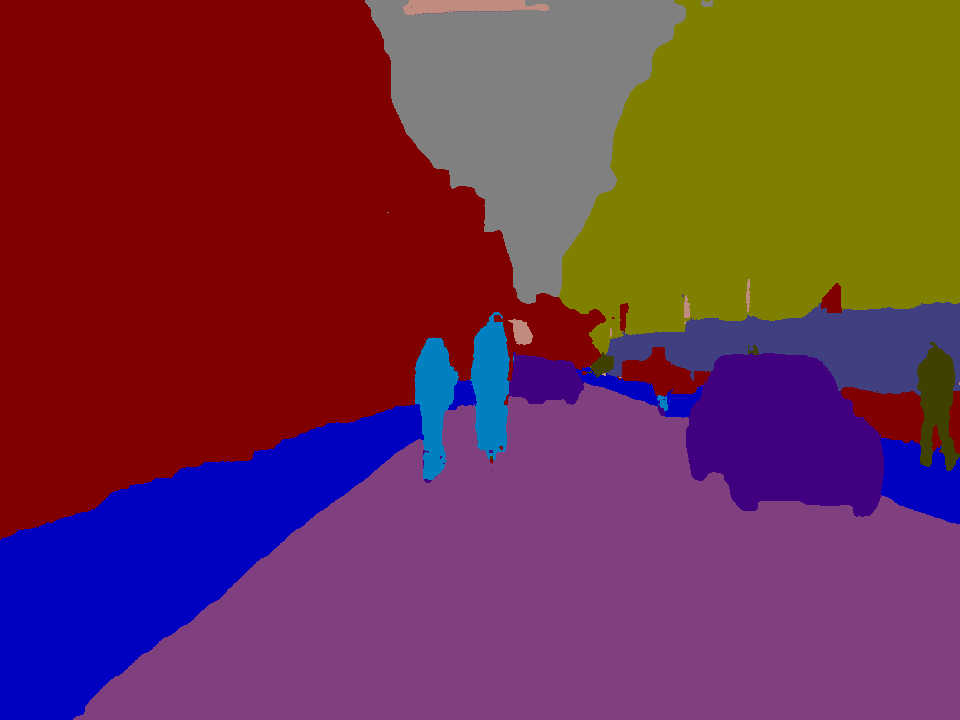}
	\includegraphics[height=\cityheight, width=\imwidth, align=b]{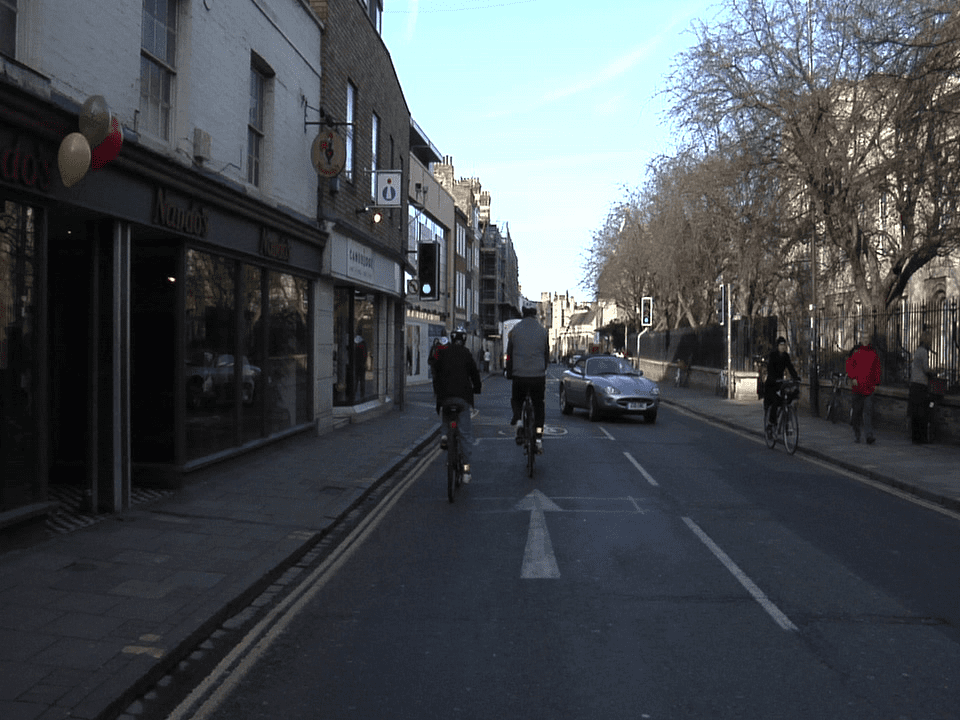}
	\includegraphics[height=\cityheight, width=\imwidth, align=b]{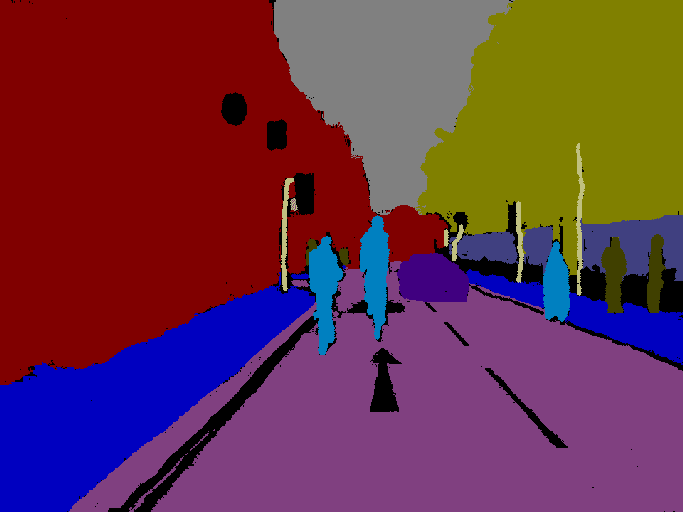}
	\includegraphics[height=\cityheight, width=\imwidth, align=b]{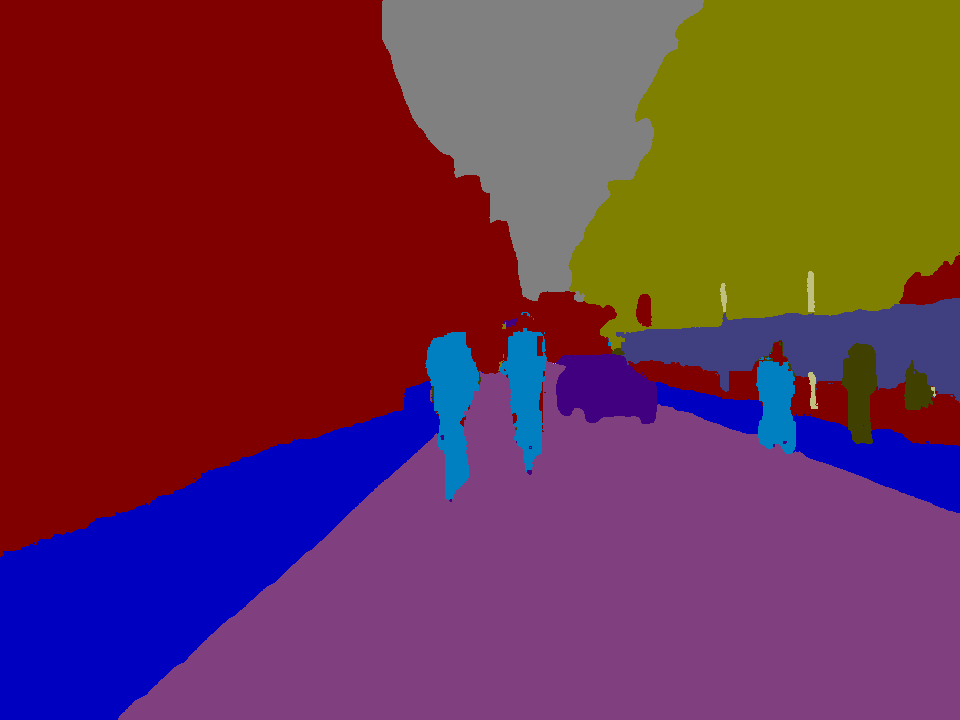}
	\\
  	\includegraphics[height=\cityheight, width=\imwidth, align=b]{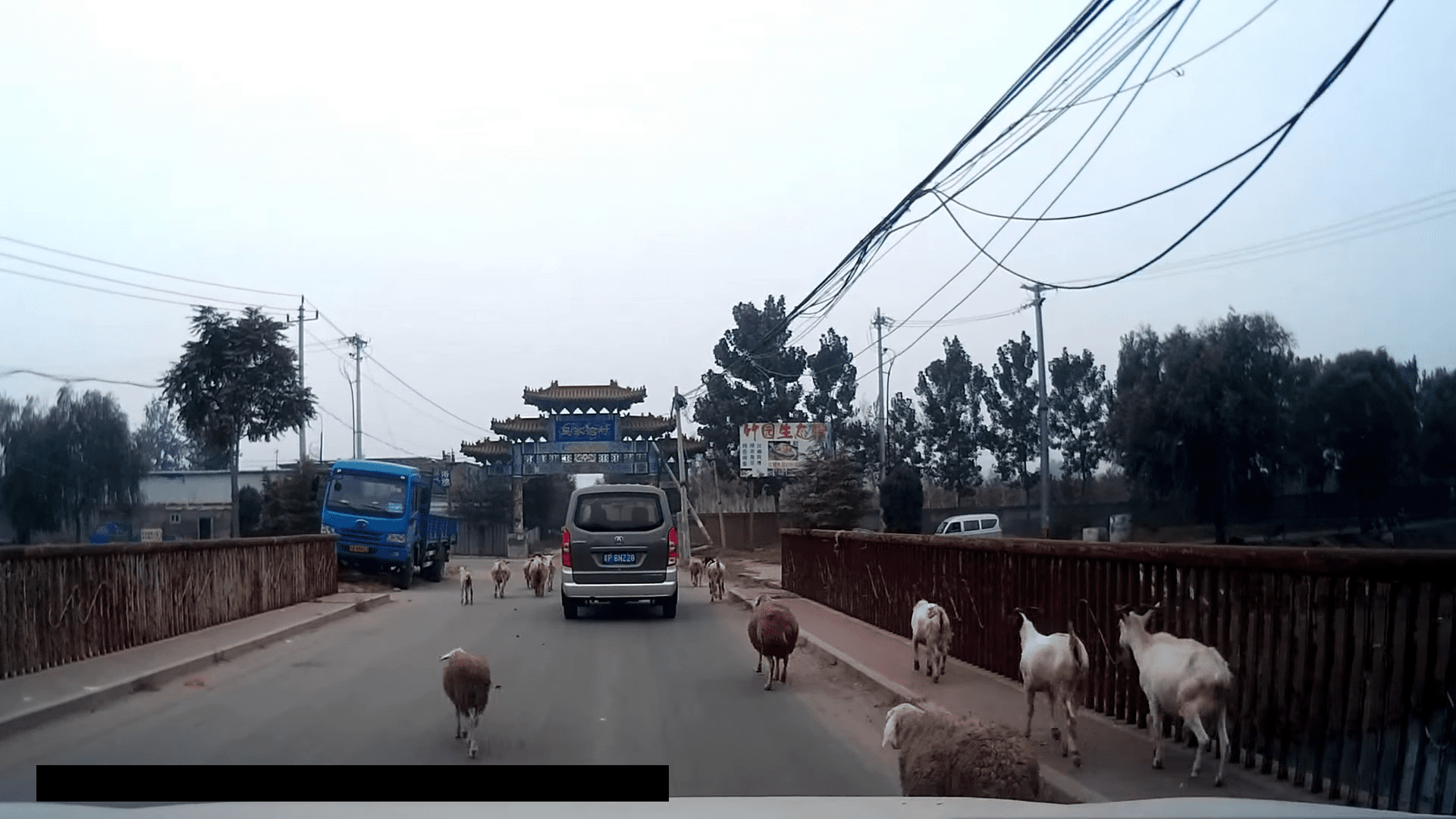}
	\includegraphics[height=\cityheight, width=\imwidth, align=b]{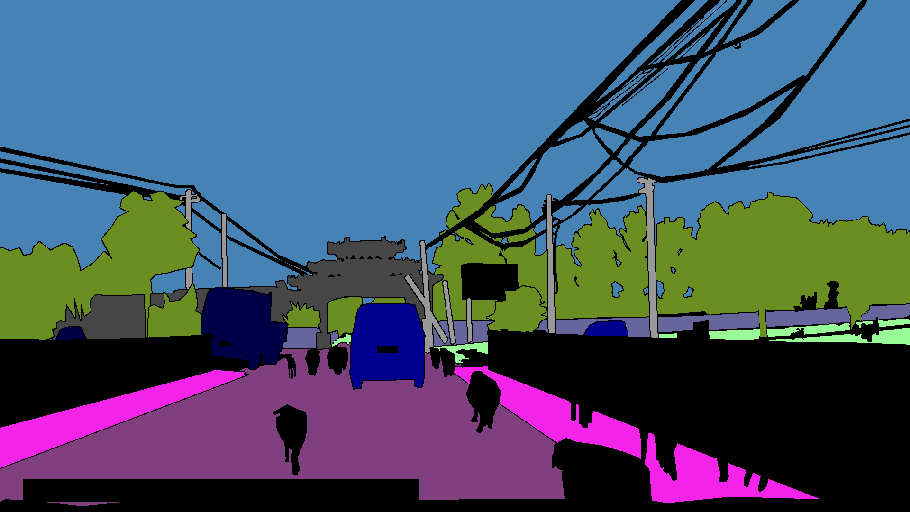}
	\includegraphics[height=\cityheight, width=\imwidth, align=b]{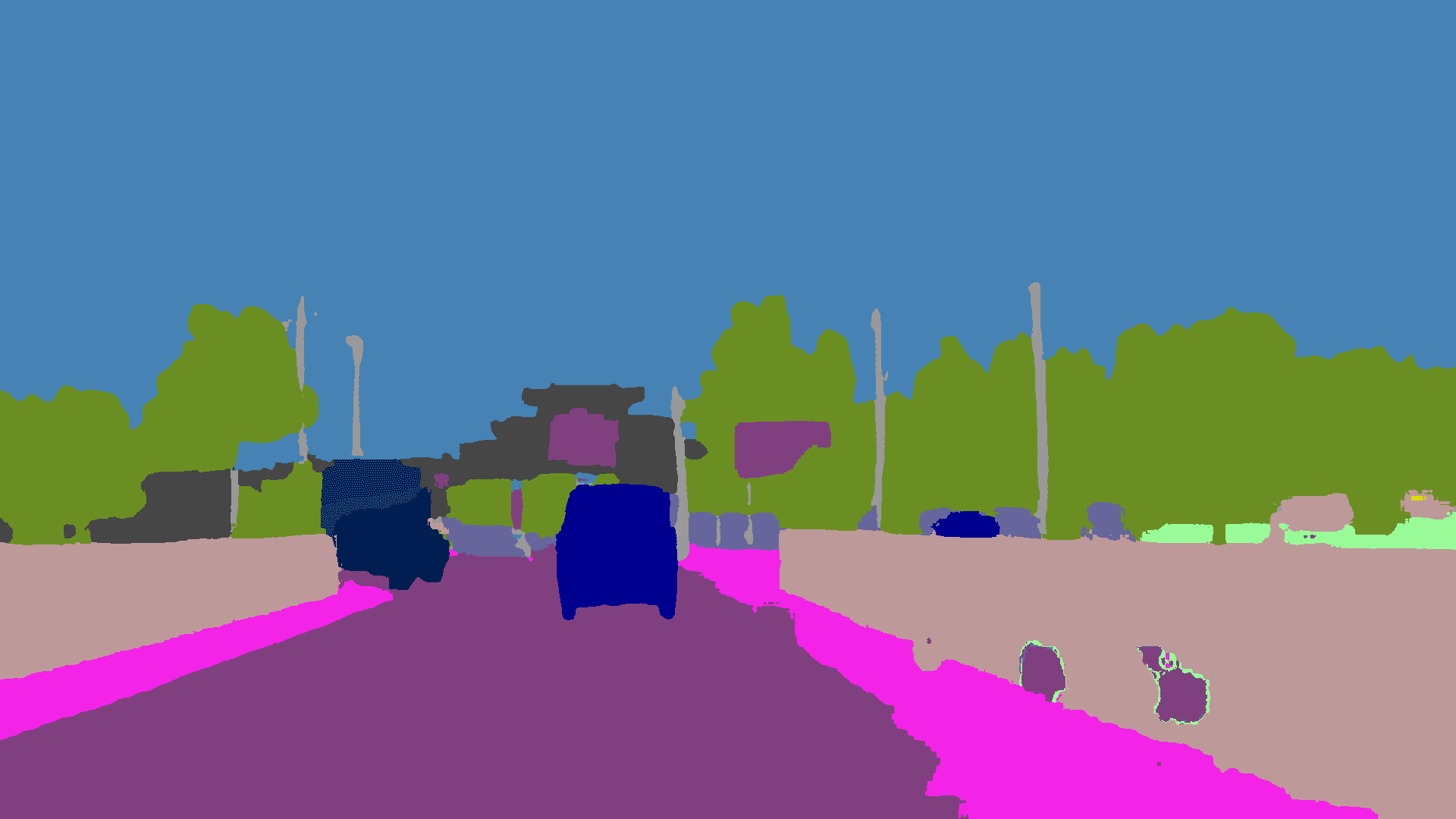}
	\includegraphics[height=\cityheight, width=\imwidth, align=b]{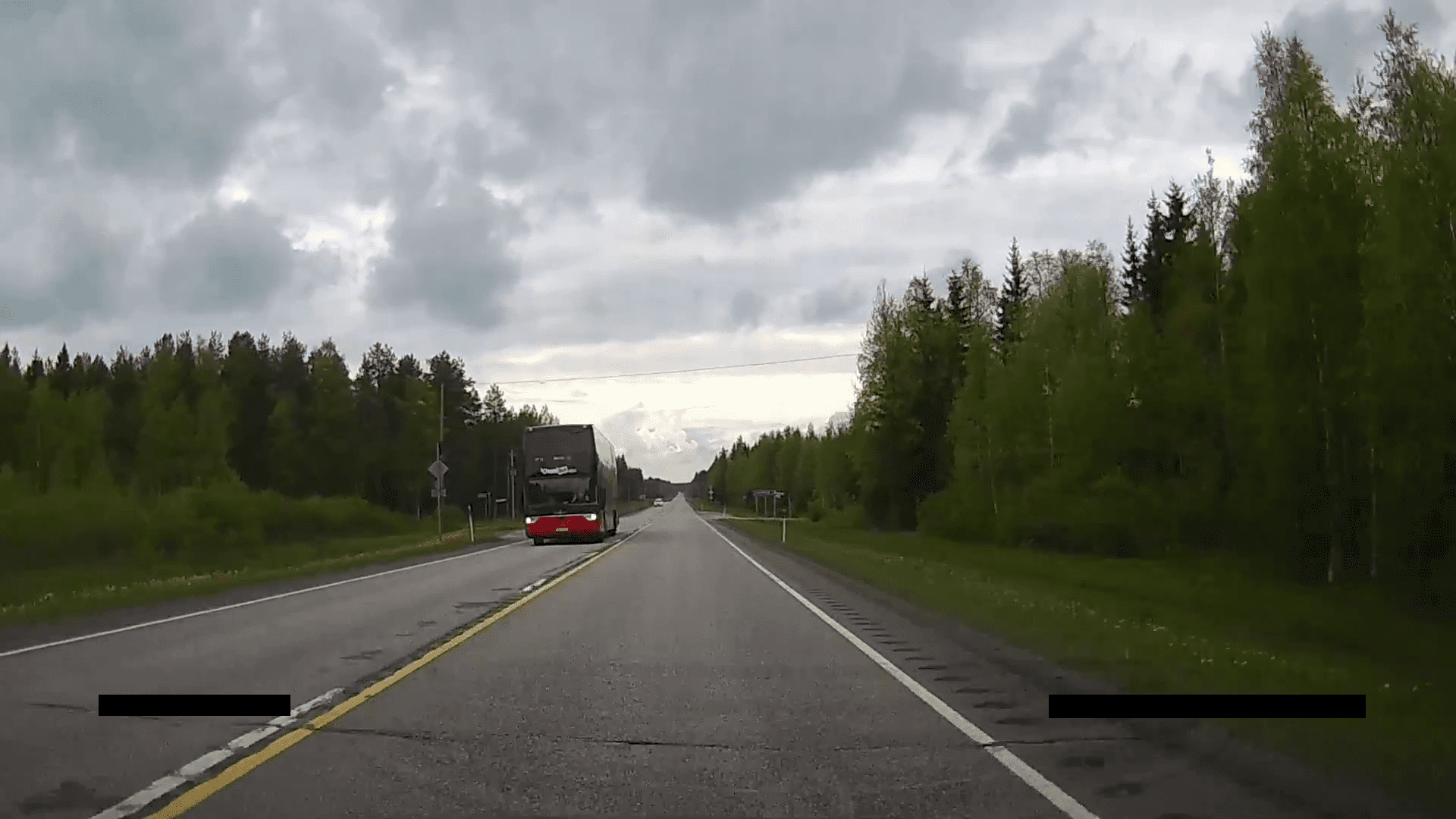}
	\includegraphics[height=\cityheight, width=\imwidth, align=b]{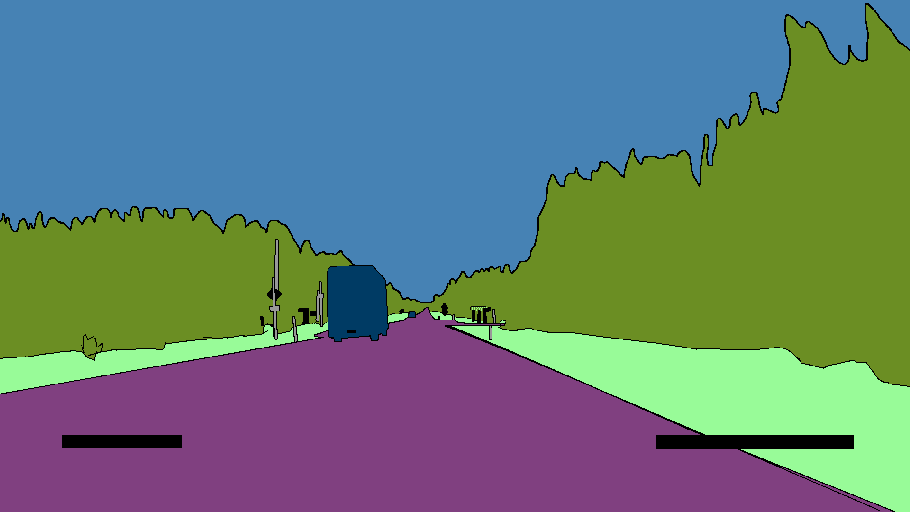}
	\includegraphics[height=\cityheight, width=\imwidth, align=b]{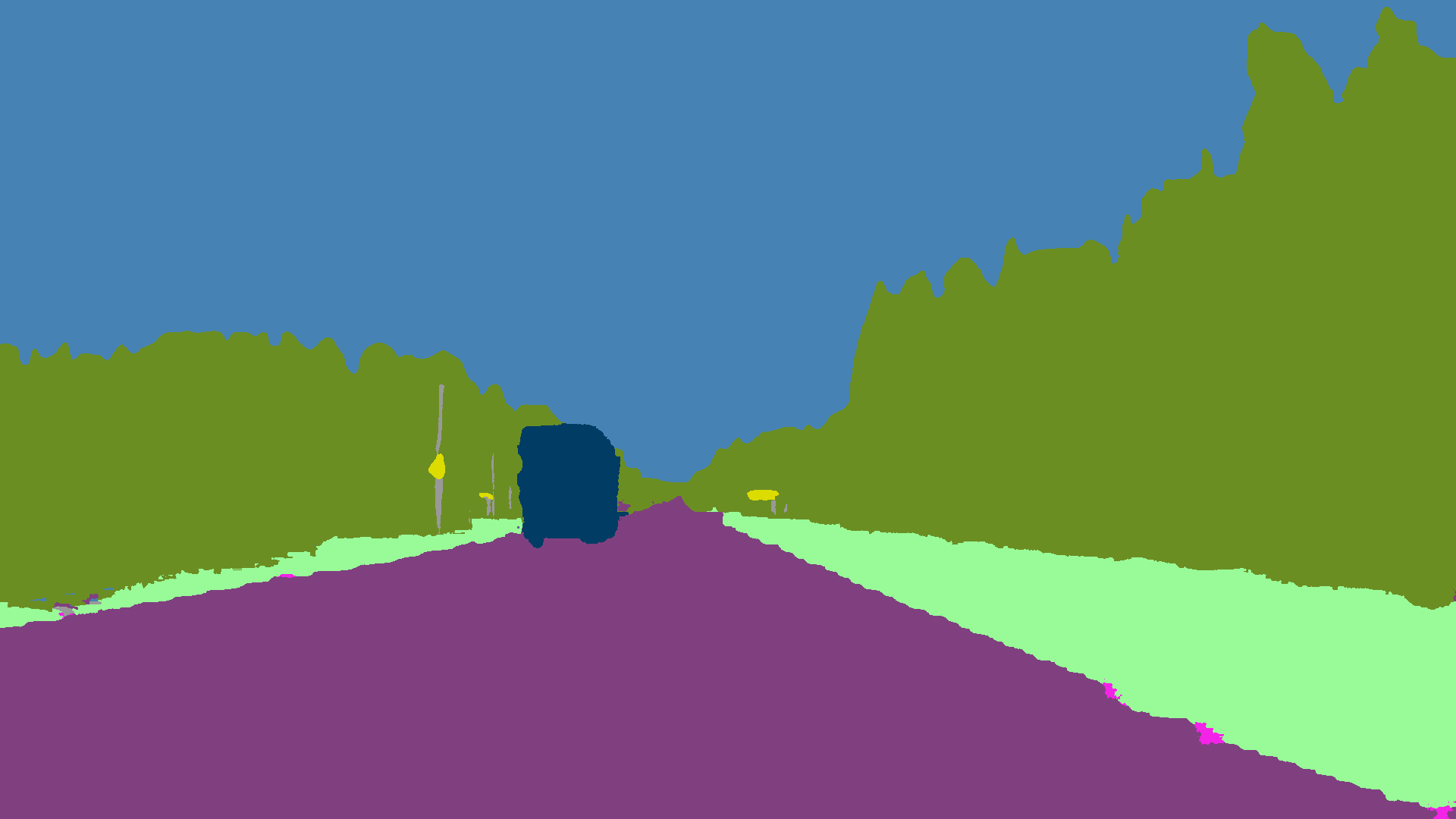} \\
	\includegraphics[height=\cityheight, width=\imwidth, align=b]{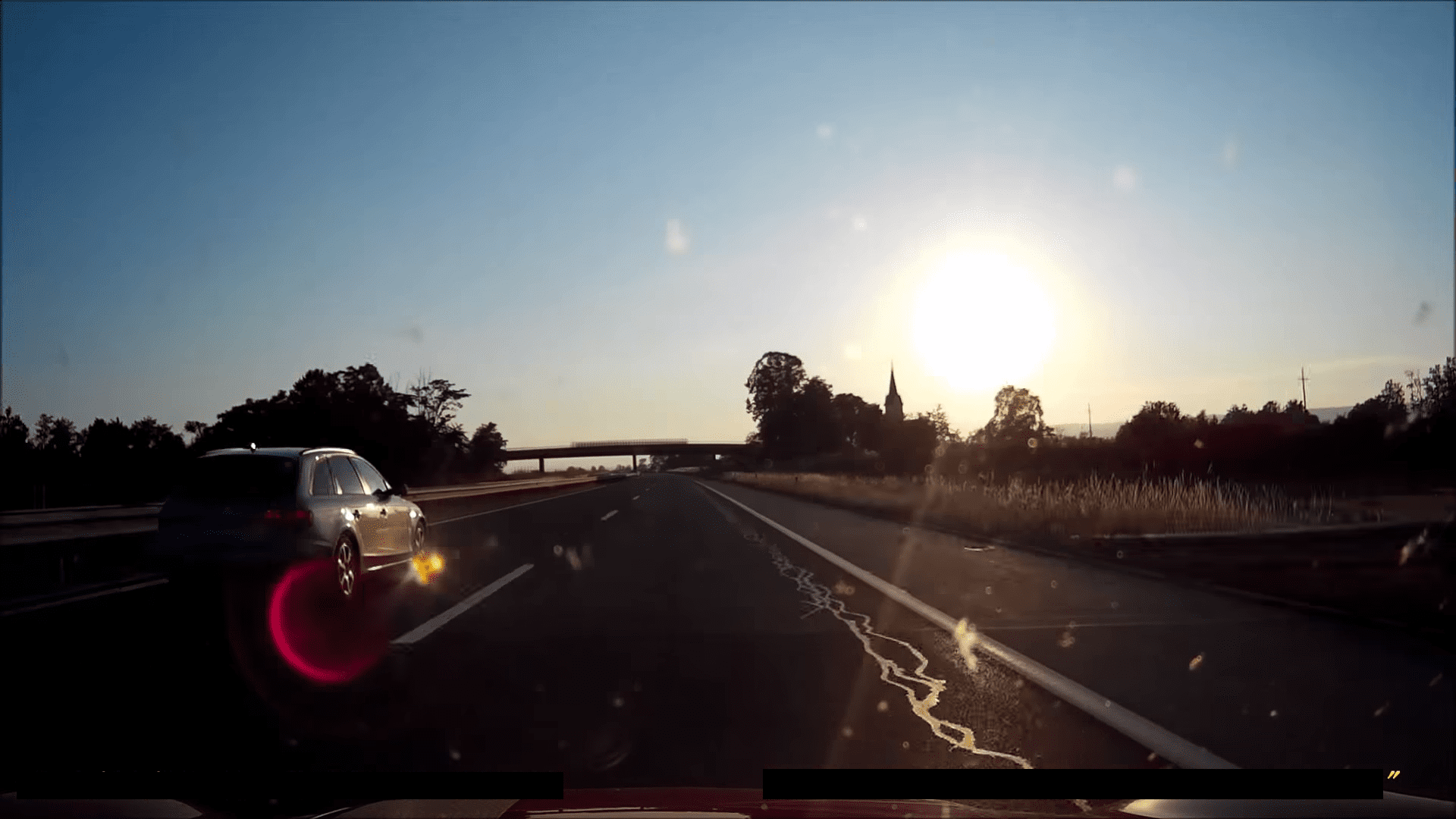}
	\includegraphics[height=\cityheight, width=\imwidth, align=b]{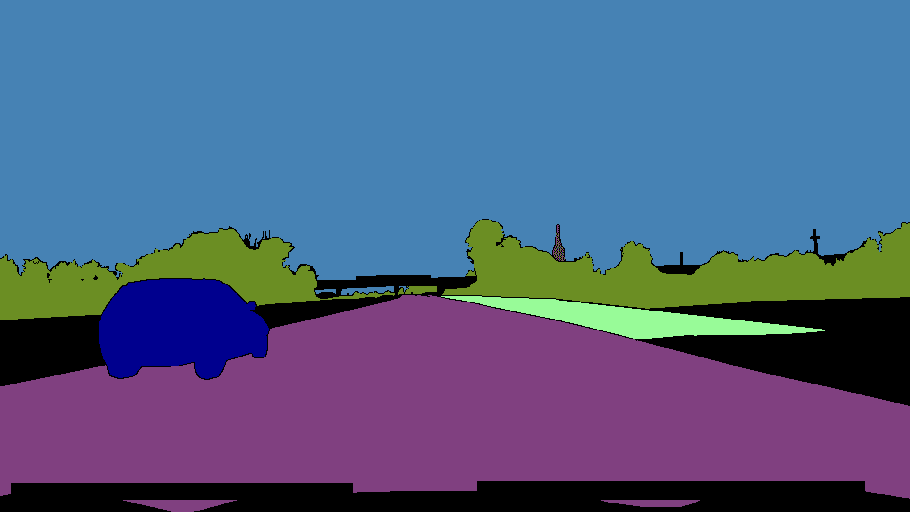}
	\includegraphics[height=\cityheight, width=\imwidth, align=b]{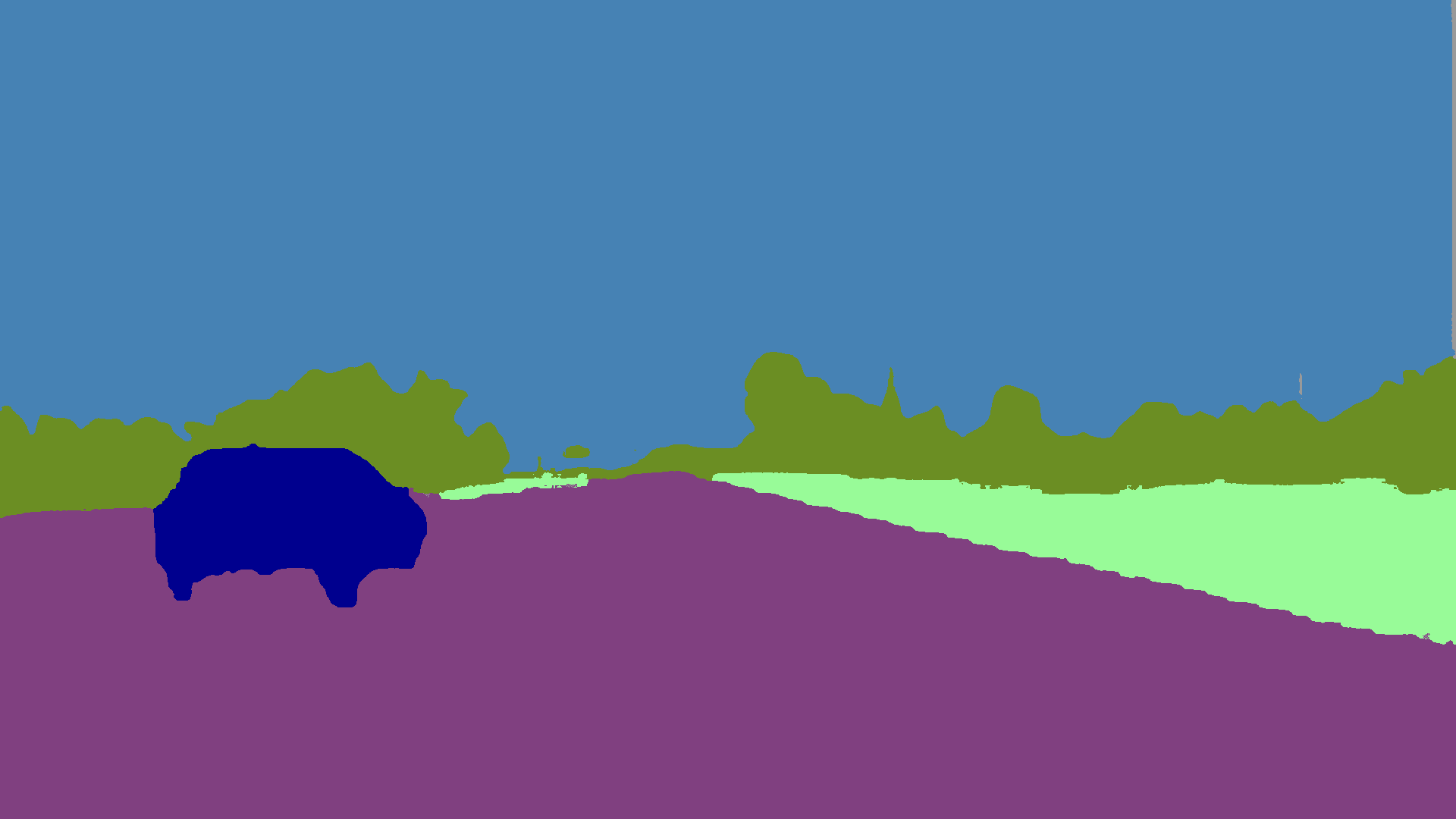}
 	\includegraphics[height=\cityheight, width=\imwidth, align=b]{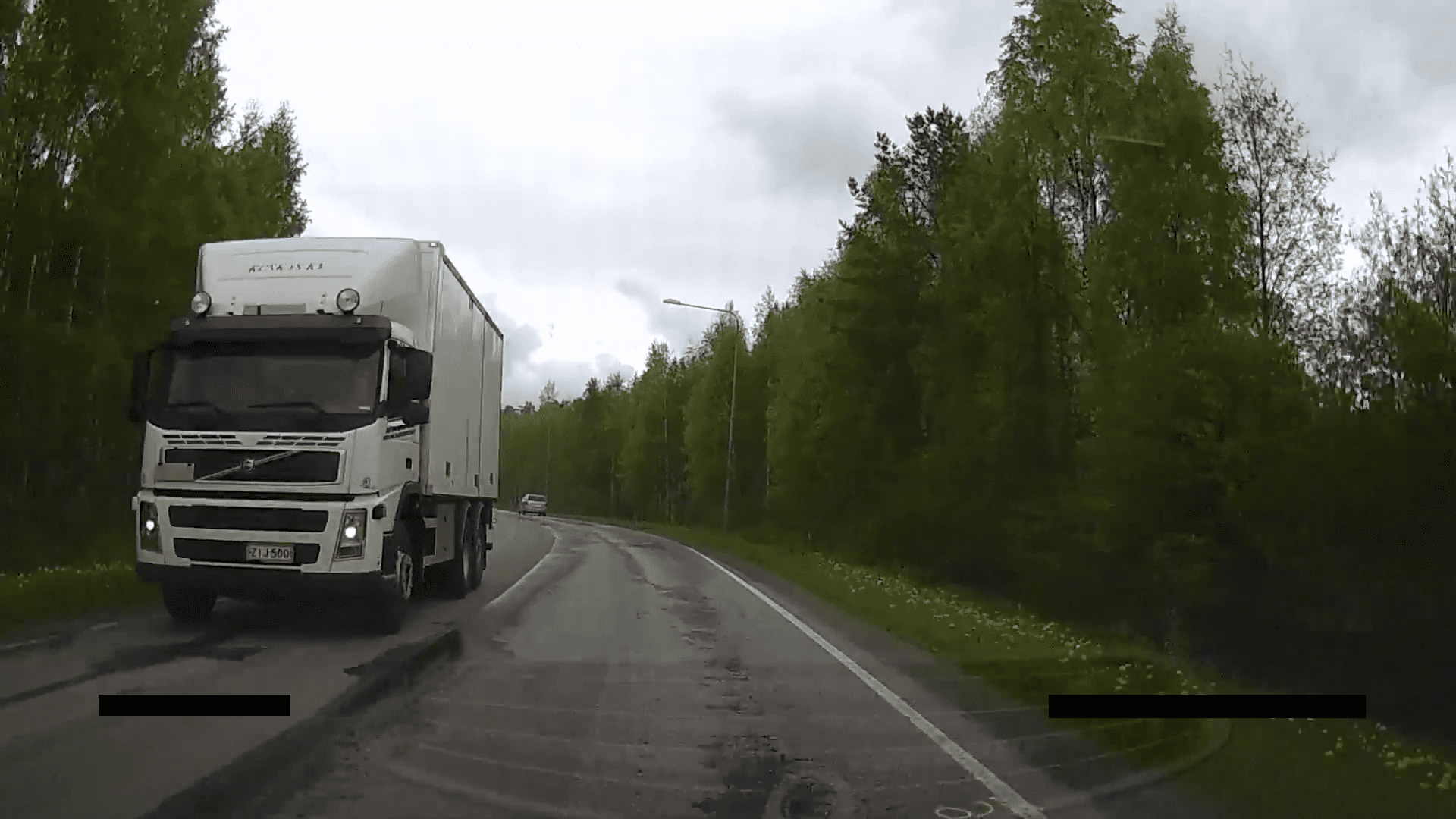}
	\includegraphics[height=\cityheight, width=\imwidth, align=b]{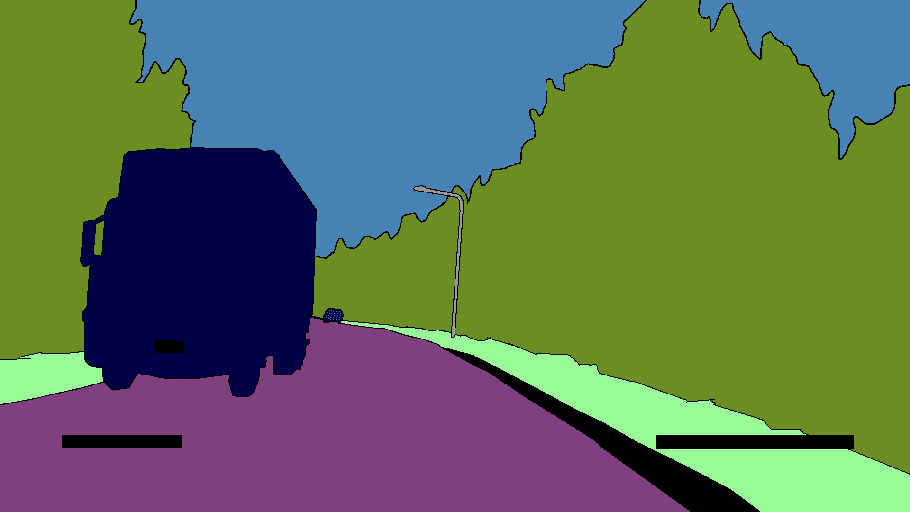}
	\includegraphics[height=\cityheight, width=\imwidth, align=b]{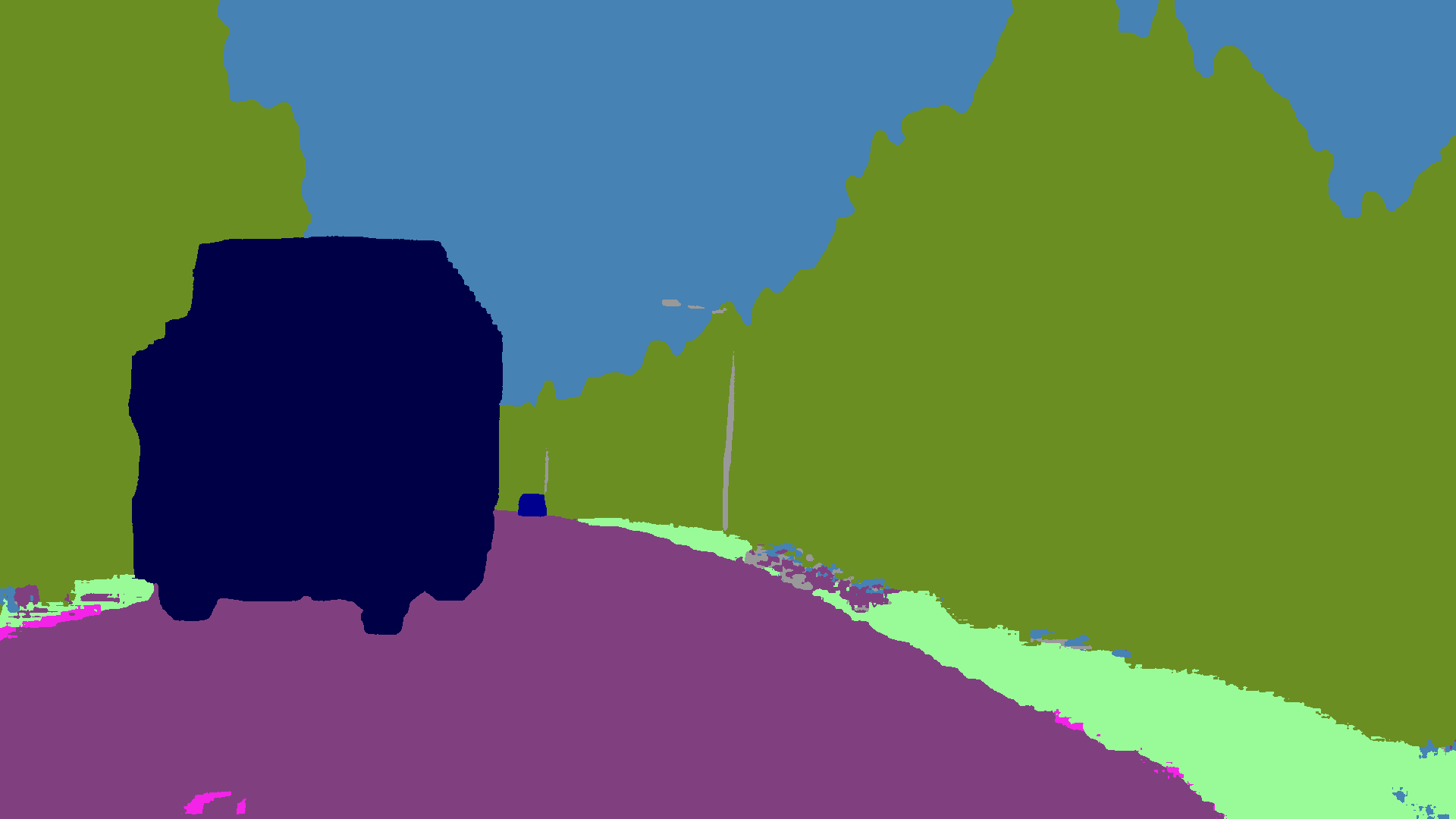}
	\rotatebox{0}{\textcolor{white}{---------}Image}
	\rotatebox{0}{\textcolor{white}{-------------}Ground Truth}
	\rotatebox{0}{\textcolor{white}{---------}Prediction}
	\rotatebox{0}{\textcolor{white}{---------------}Image}
	\rotatebox{0}{\textcolor{white}{-----------}Ground Truth}
	\rotatebox{0}{\textcolor{white}{-----------}Prediction\textcolor{white}{-------}}
	\caption{Qualitative results of semantic segmentation on MSeg {\tt test} datasets~\cite{lambert2020mseg}. From top to bottom: Pascal VOC~\cite{everingham2010pascal}, Pascal Context~\cite{mottaghi2014role}, ScanNet-20~\cite{dai2017scannet}, CamVid~\cite{brostow2009semantic} and WildDash~\cite{zendel2018wilddash} (the last two rows).}
	\label{fig:supp_visualization_stage_2_mseg_main}
\end{figure*}

\begin{figure*}[th]
	\def \imwidth {8.6cm}
    \def \vochight {2.5cm}
	\centering
	\includegraphics[height=\vochight, width=\imwidth, align=b]{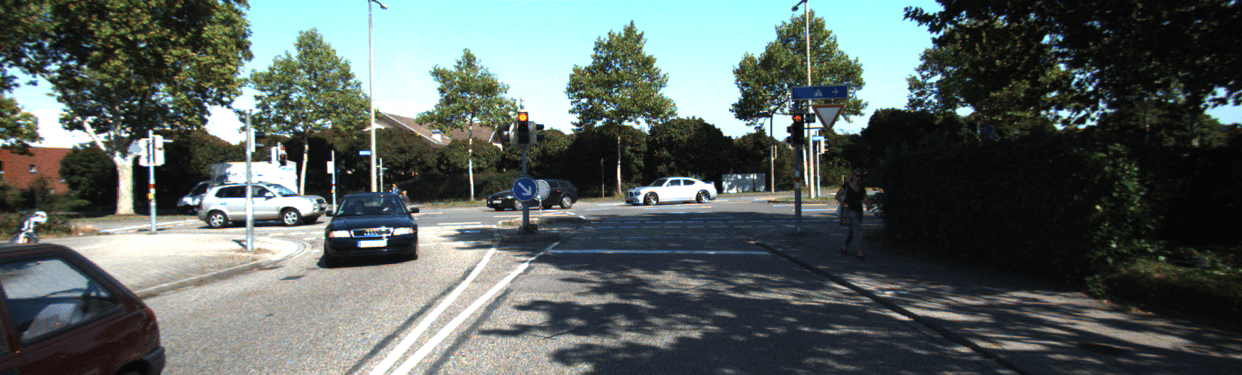} 
 	\includegraphics[height=\vochight, width=\imwidth, align=b]{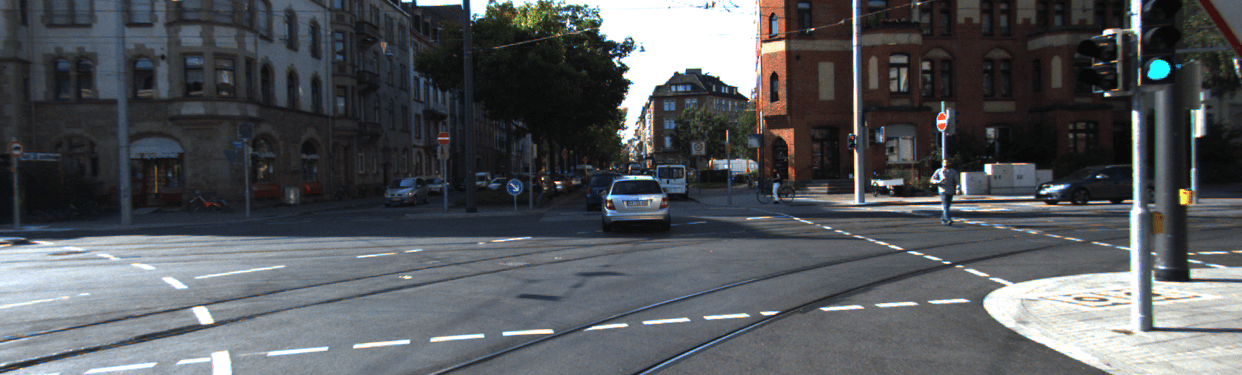} 
  \\
	\includegraphics[height=\vochight, width=\imwidth, align=b]{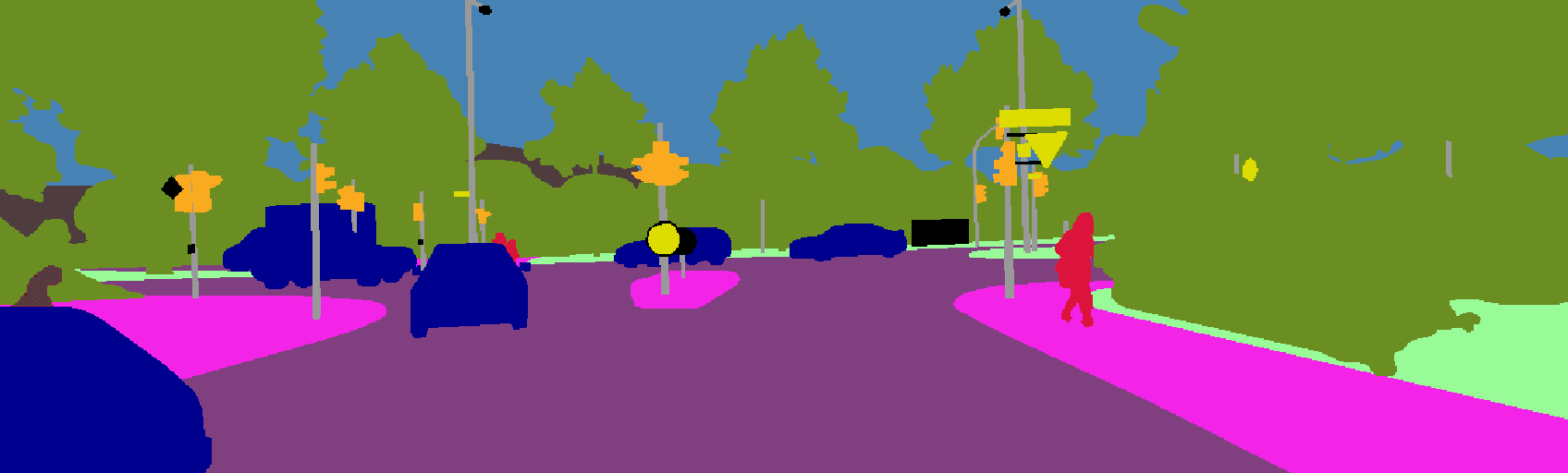} 
 	\includegraphics[height=\vochight, width=\imwidth, align=b]{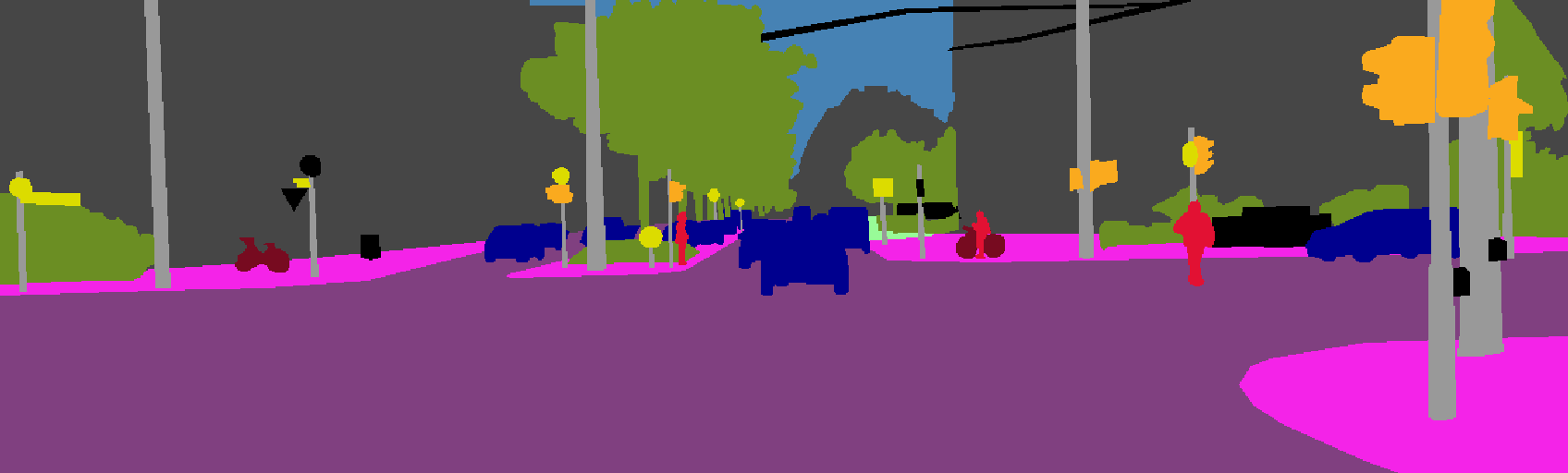} 
 \\
	\includegraphics[height=\vochight, width=\imwidth, align=b]{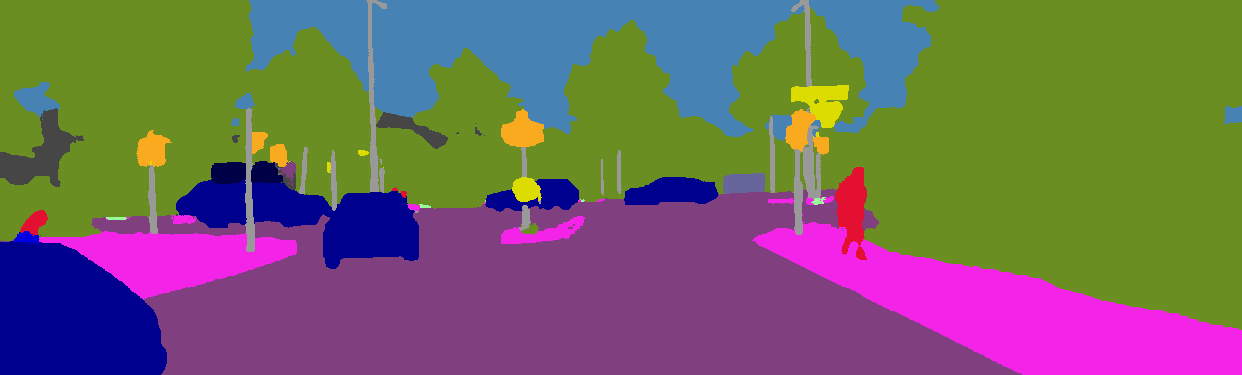} 
	\includegraphics[height=\vochight, width=\imwidth, align=b]{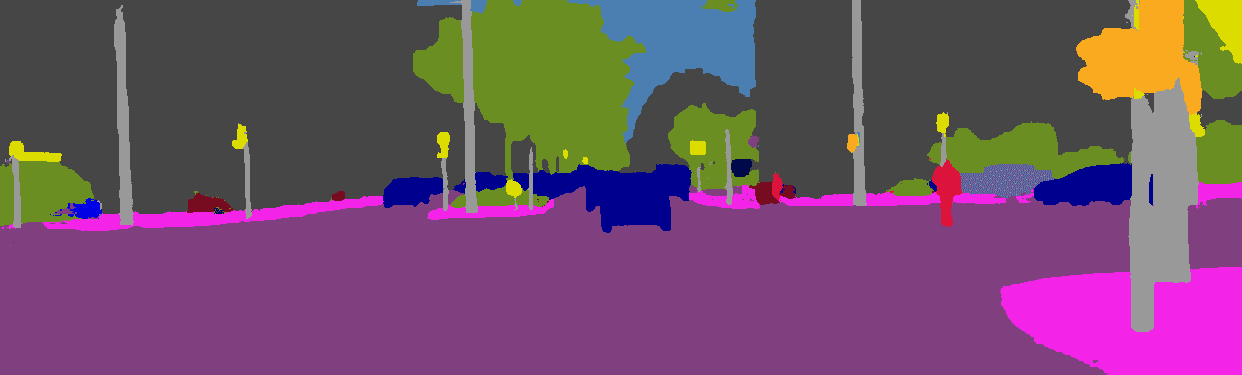}
	\\

	\caption{Qualitative results of semantic segmentation on MSeg {\tt test} dataset~\cite{lambert2020mseg} (KITTI dataset~\cite{geiger2013vision}). The $1^{st}$ row is input image, the $2^{rd}$ row is Ground Truth, and the $3^{rd}$ row is prediction result.}
	\label{fig:supp_visualization_stage_2_meg_kitti}
\end{figure*}

As shown in Figure~\ref{fig:supp_visualization_stage_2_cityscapes}, our \model{} has an accurate perception of buses, trucks and pedestrians in distance, whilst also splitting the dense and slim poles. 
In Figure~\ref{fig:supp_visualization_stage_2_ade20k}, we see that \model{} correctly recognises a wide range of furniture such as curtains, cabinets, murals, doors and toilets;
This suggests that our {\texttt maskige} generative approach can accurately represent a wide range of semantic entities. 
Figure~\ref{fig:supp_visualization_stage_2_mseg_main} and Figure~\ref{fig:supp_visualization_stage_2_meg_kitti} show the cross-domain segmentation performance
on images from previously unseen domains (Mseg {\tt test} datasets). 
It can be seen that \model{} performs well in all five datasets in the MSeg {\tt test} split~\cite{lambert2020mseg}, further validating that our generative algorithm has strong cross-domain generalization capabilities.

\section{Reproduced semantic segmentation version of UViM~\cite{kolesnikov2022uvim}}
We reproduce UViM with {\em mmsegmentation} and follow the hyperparameter and structure in the paper~\cite{kolesnikov2022uvim}.
To achieve a fair comparison with our approach, we have made some modifications: 
\textbf{(i)} we implement Swin-Large~\cite{liu2021swin} pretrained on ImageNet 22K~\cite{deng2009imagenet} as the Language model $LM$ as ours; 
\textbf{(ii)} we generate the Guiding code straightforwardly in a non-autoregressive manner; 
\textbf{(iii)} we trained 80k iterations in the first stage of UViM~\cite{kolesnikov2022uvim} and 160k iterations in the second stage. 
These modifications are necessary to ensure \textbf{a fair comparison}.

\section{Societal impact}
Given that the strong cross-domain generalization capability,  we consider our model has the potential to be used in a wide range of visual scenarios. This is desired in practical applications
due to the benefits of reducing the demands of per-domain model training and easier deployment and system management.
This is meaningful and advantageous in both economics and environment. 
On the other hand, there exist the potential to spawn abuse of our algorithm and unexpected undesirable uses. 
Therefore, it is necessary to strengthen the regulation and supervision of algorithm applications, in order to guarantee that new algorithms including ours can be used safely and responsibly for the good of the humanity and society.

\section{Limitations and Future Work}
While our study represents a significant step forward for generative segmentation, our models still fall short of the performance achieved by top discriminative models. 
One contributing factor is that decision boundaries for generative models are often less precise than those of discriminative models, resulting in less accurate object edges in segmentation. 
Another drawback is that generative models require larger amounts of data to achieve good performance, because discriminative models only learn decision boundaries, while generative models need to learn the distribution of the entire sample space. In our experiments, the performance of MSeg is better compared to Cityscapes and ADE20K, which roughly indicates this point. 

Additionally, since we convert all categories to colors, the color space is limited, and as the number of categories increases, the colors become more crowded. This can lead to confusion when using $\mathcal{X}^{-1}$ to query and predict the closest pre-defined color for each category from {\tt maskige}, especially near object edges. Therefore, it is worth trying to expand this space to higher dimensions.

Looking ahead, there are several avenues for future research in generative semantic segmentation. One promising direction is instance-level segmentation, which would enable more precise identification and separation of individual objects within an image. Additionally, we believe that it would be valuable to explore a unified model that can perform multiple vision tasks, such as segmentation, 2D object detection, depth prediction, 3D detection, and more.

Given that the second stage training of \model{} focuses on latent prior learning, new vision tasks could be inclusively added by incorporating a new posterior distribution of latent variables, without requiring any changes to the model architecture. By pursuing these directions, we believe that significant advances can be made in the field of generative semantic segmentation.


\end{document}